\documentclass{article}

\usepackage{arxiv}

\usepackage[T1]{fontenc}    
\usepackage{url}            
\usepackage{booktabs}       
\usepackage{microtype}      
\usepackage{graphicx}       
\usepackage{subcaption}     
\usepackage{float}          
\usepackage{natbib}         
\usepackage{lipsum}         
\usepackage[autostyle=tryonce]{csquotes} 

\usepackage{algorithm}      
\usepackage{algorithmicx}   
\usepackage{algpseudocode}  

\algdef{SE}[DOWHILE]{Do}{DoWhile}{\algorithmicdo}[1]{\algorithmicwhile\ #1}
\algdef{SE}[PARALLELFOR]{ParallelFor}{EndParallelFor}[1]{\textbf{parallel}
\algorithmicfor #1 \algorithmicdo}{\algorithmicend~\algorithmicfor}

\usepackage{dsfont}         
\usepackage{nicefrac}       
\usepackage{amsthm}
\usepackage{amsmath}        
\usepackage{amssymb}
\usepackage{amsfonts}       
\usepackage{mathtools}      
\usepackage{mathrsfs}

\usepackage[textsize=scriptsize]{todonotes}  
\usepackage{tikz}
\usetikzlibrary{arrows}
\usetikzlibrary{arrows.meta}
\usetikzlibrary{positioning}

\usepackage[colorlinks=true, linkcolor=., citecolor=blue, pdfusetitle]{hyperref}

\usepackage[nameinlink,capitalise,noabbrev]{cleveref}

\DeclareMathAlphabet\mathbfcal{OMS}{cmsy}{b}{n}

\newcommand{\twobyone}[2]{\left[\begin{array}{c}#1\\#2\end{array}\right]}
\newcommand{\threebyone}[3]{\left[\begin{array}{c}#1\\#2\\#3\end{array}\right]}

\newcommand{\onebythree}[3]{\left[\begin{array}{ccc}#1&#2&#3\end{array}\right]}
\newcommand{\twobytwo}[4]{\left[\begin{array}{cc}#1&#2\\#3&#4\end{array}\right]}

\newcommand{\threebythree}[9]{\left[\begin{array}{ccc}#1&#2&#3\\#4&#5&#6\\#7&#8&#9\end{array}\right]}

\newcommand{\threediag}[3]{\left[\begin{array}{ccc} #1 & & \\ & #2 & \\ & & #3 \end{array}\right]}

\newcommand{\homogen}[1]{\twobyone{#1}{1}}
\newcommand{\homogendir}[1]{\twobyone{#1}{0}}

\DeclareMathOperator*{\argmin}{argmin}
\DeclareMathOperator*{\argmax}{argmax}
\DeclareMathOperator*{\expect}{\Eds}
\newcommand{\kron}{\otimes} 

\DeclareMathOperator{\D}{D}               
\DeclareMathOperator{\Hess}{\mathsf{H}}   
\DeclareMathOperator{\diag}{diag}         
\DeclareMathOperator{\flatten}{vec}       
\DeclareMathOperator{\rank}{rank}         

\newcommand{\Eds}{\mathbb{E}} 
\newcommand{\Nds}{\mathbb{N}} 
\newcommand{\Rds}{\mathbb{R}} 

\newcommand{\Hbf}{\mathbf{H}} 

\newcommand{\Wbf}{\mathbf{W}}

\newcommand{\Acal}{\mathcal{A}}

\newcommand{\Fcal}{\mathcal{F}}
\newcommand{\Jcal}{\mathcal{J}}

\newcommand{\Scal}{\mathcal{S}} 

\newcommand{\Vcal}{\mathcal{V}}
\newcommand{\Wcal}{\mathcal{W}}

\newcommand{\Wbfcal}{\mathbfcal{W}}

\renewcommand{\top}{\mathsf{T}}     

\DeclareMathOperator{\bellop}{T_\pi}

\newcommand{\unit}{\Lambda}
\newcommand{\layer}{\unit}
\newcommand{\network}{f}

\newcommand{\Nnet}{N_{net}}

\newcommand{\weightspacehom}[2]{\Wbfcal \coloneqq \Rds^{(#1 + 1) \times n_1} \times\ldots\times
\Rds^{(n_{L-1} + 1) \times #2}}

\newcommand{\unitoperation}{\sigma\left(w_{l,k}^\top \phi_{l-1} - b_{l, k}\right)}
\newcommand{\unitdomain}{\Rds^{n_{l-1}} \times \Rds \times \Rds^{n_{l-1}} \to \Rds}

\newcommand{\unitmap}{\unit_{l, k}(w_{l, k}, b_{l, k}, \phi_{l-1}) \coloneqq \unitoperation}

\newcommand{\layeroperation}{\sigma \left( W_l^\top \phi_{l-1} + b_l\right)}
\newcommand{\layeroperationhom}{\sigma \left( W_l^\top \cdot \homogen{\phi_{l-1}} \right)}

\newcommand{\layermaphom}{\layer_l (W_l, \phi_{l-1}) \coloneqq \layeroperationhom}

\newcommand{\layercomp}{\layer_L(W_L,\cdot) \circ \ldots \circ \layer_1(W_1,\phi_{0})}

\newcommand{\netmapdomain}[2]{\Wbfcal \times \Rds^{#1} \to \Rds^{#2}}

\newcommand{\netmapsto}{(\Wbf,\phi_0) \mapsto \layercomp}

\newcommand{\netmapfull}[2]{\network \colon \netmapdomain{#1}{#2}, \quad \netmapsto}

\newcommand{\MLP}[2]{\Fcal(#1, \ldots, #2)}
\newcommand{\MLPwd}[4]{\Fcal(#1, #2 \times #3, #4)}

\newcommand{\MLPfull}[2]{\Fcal \coloneqq \big\lbrace \network(\Wbf,\cdot) \colon
\Rds^{#1} \to \Rds^{#2} ~\big|~ \Wbf \in \Wbfcal \big\rbrace}

\newtheorem{proposition}{Proposition}

\newtheorem{assumption}{Assumption}
\newtheorem{definition}{Definition}

\title{Analysis and Optimisation of Bellman Residual Errors with Neural Function Approximation}

\author{
  Martin Gottwald
  \thanks{Chair for Data Processing,
          Technical University of Munich,
          Arcisstr. 21, 80333 Munich, Germany}\\
  \texttt{martin.gottwald@tum.de} \\
  \And
  Sven Gronauer\footnotemark[1]\\
  \texttt{sven.gronauer@tum.de}\\
  \And
  Hao Shen
  \thanks{fortiss,
          Forschungsinstitut des Freistaats Bayern,
          Guerickestr. 25, 80805 Munich, Germany}\\
  \texttt{shen@fortiss.org} \\
  \And
  Klaus Diepold\footnotemark[1]\\
  \texttt{kldi@tum.de}\\
}

\hypersetup{
  pdfsubject={},
  pdfauthor={Martin Gottwald},
  pdfkeywords={Critical Point Analysis,
               Dynamic Programming,
               Gauss Newton Algorithm,
               Local Quadratic Convergence,
               Mean Squared Bellman Error,
               Residual Gradient},
}

\begin{document}

\maketitle

%
\begin{abstract}
  Recent development of Deep Reinforcement Learning has demonstrated superior performance of neural
  networks in solving challenging problems with large or even continuous state spaces.
  One specific approach is to deploy neural networks to approximate value functions by minimising
  the Mean Squared Bellman Error function.
  Despite great successes of Deep Reinforcement Learning, development of reliable and efficient
  numerical algorithms to minimise the Bellman Error is still of great scientific interest and
  practical demand.
  Such a challenge is partially due to the underlying optimisation problem being highly non-convex
  or using incomplete gradient information as done in Semi-Gradient algorithms.
  In this work, we analyse the Mean Squared Bellman Error from a smooth optimisation perspective and
  develop an efficient Approximate Newton's algorithm.\\
  First, we conduct a critical point analysis of the error function and provide technical insights
  on optimisation and design choices for neural networks.
  When the existence of global minima is assumed and the objective fulfils certain conditions,
  suboptimal local minima can be avoided when using over-parametrised neural networks.
  We construct a Gauss Newton Residual Gradient algorithm based on the analysis in two variations.
  The first variation applies to discrete state spaces and exact learning.
  We confirm theoretical properties of this algorithm such as being locally quadratically
  convergent to a global minimum numerically.
  The second employs sampling and can be used in the continuous setting.
  We demonstrate feasibility and generalisation capabilities of the proposed algorithm empirically
  using continuous control problems and provide a numerical verification of our critical point
  analysis.
  We outline the difficulties of combining Semi-Gradient approaches with Hessian information.
  To benefit from second-order information complete derivatives of the Mean Squared Bellman Error
  must be considered during training.
\end{abstract}


%
\section{Introduction}
\label{sec:introduction}
%
%
\emph{Reinforcement Learning} (RL), or more general \emph{Dynamic Programming} (DP), is a general
purpose framework to solve sequential decision making or optimal control problems.
To handle problems with large or even continuous state spaces \emph{Value Function Approximation}
(VFA) is used as an important and effective instrument \citep{bert:book12, sutt:book20}.
%
%
Such VFA methods can be categorised in two groups, namely \emph{linear} and \emph{non-linear}.
Various efficient \emph{Linear Value Function Approximation} (LVFA) algorithms have been developed
and analysed in the field, e.g. \citep{nedi:deds03, bert:chap04, parr:icml08, geis:tnn13}.
Despite their significant simplicity and convergence stability, the performance of LVFA methods
depends heavily on construction and selection of linear features, which is a time consuming
and hardly scalable process \citep{parr:icml07, rohm:jmlr13}.
Therefore, recent research efforts have focused more on
\emph{Non-Linear Value Function Approximation} (NL-VFA) methods.

%
%
As a popular non-linear mechanism, kernel tricks have been successfully adopted to VFA, and
demonstrated their convincing performance in various applications \citep{xuxi:tnn07, tayl:icml09,
bhat:nips12}.
Unfortunately, due to the nature of kernel learning, these algorithms can easily suffer from a high
computational burden due to the required number of samples.
Furthermore, kernel-based VFA algorithms can also have serious problems with over-fitting.
As an alternative, \emph{Neural Networks} (NN) have been another common and powerful approach to
approximate value functions \citep{linl:phd93, bert:book96}.
Impressive successes of NNs in solving challenging problems in pattern recognition, computer vision,
and speech recognition and game playing \citep{lecu:nature15, yudo:book15, mnih:nature15,
silver:nature17} have further triggered increasing efforts in applying NNs to VFA
\citep{hass:aaai16}.
More specifically, \emph{NN-based Value Function Approximation} (NN-VFA) approaches have
demonstrated their superior performance in many challenging domains, e.g. \emph{Atari} games
\citep{mnih:nature15} or the game \emph{Go} \citep{silver:nature16, silver:nature17}.
Despite these advances, development of more efficient NN-VFA based algorithms is
still of great demand for even more challenging applications.
So far, these impressive successes are generally only possible, if a plethora of training samples is
available.

%
%
Aside from some early work in \citep{bair:icml95}, called \emph{Residual Gradient} (RG) algorithms,
omitting gradients of the TD-target has been a common practice, see e.g. \citep{ried:ecml05}, and
is recently referred to as \emph{Semi-Gradient} (SG) algorithms \citep{sutt:book20}.
Reasons for choosing Semi-Gradients over Residual Gradients include inferior learning speed
\citep{bair:icml95}, limitation with non-Markovian feature space \citep{sutt:nips08} and
non-differentiable operators, e.g. the max-operator involved in Q-learning.
Residual Gradients are favoured for their convergence guarantees and applicability of
\enquote{classic} gradient based optimisation techniques.

%
%
In this work, we study the problem of minimising the \emph{Mean Squared Bellman Error} (MSBE) with
NN-VFA from the global analysis perspective.
More specifically, we work in the framework of geometric optimisation \citep{absi:book08}.
We conduct a critical point analysis of the MSBE including its Hessian and also derive a proper
approximation for it.
We obtain insights in the learning process and can prevent the existence of saddle points or
undesired local minima by requiring over-parametrisation of the NN-VFA and by ensuring certain
properties of the optimisation objective.
Furthermore, our analysis leads to an efficient and effective \emph{Approximate Newton}'s (AN)
algorithm.
It is further investigated for a continuous state space setting whether or not ignoring the
dependency of derivatives on the network parameters in the TD-target has a significant impact,
especially in the context of second-order optimisation.
We outline how to overcome the convergence speed issues of Residual Gradients and show that
gradients and higher order derivatives of the TD-target provide critical information about the
optimisation problem.
They are essential for implementing efficient optimisation algorithms and result in solutions with
higher quality.
Finally, we conduct several experiments to confirm the results of our critical point analysis
numerically and also investigate generalisation capabilities of NN-VFA methods empirically.

%
%
The rest of this paper is arranged as follows.
In the next section, we review the existing work regarding the construction and analysis of
algorithms.
In \cref{sec:notation}, we give an introduction to RL with VFA and provide the necessary notational
conventions around NNs since a concise notation is required for our analysis in later sections.
We conduct a critical point analysis of the MSBE when using NN-VFA in \cref{sec:theory} and present
our results separately for discrete and continuous state spaces in
\cref{ssec:crit_point_discrete,ssec:crit_point_continuous}, respectively.
\cref{sec:algorithm} is dedicated to our proposed Approximate Newton's Residual Gradient algorithm
and addresses details relevant for implementation.
In \cref{sec:experiments}, we evaluate performance and generalisation capabilities of the proposed
method in several experiments.
Finally, a conclusion is given in \cref{sec:conclusion}.

%
\section{Related Work}
\label{sec:related_work}
%
%
Recent attempts towards developing efficient NN-VFA methods follow the approach of extending the
well-studied LVFA algorithms. 
These are a general family of gradient-based temporal difference algorithms, which have been
proposed to optimise either the \emph{Mean Squared Bellman Error} \citep{bair:icml95, bair:nips99}
or the \emph{Mean Squared Projected Bellman Error} \citep{sutt:nips08, sutt:icml09}.
The work in \citep{maei:nips09} adapts the results of developing the so-called \emph{Gradient
Temporal Difference} (GTD) algorithms to a non-linear smooth VFA manifold setting.
The proposed approach requires projections onto some smooth manifold, which is practically
infeasible because the geometry of VFA manifolds is in general not available.
Similarly, the approach developed in \citep{silver:ewrl12} projects estimates of the value function
directly onto the vector subspace spanned by the parameter matrices of the NNs.
Unfortunately, no further analysis or numerical development exists besides the original work.

%
%
Such a difficulty in studying and developing NN-VFA methods is partially due to incomplete
theoretical understanding of training NNs.
The main challenge of the underlying optimisation problem is the strong non-convexity.
Although there are several efforts towards characterising the optimality of NN-training, e.g.
\citep{kawa:nips16, nguy:icml17, haef:cvpr17, yunc:iclr18}, a complete answer to the question is
still missing and demanding.
The work in \citep{shen:cvpr18, shen:corr18, shen:icml19} addresses training of NNs using theory
of differential topology and smooth optimisation.
It has highlighted the importance of over-parametrisation to ensure proper convergence to a
solution.
In this work, we introduce those techniques to the RL domain.

%
%
Lately, there has been more interest in Residual Gradient algorithms.
As described in \citep{bair:icml95}, Residual Gradient algorithms possess convergence guarantees
since they use a complete gradient of a well-defined performance objective.
Thus, they are eligible for a critical point analysis.
Unfortunately, these guarantees are coupled to solving the \emph{Double Sampling} issue, i.e., the
requirement of having several possible successors for every state to capture stochastic transitions.
In a recent work \citep{saleh:nips19}, the authors explore the application of deterministic
Residual Gradient algorithms to bypass the Double Sampling issue when using the \emph{Optimal
Bellman Operator}.
Furthermore, they characterise empirically the impact of stepwise increased noise in environments
and can motivate reviving Residual Gradient algorithms.
In this work, we investigate Residual Gradient algorithms in deterministic problems for Policy
Evaluation instead of aiming directly at the optimal value function.
This allows us to use a Policy Iteration scheme as done in \citep{gott:ewrl18}.

%
%
Another work \citep{cai:corr19}, which is also strongly related to ours, pursues a similar idea,
namely the importance of over-parametrisation.
However, the authors address Semi-Gradient algorithms and thus work in a different setting.
They show that the usage of NNs for NL-VFA with a redundant amount of adjustable parameters is
mandatory for achieving good performance.
They establish an implicit local linearisation and enable reliable convergence to a global optimum
of the Mean Squared Projected Bellman Error.
In \citep{brand:iclr20}, reliable convergence under over-parametrisation is also confirmed when
treating Semi-Gradient TD-Learning as ordinary differential equation and investigating stationary
points of the associated vector field.
The Jacobian of over-parametrised networks when evaluated for all discrete states can have full
rank, which is required for their theoretical investigation.
Using an empirical approach proposed in \citep{fu:pmlr19}, the authors arrive at the conclusion that
larger NL-VFA architectures result in smaller errors and boost convergence.
In \citep{boyi:nips19}, the role of over-parametrisation is classified as an important ingredient
for a variant of \emph{Proximal Policy Optimisation} \citep{schulman:corr17} to converge to an
optimum.
%
%
In the present work, we confirm the role of over-parametrisation when using Residual Gradient
algorithms and provide further insights from the global analysis perspective.

%
\section{Notation and Technical Preliminaries}
\label{sec:notation}
This section contains various definitions and provides a concise foundation for \cref{sec:theory}.
First, we outline the Reinforcement Learning setting in general.
Second, we introduce the usage of function approximation architectures and formulate the
optimisation problem we want to investigate.
Lastly, we define the function class and its components used for the NL-VFA method we are going to
analyse.

\subsection{Reinforcement Learning}
\label{ssec:notation_rl}
As common approach, we model the decision making problem as a \emph{Markov Decision Process} (MDP)
by defining the tuple $ (\Scal, \Acal, P, r, \gamma) $.
As state space $ \Scal $ we consider both finite countable sets of $ K $ discrete elements
$ \Scal = \left\lbrace 1, 2, \ldots, K \right\rbrace $ as well as compact subsets of finite
dimensional Euclidean vector spaces $ \Scal \subset \Rds^K $.
Depending on the state space, we denote with slight abuse of notation by $ K \coloneqq |\Scal| $
either the cardinality of the finite set or by $ K \coloneqq \dim(\Scal) $ the dimension of the
subspace.
The action space $ \Acal $ is always a finite set of discrete actions an agent can choose from.
%
%
The conditional transition probabilities $ P \colon \Scal \times \Acal \times \Scal \to [0, 1] $ are
either available directly in the form of analytic models or can be collected by using simulators and
performing roll-outs.
For discrete spaces, they define the probability $ P(s'|s, a) $ for transiting from state $ s $ to
$ s' $ when executing action $ a $.
In continuous state spaces, $ P $ takes the role of a probability density function which must be
integrated over.
A scalar \emph{reward} function $ r \colon \Scal \times \Acal \times \Scal \to [-M, M] $ with
$ M \in \Rds $ assigns an immediate and finite one-step-reward to the transition triplet
$ ( s, a, s' ) $.
Finally, $ \gamma \in (0, 1) $ represents a \emph{discount factor}, which is required to ensure
convergence of the overall expected discounted reward.
The goal of an agent is to learn a \emph{policy} $ \pi $, which maximises the expected discounted
reward.
It is sufficient to consider deterministic policies of the form $ \pi \colon \Scal \to \Acal $, as
the space of \emph{history independent Markov policies} can be proven%
\footnote{e.g. chapter one and two in \citep{bert:book12}, applies also to other statements here}
to contain an optimal policy.
%

The expected discounted reward starting in some state $ s \in \Scal $ and following the policy
$ \pi $ afterwards is called \emph{value function} and is defined as
\begin{equation}
  \label{eq:def_V_pi}
  V_\pi \colon \Scal \to \Rds,
  \quad
  s \mapsto \lim_{T\to\infty} \expect_{s_1, s_2, \ldots, s_T} \left[ \sum_{t=0}^T \gamma^t
  r(s_t, \pi(s_t), s_{t+1}) \Big| s_0 = s \right].
\end{equation}
The computation of expectations depends on the type of the state space.
For discrete sets, the expression takes the form of a weighted sum.
In continuous spaces, we need to compute an integral over the successor states.
It is well known that for a given policy $ \pi $ the value function $ V_\pi $ satisfies the
\emph{Bellman equation}, i.e., we can write for all states $ s \in \Scal $
\begin{align}
  \nonumber
  V_\pi(s)
  & = \expect_{s'} \Big[ r(s, \pi(s), s') + \gamma V_\pi (s') \Big]\\
  \label{eq:def_bell_eq}
  & = \sum_{s'} P(s,\pi(s),s') \Big(r(s, \pi(s), s') + \gamma V_\pi(s') \Big),
\end{align}
where $ s' $ is the successor state of $ s $ obtained by executing action $ a = \pi(s) $.
The second equality in \cref{eq:def_bell_eq} is only available if transition probabilities are
known and exist due to discrete state and action spaces.
Treating $ V_\pi \in \Vcal $ as a variable, where $ \Vcal $ is the space of all possible value
functions, the right hand side of \cref{eq:def_bell_eq} induces the \emph{Bellman Operator}
under the policy $ \pi $ and is denoted by $ \bellop \colon \Vcal \to \Vcal $.
It can be shown that $ \bellop $ is a contraction mapping with modulus $ \gamma $ and,
consequently, that the value function $ V_\pi $ is its unique fixed point.
Hence, we can write
\begin{equation}
  \label{eq:def_bell_eq_short}
  V_\pi(s) = (\bellop V_\pi)(s) \quad \forall s \in \Scal.
\end{equation}
Finally, given an arbitrary $ V \in \Vcal $, the difference of both sides in
\cref{eq:def_bell_eq_short} is known as the \emph{one-step Temporal Difference error}
\begin{equation}
  \label{eq:TD-error}
  \delta(s) \coloneqq V(s) - \left( \bellop V \right)(s) \quad \forall s \in \Scal.
\end{equation}
The term \emph{Temporal Difference} (TD) emphasizes the occurrence of the current state in $ V(s) $
and the successor state $ s' $ in $ (\bellop V)(s) $.
In this context one also calls the application of the Bellman operator \emph{TD target}.
The TD error is non-zero for all but the correct value function.
Thus, we can use $ \delta $ to convert the fixed point iteration into a root finding problem.
By combining the squared TD-error for all states we obtain a performance objective that can be
minimised.
We will use this objective together with function approximation architectures to conduct our
critical point analysis in a later section.

To avoid extensive computation when working with policies and state-only value functions, it is
possible to define similar to \cref{eq:def_V_pi} so-called \emph{$ Q $-factors} as
\begin{equation}
  \label{eq:def_Q_pi}
  Q_\pi \colon \Scal \times \Acal \to \Rds,
  \quad
  s,a \mapsto \lim_{T\to\infty} \expect_{s_1, s_2, \ldots, s_T} \left[ \sum_{t=0}^T \gamma^t
  r(s_t, \pi(s_t), s_{t+1}) \Big| s_0 = s, a_0 = a \right]
\end{equation}
and the corresponding \emph{Bellman equation in Q}
\begin{equation}
  \label{eq:def_bell_eq_in_Q}
  Q_\pi(s, a) = \expect_{s'} \Big[ r(s, a, s') + \gamma Q_\pi (s', \pi(s'))\Big]
              = (T_\pi Q_\pi)(s, a)
\end{equation}
for all $ (s, a) \in \Scal \times \Acal $.
While all algorithms work with $ Q $-factors as they do for state-only value functions, it is now
possible to define a \emph{greedily induced policy} (GIP) compactly as
$ \pi' \colon \Scal \to \Acal $ with
\begin{equation}
  \label{eq:def_gip}
  \pi'(s) \in \argmax_{a \in \Acal} Q_\pi(s, a).
\end{equation}
The new GIP $ \pi' $ satisfies $ \pi' \geq \pi $ in the sense that $ Q_{\pi'}(s,a) \geq
Q_\pi(s,a) $ for all states and actions with the equality being true for optimal policies.

\subsection{Reinforcement Learning with Function Approximation}
\label{ssec:definition_func_approx}
If a state space is finite but too large to be stored in memory, or even has to be treated as
infinite due to its size, an exact representation of the value function $ V_\pi(s) $ is practically
impossible.
This phenomenon is often referred to as the \emph{curse of dimensionality}.
Also, for continuous state spaces, where a proper iteration over all states is not available due to
the lack of a natural discrete formulation, exact representations are out of reach.
An accurate value function approximation is thus useful and necessary for representing the actual
value function $ V_\pi $ of the current policy $ \pi $ in any computer system.

Let $ F \colon \Scal \to \Rds $ be an arbitrary value function approximation and let us now denote
by $ \Vcal $ the space of approximations considered.
If we further restrict the decision making problem to be \emph{ergodic}, there exists a \emph{steady
state distribution} $ \xi_i \in (0,1) $ for each%
\footnote{for continuous spaces $ \xi $ is also a probability density function and must be used
with integrals}
state $ s_i $, which allows us to create a quality assessment for the approximation when combined
with the aforementioned objective \cref{eq:TD-error}.
The combination of a $ \xi $-weighted norm and the TD-error yields the \emph{Mean Squared Bellman
Error}
\begin{align}
  \label{eq:msbe_norm_xi}
  \text{MSBE}(F) = \frac12 \Big\| F - (\bellop F) \Big\|^2_\xi,
\end{align}
which accepts a value function approximation $ F \in \Vcal $ and returns a scalar value.
The factor $ \nicefrac12 $ is included for convenience.
As the MSBE contains a norm over a function space, the realisation of $ \left\|\cdot\right\|_{\xi} $
comes in two different ways, depending on the type of state space.
In discrete spaces we can enumerate all states and write down directly
\begin{align}
  \label{eq:msbe_discrete}
  \text{MSBE}(F)
  = \frac12 \Big\| F - (\bellop F) \Big\|_\xi^2
  = \frac12 \sum_{s \in \Scal} \xi(s) \Big( F(s) - (\bellop F)(s) \Big)^2.
\end{align}
Since the steady state distribution is a probability, we can treat the summation also as expectation
and approximate it by Monte Carlo sampling.
We have
\begin{align}
  \label{eq:msbe_mc_discrete}
  \text{MSBE}(F)
  = \frac12 \expect_{s \in \Scal} \left[  \Big( F(s) - (\bellop F)(s) \Big)^2 \right]
  \approx \frac{1}{2N} \sum_{i=1}^N \Big( F(s_i) - (\bellop F)(s_i) \Big)^2,
\end{align}
where samples $ s_i $ are drawn according to the environment.
In continuous spaces, the $ \xi $-norm is typically based on integrals
\begin{align}
  \label{eq:msbe_continuous}
  \text{MSBE}(F)
  = \frac12 \Big\| F - (\bellop F) \Big\|_\xi^2
  = \frac12 \int_\Scal \xi(s) \Big( F(s) - (\bellop F)(s) \Big)^2 ds.
\end{align}
Here, $ \xi $ takes the role of a normalised probability density function such that we can make use
of Monte Carlo Integration.
%
It allows us to write integrals as summations over samples
\begin{align}
  \label{eq:msbe_mc_continuous}
  \text{MSBE}(F)
  = \frac12 \int_\Scal \xi(s) \Big( F(s) - (\bellop F)(s) \Big)^2 ds
  \approx \frac{1}{2N} \sum_{i=1}^N \Big( F(s_i) - (\bellop F)(s_i) \Big)^2.
\end{align}
From \cref{eq:msbe_mc_discrete,eq:msbe_mc_continuous} it becomes clear that a realisation of an
optimisation procedure for both types of state spaces results in the same instructions for a
computer.
As the only requirement, one has to ensure that the integrand $ \left( F(s) - (\bellop F)(s)
\right) $ in \cref{eq:msbe_continuous} is square integrable, i.e., an element of $ L^2 $.
This is not a problem for the definition of $ V_\pi $ as the infinite discounted sum in
\cref{eq:def_V_pi}.
Already necessary assumptions such as finite rewards $ \left|r(\cdot)\right|< M $ for some $ M > 0 $
and bounded state spaces guarantee square integrability.
The system dynamics must be chosen to fit implicitly into these conditions.
Still, care must be taken for the selected function space $ \Vcal $.
If using for example neural networks with arbitrary non-linearities as function class, square
integrability might not be ensured.
Furthermore, neural networks are defined for the whole Euclidean space, but the integral covers
only a bounded subspace.
Hence, to avoid issues at the boundary of the state space, the integral should be extended to the
whole Euclidean space instead of $ \Scal $.
However, this stands in conflict with the assumptions required for RL and also can affect in turn
the square integrability of $ V_\pi $.
We leave these theoretical considerations as a topic for future research.

The extension of the MSBE to $ Q $-factors is in both settings straightforward.
One has to add another sum or integral over all actions to the definition of the MSBE and include
the action as second input to $ F $.

Reinforcement Learning with Function Approximation now manifests itself in the optimisation problem
\begin{equation}
  \label{eq:msbe_argmin}
  F_\pi \in \argmin_{F \in \Vcal} \text{MSBE}(F),
\end{equation}
where $ F_\pi $ is an optimal approximation to $ V_\pi $ in terms of minimising the MSBE.
In general we have $ F_\pi \neq V_\pi $ for at least some states $ s \in \Scal $ and only aim to be
close enough to $ V_\pi $.
By the choice of $ \xi $ as weighting for the norm, we also maintain for linear function
approximation architectures the contraction properties of $ \bellop $ when combining it with a
linear projection onto that function space.
The accuracy of the solution $ F_\pi $ as defined in \cref{eq:msbe_argmin} is known to be bounded by
%
\begin{equation}
  \label{eq:msbe_bound}
  \big\| F_\pi - V_\pi \big\|_\xi \leq \frac{1 + \gamma}{1 - \gamma}
  \inf_{F \in \Vcal} \big\| F - V_\pi \big\|_\xi.
\end{equation}
Obviously, the challenge is to find a function space $ \Vcal $ which we can optimise over and which
still contains a good approximation for the desired function $ V_\pi $.

\subsection{Multi Layer Perceptrons}
\label{ssec:notation_mlp}
To approximate value functions, we deploy in this paper the classic feed forward fully connected
neural network, a.k.a. \emph{Multi-Layer Perceptron} (MLP).
We summarise in the following the well-known definition of an MLP such that for the remaining
document a concise notation exists with the goal to avoid any possible source of confusion.

Let us denote by $ L $ the number of layers in the MLP structure, and by $ n_l $ the number of
processing units in the $ l $-th layer with $ l = 1, \ldots, L $.
By $ l = 0 $ we refer to the input layer with $ n_0 $ units.
The value of $ n_0 $ depends on the state space and its type.
For discrete spaces, we have $ n_0 = 1 $ such that a single state can be processed directly as
natural number by the network.
In a continuous setting, we use $ n_0 = K $ units matching the $ K $-dimensional state vectors.
We always restrict the number of nodes in the output layer $ l = L $ to $ n_L = 1 $.

Let $ \sigma \colon \Rds \to \Rds$ be a unit activation function and denote by
$ \dot \sigma \colon \Rds \to \Rds $ its first derivative with respect to the input.
The unit activation function $ \sigma $ and its derivatives act element-wise on non-scalar values.
Depending on the concrete choice for $ \sigma $, the domain and image might have to be changed.
Traditionally, the activation function $ \sigma $ is chosen to be non-constant, bounded, continuous
and monotonically increasing (e.g. the \emph{Sigmoid} function).
More recent popular choices consider unbounded functions like (\emph{Leaky-}) \emph{ReLU},
\emph{SoftPlus} or the \emph{Bent-Identity}\footnote{also abbreviated as \emph{Bent-Id}}.
In this work, we further restrict the choice for the activation function to \emph{smooth},
\emph{unbounded} and \emph{strictly monotonically increasing} functions such as SoftPlus,
Bent-Identity or also the \emph{Identity} function itself, which is used in the last layer.
The latter two are used in this work.

For the $ (l, k) $-th unit in an MLP architecture, i.e., the $ k $-th unit in the $ l $-th layer, we
define the corresponding \emph{unit mapping} $ \unit_{l, k} \colon \unitdomain $ as
\begin{equation}
  \label{eq:unit_map}
  \unitmap,
\end{equation}
where $ \phi_{l-1} \in \Rds^{n_{l-1}} $ denotes the output from layer $ (l-1) $.
The terms $ w_{l, k} \in \Rds^{n_{l-1}} $ and $ b_{l, k} \in \Rds $ are a parameter vector and a
scalar bias associated with the $ (l, k) $-th unit, respectively.
Next, we can define the $ l $-th \emph{layer mapping} by stacking all \emph{unit mappings} of
layer $ l $ as
\begin{align}
  \nonumber
  \layer_l (W_l, b_l, \phi_{l-1})
  \coloneqq&
  \onebythree
  {\Lambda_{l, 1}(w_{l, 1}, b_{l, 1}, \phi_{l-1})}
  {\ldots}
  {\Lambda_{l, n_l}(w_{l, n_l}, b_{l,n_l}, \phi_{l-1})}^\top\\
  \label{eq:layer_map_pre}
  =&~ \layeroperation,
\end{align}
with $ W_l \coloneqq \left[ w_{l, 1}, \ldots, w_{l, n_l}\right] \in \Rds^{n_{l-1} \times n_l} $
being the $ l $-th parameter matrix and
$ b_l \coloneqq \left[b_{l, 1}, \cdots, b_{l, n_l} \right] \in \Rds^{n_{l-1}} $ the $ l $-th bias
vector.
It is convenient to store the bias vector as an additional row of the matrix and extend the layer
input with a constant value of $ 1 $.
Thus, we can write \cref{eq:layer_map_pre} equivalently as
\begin{equation}
  \label{eq:layer_map}
  \layermaphom
\end{equation}
by using a larger parameter matrix $ W_l \in \Rds^{(n_{l-1} + 1) \times n_l} $.
Next, let us define the overall function represented by the MLP.
First, denote by $ \phi_0 \in \Rds^K $ the network input.
Then, the output at the $ l $-th layer is defined recursively as
$ \phi_l \coloneqq \layer_l(W_l,\phi_{l-1}) $.
Note that the last layer in an MLP employs the identity map as activation function and is thus only
an affine mapping.
Finally, by denoting the set of all parameter matrices in the MLP by $ \weightspacehom{n_0}{1} $,
we can compose all layer-wise mappings to define for a set of parameters $ \Wbf \in \Wbfcal $ the
overall \emph{network mapping} as
\begin{equation}
  \label{eq:network_map}
  \netmapfull{n_0}{},
\end{equation}
which contains in total $ \Nnet $ parameters with $ \Nnet $ being computed by
\begin{equation}
  \label{eq:N_net}
  \Nnet = \sum_{l=1}^L (n_{l-1} + 1) \cdot n_{l}.
\end{equation}
With this construction, we can define the set of parametrised functions belonging to an MLP by
writing
\begin{equation}
  \label{eq:mlp_class}
  \MLPfull{n_0}{}.
\end{equation}
With slight abuse of notation, we refer by $ \MLP{n_0, n_1}{n_{L-1}, 1} $ to a concrete function
class by specifying the architecture of the MLP, i.e., the number of processing units in each layer
as well as the input and output dimensions.
Sometimes it is more convenient to describe an MLP by its depth $ d $ and the identical width $ w $
of all hidden layers.
In this case we write $ \MLPwd{n_0}{w}{d}{1} $.
The type of non-linearity is typically fixed and mentioned separately.

\subsection{Optimisation Approaches}
\label{ssec:optimisation_approach}
For solving the optimisation problem in \cref{eq:msbe_argmin}
, i.e., obtaining the approximated value function with smallest MSBE,
two fundamental approaches using gradient information exist:
\begin{itemize}
  \item \emph{Direct Algorithms} \citep{bair:icml95},
   a.k.a. \emph{Semi-Gradient} \citep{sutt:book20};

  \item \emph{Residual Gradient} algorithms \citep{bair:icml95}.
\end{itemize}
Both approaches are descent algorithms.
The difference between them resides in the choice of descent directions.
For Residual Gradient algorithms, the complete gradient is calculated.
In Semi-Gradient algorithms, one ignores the dependence on the parameters through the value
function inside the Bellman Operator.
We refer to \citep{zhang:aamas20} and references therein for a comparison of Semi-Gradient and
Residual Gradient algorithms.

Next, we proceed with the analysis of critical points of \cref{eq:msbe_norm_xi} for discrete and
continuous spaces.
We only investigate Residual Gradient algorithms as there is no insight to be gained from a critical
point analysis, when we use a direction for descending, which is not always pointing in the same
half space as the gradient.

%
\section{A Critical Point Analysis of the MSBE}
\label{sec:theory}
In the following, we present our theoretical investigation of Residual Gradient algorithms by
analysing the critical points of the MSBE.
Thereby, we follow the work in \citep{shen:cvpr18} and translate those insights to the RL domain.
First, we investigate discrete state spaces and derive conditions under which learning the exact
value function works reliably.
Second, we extend our analysis to continuous spaces by changing to a sampling based approximation of
the MSBE and adapt our conditions accordingly.
We characterise the importance of over-parametrisation, give design principles for MLPs and
unveil a connection to other state of the art algorithms.

\subsection{For Discrete State Spaces}
\label{ssec:crit_point_discrete}
In discrete problems, we have two choices to approach a critical point analysis.
Either we seek for an approximation architecture that is exact for each of the $ K $ states or
we are interested in an architecture, which is exact for only $ N \ll K $ sampled but unique states.
While the second case simplifies the analysis and relaxes restrictions as shown later for
continuous spaces, this also means that we have to address generalisation to states outside of the
sampled ones.
As the sampling case with discrete states would match almost completely our approach for continuous
state spaces, we present here only the exact learning assumption.
This implies that we are interested in conditions for an MLP that would allow for a perfect solution
to any state.

We consider MLPs of the form $ \MLP{1, n_1}{n_{L-1}, 1} $, i.e., fully connected feed forward
networks with a one dimensional input and output and arbitrary depth or width.
With discrete and finite spaces we can evaluate an MLP $ f \in \Fcal$ for every available state
$ s \in \Scal $ and collect the evaluations of $ f $ as vector in $ \Rds^K $, where $ K $ is the
number of states.
As the simplest approach, we encode the discrete states with natural numbers for the single input
unit and, thus, can treat $ F(\Wbf) \coloneqq [ f(\Wbf, 1) \ldots f(\Wbf, K)]^\top \in \Rds^K $ as
the approximated value function for all states.
Next, we define the \emph{Bellman Residual} vector for a policy $ \pi $ in matrix form as
\begin{equation}
  \label{eq:TD-error-vec}
  \Delta_\pi(\Wbf)
  = F(\Wbf) - \bellop F(\Wbf)
  = F(\Wbf) - P_\pi \Big( R_\pi + \gamma F(\Wbf) \Big),
\end{equation}
where the term $ R_\pi $ contains the collection of all one-step-rewards suitable for the matrix
form of the Bellman Operator and $ P_\pi $ is the transition probability matrix under policy
$ \pi $.
We use a capital $ \Delta $ instead of $ \delta $ to emphasize that we consider all transitions in
here instead of just a single one as in \cref{eq:TD-error}.
Using a diagonal matrix $ \Xi \coloneqq \diag(\xi_1, \ldots, \xi_K) $ consisting of the steady state
distributions $ \xi_i $ for all states, we can rewrite the norm in \cref{eq:msbe_discrete} as the
\emph{Neural Mean Squared Bellman Error} (NMSBE) function
\begin{equation}
  \label{eq:nmsbe}
  \Jcal(\Wbf) \coloneqq \frac12 \Delta_\pi(\Wbf)^\top \Xi \Delta_\pi(\Wbf).
\end{equation}
%
%
It is important to notice that the NMSBE function is generally \emph{non-convex} in $ \Wbf $, and
even worse, it is also \emph{non-coercive} \citep{guel:book10}.
Namely, for $ \Wbf \to \infty $ we do not necessarily have $ \Jcal(\Wbf) \to \infty $.
Thus, the existence and attainability of global minima of $ \Jcal(\Wbf) $ are not guaranteed in
general.
Nevertheless, when appropriate non-linear activation functions are employed in hidden layers, exact
learning of finite samples can be achieved with sufficiently large architectures \citep{shah:tnn99}.
Thus, for our analysis, it is safe to assume that there exists an MLP able to approximate the value
function $ V_\pi $.
\begin{assumption}[Existence of exact approximator]
  \label{ass:existence_exact_approximator}
  Let $ V_\pi \colon \Scal \to \Rds $ be the exact value function under policy $ \pi $.
  There exists at least one MLP architecture $ \Fcal $ as defined
  in \cref{eq:mlp_class} together with a set of parameters $ \Wbf^* \in \Wbfcal $ such that
  the output of $ F $ and $ V_\pi $ coincide, i.e., we have
  \begin{equation*}
    f(\Wbf^*, s) = V_\pi(s) \quad \forall s \in \Scal.
  \end{equation*}
\end{assumption}
To conduct a critical point analysis of the NMSBE function we first compute the derivatives.
By the linearity of the Bellman Operator $ \bellop $, we obtain the directional derivative of
$ \Jcal $ at the point $ \Wbf \in \Wbfcal $ in direction $ \Hbf \in \Wbfcal $ as
\begin{equation}
  \label{eq:diff_map_J_discrete}
  \D \Jcal(\Wbf)[\Hbf] = \Delta_\pi(\Wbf)^\top \Xi ( I_K - \gamma P_\pi) \D F(\Wbf)[\Hbf],
\end{equation}
with $ \D F(\Wbf)[\Hbf] $ being the differential map of the MLP and $ I_K $ the $ K \times K $
identity matrix.
As the function $ F(\Wbf) $ is simply a superposed function evaluated at each state, we just need
to compute the directional derivative of the MLP evaluated at a specific state $ s $, i.e.,
$ \D f(\Wbf, s)[\Hbf] $.
Furthermore, the directional derivative of $ f(\Wbf, s) $ is a linear operator, hence the evaluation
of the directional derivative of $ f $ for all states can be expressed as a matrix vector
multiplication and results in
\begin{equation}
  \label{eq:diff_map_f}
  \D F(\Wbf)[\Hbf] =
  \underbrace{
  \left.
  \Big[ \flatten(\nabla_\Wbf f(\Wbf, 1)) \ldots \flatten(\nabla_\Wbf f(\Wbf, K))\Big]
   \right.^\top
  }_{\eqqcolon G(\Wbf) \in \Rds^{K \times \Nnet}} \flatten(\Hbf),
\end{equation}
where $ \flatten( \nabla_\Wbf f(\Wbf, s) ) \in \Rds^{\Nnet} $ is the gradient of $ f $ with respect
to the parameters evaluated at $ \Wbf $ for state $ s $ under the Euclidean norm.
The operation $ \flatten(\cdot) $ transforms a matrix into a vector by stacking its columns.
It acts on collections of matrices by concatenating the results of each individual vectorisation.
The matrix $ G(\Wbf) $ takes the role of the Jacobian for the evaluation of all states $ F(\Wbf) $.
Now, we can characterise critical points of the NMSBE function $ \Jcal $ from \cref{eq:nmsbe} by
setting its gradient to zero, i.e., $ \nabla_{\Wbf} \Jcal(\Wbf) = 0 $.
Combining the results from \cref{eq:diff_map_J_discrete,eq:diff_map_f}
together with \emph{Riesz' Representation Theorem}
yields the critical point condition
\begin{equation}
  \label{eq:crit_point_cond_disc}
  \nabla_\Wbf \Jcal(\Wbf) \coloneqq
  G(\Wbf)^\top \left( I_K - \gamma P_\pi \right)^\top \Xi \Delta_\pi(\Wbf)
  \stackrel!= 0,
\end{equation}
%
which is the counterpart\footnote{In this paper we used $ G $ in place of $ P $ to avoid confusion
with the transition probability matrix} to \namecref{eq:crit_point_cond_disc} (19) in
\citep{shen:cvpr18} for the Reinforcement Learning setting with exact learning.
We derive the following proposition.
\begin{proposition}[Suboptimal local minima free condition]
  \label{prop:local_free_strong}
  Let an MLP architecture $ \Fcal $ satisfy \Cref{ass:existence_exact_approximator}.
  If the rank of the matrix $ G(\Wbf) $ as constructed in \cref{eq:diff_map_f} is equal to $ K $
  for all $ \Wbf \in \Wbfcal $, then any extremum $ \Wbf^* \in \Wbfcal $ realises the true value
  function $ V_\pi $, i.e., $ f(\Wbf^*, s) = V_\pi(s)~\forall s \in \Scal $.
  Furthermore, the NMSBE function $ \Jcal $ is free of suboptimal local minima.
\end{proposition}
\begin{proof}
  Since the underlying state transitions under policy $ \pi $ are required to be Markovian and
  ergodic, both terms $ \Xi $ and $ (I_{K} - \gamma P_\pi) $ have full rank.
  Consequently, the expression $ \Xi (I_{K} - \gamma P_\pi) G(\Wbf^*)^\top $ also has rank $ K $, if
  we claim that $ G(\Wbf^*) $ has full rank.
  Hence, there is only the trivial solution left for the linear system in
  \cref{eq:crit_point_cond_disc}, meaning that the Bellman Residual $ \Delta_\pi(\Wbf^*) $ must be
  exactly zero for all states.
  Since the Bellman Residual is only zero for the unique fixed point $ V_\pi $ of the operator
  $ \bellop $, \Cref{ass:existence_exact_approximator} implies that $ \Wbf^* $ corresponds to the
  true value function.
  Furthermore, the Bellman Residual appears as factor in the NMSBE.
  Hence, at any critical point the error vanishes and there are no suboptimal local minima.
\end{proof}
To make use of \Cref{prop:local_free_strong}, one needs to investigate, under what conditions the
matrix $ G(\Wbf) $ has full column rank.
The immediate risk of being exactly zero can be eliminated by choosing proper activation functions
without zero derivatives for finite inputs.
To see this we have to look at the structure of $ G(\Wbf) $, which takes the form
\begin{equation}
  \label{eq:def_G_W}
  G(\Wbf) =
  \threebythree
  {\Psi_{1}^\top \left( I_{n_1} \kron {\phi^{(1)}_{0}}^\top \right)}
  {\ldots}
  {\Psi_{L}^\top \left( I_{n_L} \kron {\phi^{(1)}_{L-1}}^\top \right)}
  {\vdots}{\ddots}{\vdots}
  {\Psi_{1}^\top \left( I_{n_1} \kron {\phi^{(K)}_{0}}^\top \right)}
  {\ldots}
  {\Psi_{L}^\top \left( I_{n_L} \kron {\phi^{(K)}_{L-1}}^\top \right)},
\end{equation}
where the matrices $ \Psi_l \in \Rds^{n_l \times n_L} $ and Kronecker products result from the
layer wise definition of an MLP.
The $ \Psi_l $ are products of diagonal matrices containing the derivative of $ \sigma $ and the
parameter matrices of various layers, hence a proper choice for $ \sigma $ prevents $ G(\Wbf) $ from
becoming zero.
A detailed construction is available in the appendix.

Maintaining the rank of $ G(\Wbf) $ means to choose MLPs in such a way that one minimises the
risk for rows of the matrix to lie in a shared subspace.
As in the original multidimensional regression setting, the overall block structure of $ G(\Wbf) $
remains for RL applications.
However, we have in the RL setting the advantage that each block in $ G(\Wbf) $ reduces to a
single row, because the output layers are always scalar, i.e., $ n_L = 1 $.
For the same amount of sampled inputs, there are less possibilities to have linear dependent rows.
Although there are no theoretical guarantees, most matrices have full rank in practical
applications, where noise or sampling is present.

Unfortunately, the requirement of using over-parametrised MLPs, i.e.,
\begin{equation}
  \label{eq:overparam_discrete}
  \Nnet \geq K,
\end{equation}
prevents a direct application of the theory.
If we need as many adjustable parameters in an MLP as there are unique states, we could use a
tabular representation in the first place and avoid dealing with non-linear function approximation.
Note that the strong condition in \cref{eq:overparam_discrete} stems from the assumption of being
exact for all states.
Typically, discrete states are not processed directly by an MLP but passed through an encoding
step such as assigning random features.
This leads to a potentially weaker lower bound $ \tilde K < K $, since several states could receive
an identical feature vector.

If the NMSBE is changed to be an approximation based on sampling, as done for continuous spaces in
\cref{ssec:crit_point_continuous}, we can reduce the number of MLP parameters from $ K $ down to
$ N \ll K $ samples.
But as an immediate consequence, generalisation becomes an issue as a result of the MLP being only
accurate by design for finitely many points in the state space.
Investigating the predictive capabilities for the remaining states of a discrete space, which not
necessarily has a proper definition for a metric, is beyond the analysis presented in this work.

In order to obtain an efficient and effective algorithm one can employ Newton-type optimisation
procedures.
Furthermore, to ease the work involved in the Hessian, we aim at \emph{Approximated Newton's} (AN)
algorithms.
A closer look reveals that the differential map as shown in \cref{eq:diff_map_J_discrete} is a
candidate for the \emph{Gauss Newton} (GN) approximation as in the non-linear regression setting.
Indeed, for the second directional derivative of $ \Jcal $ at $ \Wbf $ with two directions
$ \Hbf_1, \Hbf_2 \in \Wbfcal $ 
we have
\begin{align}
  \nonumber
  \D^2 \Jcal(\Wbf)[\Hbf_1,\Hbf_2] =& \;
  \Delta_\pi(\Wbf)^\top \Xi \left( I_K - \gamma P_\pi \right) \D^2 F(\Wbf)[\Hbf_1,\Hbf_2] \\
  \label{eq:true_hessian_J}
  &+ \D F(\Wbf)[\Hbf_1]^\top \left( I_K - \gamma P_\pi \right)^\top \Xi \left( I_K - \gamma P_\pi
  \right) \D F(\Wbf)[\Hbf_2],
\end{align}
where we see that the first summand from the right hand side vanishes at any critical point
$ \Wbf^* \in \Wbfcal $ according to \Cref{prop:local_free_strong}.
Thus, the evaluation of the Hessian of the NMSBE function at $ \Wbf^* $ is given by
\begin{equation}
  \label{eq:hessian_J}
  \D^2 \Jcal(\Wbf^*)[\Hbf_1,\Hbf_2] =
  \flatten(\Hbf_1)^\top
  \underbrace{
  G(\Wbf^*)^\top \left( I_K - \gamma P_\pi \right)^\top
  \Xi
  \left( I_K - \gamma P_\pi \right) G(\Wbf^*)
  }_{\Hess_\Wbf \Jcal(\Wbf^*) \in \Rds^{\Nnet \times \Nnet}}
  \flatten(\Hbf_2).
\end{equation}
This corresponds to the Gauss Newton approximation for non-linear least squares regression, i.e.,
defining the Hessian as product of the Jacobian and its transpose.
Our characterisation of critical points reveals this possibility for approximation as a side
benefit.
Using naively the product of the model's Jacobians $ G(\Wbf^*) $ as approximation would ignore the
additional structure coming from the Bellman Operator.

To ensure proper behaviour for a GN algorithm, we need to characterise the Hessian
$ \Hess_{\Wbf} \Jcal(\Wbf^*) $ of the NMSBE at all critical points.
Its quadratic form leads to the following result for MLPs.
\begin{proposition}[Properties of the approximated Hessian]
  \label{prop:hessian_rank}
  The Hessian of the NMSBE function $ \Jcal $ at any critical point $ \Wbf^*$ is always positive
  semi-definite.
  Furthermore, its rank is bounded from above by
  \begin{equation*}
    \rank(\Hess_{\Wbf} \Jcal(\Wbf^*)) \leq K,
  \end{equation*}
  if the MLP satisfies \cref{eq:overparam_discrete}.
\end{proposition}
\begin{proof}
  Positive semi-definiteness of $ \Hess_{\Wbf} \Jcal(\Wbf^*) $ follows from its symmetric
  definition.
  As before $ \Xi $ and $ (I_{K} - \gamma P_\pi) $ have full rank.
  The rank of $ G(\Wbf^*) $ is at most $ K $.
  For $ \Hess_{\Wbf} \Jcal(\Wbf^*) $ being the product of these matrices we get the upper bound on
  its rank.
\end{proof}
It is interesting to see that the rank condition from \cref{eq:overparam_discrete} also allows the
Hessian to become positive definite if the matrix $ G(\Wbf) $ has full column rank.
This has significant consequences for the optimisation problem.
A positive definite Hessian at all critical points means that they are all local minima, thereby
supporting further \Cref{prop:local_free_strong}.
There are no saddle points or maxima, where a gradient based optimisation strategy could get stuck.
Thus, over-parametrisation of MLPs is not only important for \Cref{prop:local_free_strong} but also
necessary from the algorithmic perspective.
We confirm the proposed approximation for the Hessian in discrete state spaces with exact learning
numerically in \cref{ssec:quad_conv_demo}.

\subsection{For Continuous State Spaces}
\label{ssec:crit_point_continuous}
In high dimensional continuous state spaces, say $ K > 6 $, an exact representation of the value
function based on a fine grained partitioning of $ \Scal $ is typically impossible.
This is due to the \emph{Curse of Dimensionality} and we are forced to work directly with the
continuous space, which causes MLPs to be of the form $ \MLP{K}{1} $ to accept $ K $ dimensional
state vectors as input.

Since there cannot exist a transition probability matrix in continuous spaces and its corresponding
discrete steady state distribution $ \Xi $, the NMSBE as shown in \cref{eq:nmsbe} is not available
here.
We must work with a finite number of samples $ N \in \Nds $ to approximate the loss as done in
\cref{eq:msbe_mc_continuous}.

Another limitation for Residual Gradient algorithms is the so-called \emph{Doubling Sampling Issue}.
Due to the expectation inside the Bellman Operator for every single sample $ s_i \in \Scal $ many
possible successors $ s_i' $ are necessary to approximate this expectation empirically
\citep{bair:icml95}.
The Double Sampling Issue can be bypassed if an accurate model
containing a description of stochastic transitions
is available or if one has access to a simulator, where the state can be set freely to collect its
successors.
However, if one wishes to learn in a model free manner or with rather limited and less powerful
simulations, collecting successor samples becomes problematic.
In a recent work \citep{saleh:nips19}, the authors rediscovered the application of Residual Gradient
algorithms in deterministic environments.
They are motivated by their observation that many environments and RL testbeds are deterministic or
posses only a small amount of noise.
Thus, for our analysis, we follow the same strategy and
restrict ourself to deterministic MDPs and analyse the algorithm in its purest form.
The one-step TD-error from \cref{eq:TD-error} now simplifies for the
$ i $-th sample to
\begin{equation}
  \label{eq:TD-error-det}
  \delta(s_i, s_i') \coloneqq V(s_i) - r(s_i, \pi(s_i), s_i') - \gamma V(s_i'),
\end{equation}
where $ s_i' $ is the successor of $ s_i $ when executing $ \pi(s_i) $.
As before, we collect the evaluation of the MLP for all $ N $ sampled states $ s_i $ as vector and
denote it by $ F(\Wbf) \coloneqq [ f(\Wbf, s_1) \ldots f(\Wbf, s_N) ]^\top \in \Rds^N $.
Next, we rewrite the loss of \cref{eq:msbe_mc_continuous} accordingly and obtain the sampled NMSBE
for continuous state spaces as
\begin{align}
  \nonumber
  \widetilde \Jcal(\Wbf)
  &\coloneqq \frac{1}{2N} \sum _{i=1}^N \Big(
  \underbrace{
  f(\Wbf, s_i) - r(s_i, \pi(s_i), s_i') - \gamma f(\Wbf, s_i')
  }_{\delta(s_i, s_i')}
  \Big)^2 \\
  \label{eq:nmsbe_continuous}
  &= \frac{1}{2N}
  \onebythree{\delta(s_1, s_1')}{\ldots}{\delta(s_N, s_N')} \cdot
  \threebyone{\delta(s_1, s_1')}{\vdots}{\delta(s_N, s_N')}
  = \frac{1}{2N} \Delta_\pi(\Wbf)^\top \Delta_\pi(\Wbf),
\end{align}
where $ \Delta_{\pi}(\Wbf) \in \Rds^N $ now takes the form
\begin{equation}
  \label{eq:TD-error-vec-continuous}
  \Delta_\pi(\Wbf) = F(\Wbf) - R_\pi - \gamma F'(\Wbf)
\end{equation}
if we denote by $ F' $ the evaluation of $ f $ for all successor states.
For this loss, similar lines of thought apply as in the discrete setting, but analogously to
\citep{shen:cvpr18}, we now consider a finite set of sample states at which the corresponding
value function is approximated exactly.
We formulate this situation for Residual Gradient algorithms precisely in the next definition.
\begin{definition}[Finite exact approximator]
  \label{def:finite_exact_approximator}
  Let $ V_\pi \colon \Scal \to \Rds $ be the value function under policy $ \pi $.
  Given $ N $ sample states $ s_i \in \Scal $ we call an MLP $ f \in \Fcal $, which satisfies
  \begin{equation*}
    f(\Wbf, s_i) = V_\pi(s_i) \quad \forall i=1, \ldots, N
  \end{equation*}
  for some parameters $ \Wbf \in \Wbfcal $, a finite exact approximator of $ V_\pi $ based on the
  $ N $ sample states.
\end{definition}
As in the discrete setting, we can choose sufficiently rich MLP architectures $ \Fcal $ and, thus,
assume also in the continuous setting the existence of such an approximator.
\begin{assumption}[Existence of finite exact approximator]
  \label{ass:existence_finite_exact_approximator}
  Let $ V_\pi \colon \Scal \to \Rds $ be the value function of policy $ \pi $.
  Given $ N $ unique samples $ s_i \in \Scal $, there exists at least one MLP architecture
  $ \Fcal $ as defined in \cref{eq:mlp_class} together with a set of parameters
  $ \Wbf^* \in \Wbfcal $ such that the MLP $ f(\Wbf^*, \cdot) \in \Fcal $ is a finite exact
  approximator of $ V_\pi $ according to \Cref{def:finite_exact_approximator}.
\end{assumption}
A key challenge of \Cref{def:finite_exact_approximator} and
\Cref{ass:existence_finite_exact_approximator} resides in the sampling based approximation of the
NMSBE.
With only finitely many samples, it can happen that the connection between states and their
successors is lost in some parts of the state space.
Whether this happens solely depends on the MDP and the distance of samples.
Hence, the NMSBE might possess undesired degrees of freedom and \cref{eq:TD-error-det} no longer
applies to all sample states.
Unfortunately, \Cref{ass:existence_finite_exact_approximator} ensures, that we can always find an
MLP $ f $ with parameters $ \Wbf^* $ to make $ \Delta_\pi(\Wbf) $ exactly zero.
However, in such a scenario, this MLP will not be a finite exact approximator according to
\Cref{def:finite_exact_approximator}.
\begin{proposition}[The NMSBE includes bad solutions]
  \label{prop:bad_solutions_NMSBE}
  The NMSBE as shown in \cref{eq:nmsbe_continuous} contains minima $ \Wbf^* \in \Wbfcal $, for which
  the MLP $ f $ is not a finite exact approximator according to
  \cref{def:finite_exact_approximator}.
\end{proposition}
\begin{proof}
  Denote by $ \Wbfcal^* \coloneqq  \left\{\left. \Wbf \in \Wbfcal ~\right| \Delta_{\pi}(\Wbf) = 0
  \right\} $ the set of all MLP with minimal NMSBE.
  It coincides with critical points as we see later.
  By $ \Wbfcal' \coloneqq  \left\{ \left. \Wbf \in \Wbfcal ~\right| f(\Wbf, s_i) = V_\pi(s_i)
  ~\forall i \right\} $ we denote the set of all finite exact approximators.
  Both sets are non-empty due to \Cref{ass:existence_exact_approximator}.
  In general we have $ \Wbfcal' \subset \Wbfcal^* $ such that $ \Wbfcal^* \setminus \Wbfcal' \neq
  \{\} $, because if for a single sample state the connection to its successors is not present, then
  the MLP can approximate any value for that state without affecting the NMSBE.
  Thus, there exist critical points of the NMSBE for which at some states the MLP is arbitrarily
  far away from $ V_\pi $.
\end{proof}
Whether the mismatch of both sets becomes problematic, depends on generalisation capabilities of
MLPs.
Thus, these capabilities are crucial and we investigate them empirically as part of the experiments.

Our critical point analysis follows the same steps as in \cref{ssec:crit_point_discrete}.
First, we need the differential map of \cref{eq:nmsbe_continuous}.
It takes the form
\begin{align}
  \nonumber
  \D \widetilde \Jcal(\Wbf)[\Hbf]
  &= \frac1N \Delta_\pi(\Wbf)^\top \Big( \D F(\Wbf)[\Hbf] - \gamma \D F'(\Wbf)[\Hbf] \Big)\\
  \label{eq:diff_map_J_continuous}
  &= \frac1N \Delta_\pi(\Wbf)^\top \Big( G(\Wbf) - \gamma G'(\Wbf) \Big) \flatten(\Hbf),
\end{align}
where the definition of $ G(\Wbf) \in \Rds^{N \times \Nnet} $ applies as in the discrete setting.
Detailed steps for its derivation are provided in the appendix.
The term $ G' $ results from using $ s_i' $ as input.
Using this map, critical points can be characterised by setting the gradient to zero
\begin{equation}
  \label{eq:crit_point_cond_cont}
  \nabla_\Wbf \widetilde \Jcal(\Wbf) \coloneqq
  \frac1N
  \Big(
  \underbrace{
  G(\Wbf) - \gamma G'(\Wbf)
  }_{\eqqcolon \widetilde G(\Wbf) \in \Rds^{N \times \Nnet}}
  \Big)^\top
  \Delta_\pi(\Wbf)
  \stackrel!= 0.
\end{equation}
Apparently, the critical point condition for the continuous setting based on $ N $ unique samples in
\cref{eq:crit_point_cond_cont} takes a similar form as that of the discrete setting in
\cref{eq:crit_point_cond_disc}.
But since we no longer investigate the exact learning scenario, we have to reformulate
\Cref{prop:local_free_strong} and obtain a slightly different version.
\begin{proposition}[Suboptimal local minima free condition]
  \label{prop:local_free_strong_continuous}
  Let an MLP architecture $ f \in \Fcal $ satisfy \Cref{ass:existence_finite_exact_approximator}.
  If the rank of the matrix $ \widetilde G(\Wbf) $ as defined in \cref{eq:crit_point_cond_cont} is
  equal to $ N $ for all $ \Wbf \in \Wbfcal $, then any extremum $ \Wbf^* \in \Wbfcal $ achieves
  zero NMSBE.
  Furthermore, the NMSBE function $ \widetilde \Jcal $ of \cref{eq:nmsbe_continuous} is free of
  suboptimal local minima.
\end{proposition}
\begin{proof}
  As before, \cref{eq:crit_point_cond_cont} defines a linear equation system in the sampled Bellman
  Residual vector $ \Delta_\pi $.
  If we claim that $ \widetilde G(\Wbf) $ has full rank, there is only the trivial solution left for
  $ \Delta_\pi(\Wbf) $.
  By \Cref{ass:existence_finite_exact_approximator} we know that such an MLP exists.
  Furthermore, the Bellman Residual appears as factor in the sampled NMSBE.
  Hence, at any critical point the error vanishes completely and there are no suboptimal local
  minima.
\end{proof}
We now address the requirements for $ \widetilde G(\Wbf) $ since this matrix forms the backbone of
\Cref{prop:local_free_strong_continuous}.
As the first requirement, $ \widetilde G(\Wbf) $ has to be non-zero to define a proper equation
system and enforce a trivial solution in terms of $ \Delta_\pi(\Wbf) $ in
\cref{eq:crit_point_cond_cont}.
For both differential maps $ G(\Wbf) $ and $ G'(\Wbf) $ the design principles apply individually.
Hence it is unlikely in practice that for $ N $ unique sample states simultaneously both matrices
vanish elementwise on their own.
More troublesome is the distance between a state $ s $ and its successor $ s' $.
If $ \left\|s - s' \right\| \to 0 $, which happens for example whenever $ s $ is getting close to a
fixed point of the system dynamics, then the discount factor $ \gamma \in (0, 1) $ prevents a
perfect cancellation of $ G(\Wbf) $ with $ G'(\Wbf) $.
Aside from those fixed points in the state space, it is again unlikely to observe perfect
cancellation of these two matrices in practice.
As the second requirement, the rank of $ \widetilde G(\Wbf) $ is important.
Obviously, it is more difficult to make concise statements compared to discrete and exact learning
setting.
For the rank of the sum of two matrices, the only known inequality is
\begin{equation*}
  \rank \left( A + B \right) < \rank \left( A \right) + \rank \left( B \right),
\end{equation*}
which implies that we still have to increase the rank of $ G(\Wbf) $ and $ G'(\Wbf) $ individually
to push the upper bound for the rank of $ \widetilde G(\Wbf) $ high enough to allow for full rank.
This leads us to similar design principles for the MLP as in the discrete setting.
Hence, the more complex $ \widetilde G(\Wbf) $ still complies to considerations of the discrete
setting regarding linear dependent rows.
Of course these properties and requirements come without any guarantees, i.e., we expect that
carefully constructed examples (yet almost pathological) exist, where a drop of the rank of
$ \widetilde G(\Wbf) $ happens.
But in practice, where numerical errors and sampled quantities are present, we do not consider this
to become a problem.
When taking a closer look at $ \widetilde G(\Wbf) $ itself, we find that the overall block structure
of $ G(\Wbf) $ as shown in \cref{eq:def_G_W} remains.
We have
\begin{equation}
  \label{eq:def_tilde_G_W}
  \widetilde G(\Wbf) \!=\!
  \threebythree
  {\!\!\!
   \Psi_{1} ^\top \!\left( I_{n_1} \!\kron {\phi _{0}}^{(1)} \!\right)^\top \!\! - \gamma
   \Psi_{1}'^\top \!\left( I_{n_1} \!\kron {\phi'_{0}}^{(1)} \!\right)^\top}
  {\!\!\!\!\!\ldots\!\!\!\!\!}
  {\Psi_{L} ^\top \!\left( I_{n_L} \!\kron {\phi _{L-1}}^{(1)} \!\right)^\top \!\! - \gamma
   \Psi_{L}'^\top \!\left( I_{n_L} \!\kron {\phi'_{L-1}}^{(1)} \!\right)^\top
   \!\!\!}
  {\vdots}{\!\!\!\!\!\ddots\!\!\!\!\!}{\vdots}
  {\!\!\!
   \Psi_{1} ^\top \!\left( I_{n_1} \!\kron {\phi _{0}}^{(N)} \!\right)^\top \!\! - \gamma
   \Psi_{1}'^\top \!\left( I_{n_1} \!\kron {\phi'_{0}}^{(N)} \!\right)^\top}
  {\!\!\!\!\!\ldots\!\!\!\!\!}
  {\Psi_{L} ^\top \!\left( I_{n_L} \!\kron {\phi _{L-1}}^{(N)} \!\right)^\top \!\! - \gamma
   \Psi_{L}'^\top \!\left( I_{n_L} \!\kron {\phi'_{L-1}}^{(N)} \!\right)^\top
   \!\!\!}.
\end{equation}
The vectors $ \phi_l \in \Rds^{n_l} $ result from the evaluation of all layers for state $ s $ and
$ \phi'_l $ from using $ s' $.
Similarly, the matrices $ \Psi $ and $ \Psi' $ use $ s $ and $ s' $ for their computation.
We provide slightly more detailed construction steps in the appendix.
Unfortunately, we see from \cref{eq:def_tilde_G_W} that no additional statements are possible.

Various authors report a slow convergence of RG algorithms due to the similarity of $ G(\Wbf) $
and $ G'(\Wbf) $, e.g. \citep{zhang:aamas20, bair:icml95}.
We can identify two remedies here.
The first is already visible in \cref{eq:crit_point_cond_cont} or \cref{eq:def_tilde_G_W}, where
the contribution of the successor $ G'(\Wbf) $ comes with the prefactor $ \gamma $.
If using $ n $-step returns, as done for example in \citep{mnih:pmlr16}, the discount factor would
come with higher powers, thus, more efficiently taking away the cancelling effect of $ G'(\Wbf) $
onto $ G(\Wbf) $ if the successor stays after several steps still close to its original state.
For large enough lookahead, $ \gamma^n $ would become small enough to allow for ignoring
$ G'(\Wbf) $ altogether, which also helps with the desired full rank of $ \widetilde G(\Wbf) $.
Also, Semi-Gradient issues such as the limited applicability of classic optimisation methods seem
to be avoidable, because for long enough lookahead the omitted dependence of derivatives with
respect to MLP parameters in the TD target vanishes naturally.
Furthermore, it is interesting to see that the extension of $ n $-step returns to a full
TD($ \lambda $) algorithm is a core component in \emph{Proximal Policy Optimisation}
\citep{schulman:corr17} or \emph{Generalized Advantage Estimation} \citep{schulman:iclr16}.
Using TD($ \lambda $), one obtains per sampled state more information without increasing their
amount.
Hence, the required number of parameters to obtain a full rank for $ G(\Wbf) $ does not increase.
The second remedy is related to vanishing gradients of $ \widetilde \Jcal $ if $ s $ and $ s' $ are
close by.
Here, the natural solution is to use second-order gradient descent, which takes the curvature into
account to define a descent direction.
Hence, we propose to employ again a \emph{Gauss Newton Residual Gradient Algorithm}.
%
Other approaches to overcome curvature issues such as momentum based descent algorithms are too
reliant on the dynamical runtime behaviour as well as on initialisation.
They are prone to excessive hyper parameter tuning and frequent restarts.
In the worst case, they complicate reproducibility, which is also a reason, why we have decided to
use second-order optimisation.

As before, to see the possibility for a GN approximation of the Hessian, we first write down the
second-order differential map of $ \widetilde \Jcal $ at point $ \Wbf \in \Wbfcal $ for two
directions $ \Hbf_1, \Hbf_2 \in \Wbfcal $
\begin{align}
  \nonumber
  \D^2 \widetilde \Jcal(\Wbf)[\Hbf_1,\Hbf_2] =& \;
  \frac1N \Delta_\pi(\Wbf)^\top \Big( \D^2 F(\Wbf)[\Hbf_1,\Hbf_2] - \gamma \D^2
  F'(\Wbf)[\Hbf_1,\Hbf_2] \Big) \\
  \label{eq:true_hessian_J_continuous}
  &+ \frac1N \D \Delta_\pi(\Wbf)[\Hbf_1]^\top \Big( \D F(\Wbf)[\Hbf_2] - \gamma
  \D F'(\Wbf)[\Hbf_2] \Big).
\end{align}
Since the first summand contains the Bellman Residual as factor it vanishes at any critical point
$ \Wbf^* $ of $ \widetilde \Jcal $ due to the assumption of exact learning at sample states.
This removes the contribution of second-order derivatives of the MLP and allows us to simplify the
Hessian to
\begin{equation}
  \label{eq:hessian_J_continuous}
  \D^2 \widetilde \Jcal(\Wbf^*)[\Hbf_1,\Hbf_2] = \frac1N
  \flatten(\Hbf_1)^\top
  \underbrace{
  \Big(G(\Wbf^*) - \gamma G'(\Wbf^*)\Big)^\top
  \Big(G(\Wbf^*) - \gamma G'(\Wbf^*)\Big)
  }_{\Hess_\Wbf \widetilde \Jcal(\Wbf^*) \in \Rds^{\Nnet \times \Nnet}}
  \flatten(\Hbf_2).
\end{equation}
It becomes clear that \Cref{prop:hessian_rank} applies almost unchanged.
\begin{proposition}[Properties of the approximated Hessian]
  \label{prop:hessian_rank_continuous}
  The Hessian of the NMSBE function $ \widetilde \Jcal $ from \cref{eq:nmsbe_continuous} is at any
  critical point $ \Wbf^* $ always positive semi-definite.
  Furthermore, its rank is bounded from above by
  \begin{equation*}
    \rank(\Hess_{\Wbf} \widetilde \Jcal(\Wbf^*)) \leq N.
  \end{equation*}
\end{proposition}
\begin{proof}
  Positive semi-definiteness follows from the symmetric definition.
  The rank of the matrix $ \left(G(\Wbf^*) - \gamma G'(\Wbf^*) \right) $ is at most $ N $.
  For $ \Hess_\Wbf \widetilde \Jcal(\Wbf^*) $ being the product of these we get the upper bound on
  its rank.
\end{proof}
The question, whether $ \Hess_\Wbf \widetilde \Jcal(\Wbf^*) $ is positive definite or only
semi-definite, is the same as in the discrete setting.
We just have to use $ \widetilde G(\Wbf^*) $ in place of $ G(\Wbf^*) $ alone and still arrive at
\begin{equation}
  \label{eq:overparam_continuous}
  \Nnet \geq N
\end{equation}
to allow the rank to become full.
We want to emphasize that this requirement for the Hessian is far less restricting than in the
discrete setting with exact learning.
Also, computational concerns regarding a second-order optimisation method are no longer that severe
with nowadays hardware capabilities.
MLPs can become large enough for RL applications while still allowing for working with Hessians in
a reasonable time.
We further address practical concerns in \cref{sec:algorithm} and as part of the experiments.

In summary, we have two kind of approaches, namely exact learning and sampling based approximation.
For discrete state spaces we have a sound and exact algorithm with verifiable local quadratic
convergence as shown later.
Furthermore, this algorithm can be converted to sampling based loss definition at any time without
extra effort.
For continuous state spaces only a sampling based algorithm is realisable.
As we have shown, it possesses a matching behaviour with only slightly altered propositions.
Thus, the only major difference is the formulation of the loss.
By using sampling we allow for broader applications without sacrificing theoretical properties.
The last uncertainty left is how many sample states are required, i.e., how to select the size of
$ N $ for a certain MDP.
We treat this problem concerning sampling complexity as future research.

%
\section{A Gauss Newton Residual Gradient Algorithm}
\label{sec:algorithm}
In this section, we provide details regarding our concrete algorithm and make use of the results
from our analysis.
The focus lies on aspects that are relevant on their own, i.e., decoupled from any experiment, but
fit no longer into the previous section as they are only due to the implementation in a computer
system.
We give pseudo code as it is used in the experiments and demonstrate numerically theoretical
properties of a Gauss Newton algorithm using the discrete exact learning setting.

\subsection{Second-Order Algorithms}
\label{ssec:practical_details}
A frequent concern regarding second-order algorithms is the computational effort involved in models
with many parameters.
Typical architectures of MLPs used in RL applications contain around two hidden layers with
approximately 50 units in each layer.
For such MLPs, $ \Nnet $ falls in the range of $ 2000 $ parameters, for which second-order
optimisation is manageable with reasonable time and storage as we demonstrate with our experiments.

More severe is the root seeking behaviour of a Newton's method.
By using second-order information in a gradient descent optimisation procedure, we aim directly at
any root in the gradient vector field of the NMSBE.
Thus, the optimiser is also attracted by (local) maxima or even saddle points.
Fortunately, as we have already elaborated after
\Cref{prop:hessian_rank,prop:hessian_rank_continuous}, these types of extrema do not exist if our
assumptions are satisfied.
On top of that, only at critical points of $ \widetilde \Jcal(\Wbf) $ our approximated Hessian is
exact.
This makes it necessary to define the Approximated Newton's step $ \eta \in \Wbfcal $ for an
arbitrary point in the parameter space as the solution to a regularised linear equation system.
Using the gradient from \cref{eq:diff_map_J_continuous} directly and the Hessian from
\cref{eq:hessian_J_continuous} combined with an identity matrix times a small factor we solve
\begin{equation}
  \label{eq:newtons_direction}
  \left( \Hess_\Wbf \widetilde \Jcal(\Wbf) + c ~ I_{\Nnet} \right) \eta = \nabla_\Wbf
  \widetilde \Jcal(\Wbf)
\end{equation}
for $ \eta $, where $ c = 10^{-5} $ controls the strength of regularisation.
All experiments use this value if not otherwise stated.
The regularisation is important, because the approximated Hessian is only valid in a restricted
neighbourhood around critical points.
Outside of critical points, the regularisation ensures that the gradient direction is always part
of the Newton's step.
Thus, this disturbance would also help with avoiding saddle points, as the descent direction is not
coinciding completely with a correct Newton's direction.
Additionally, for first-order only information saddles and maxima are numerically unstable.

In our implementation, we do not make use of automatic differentiation frameworks due to the
following reasons.
First, they are not able to introduce approximations to symbolic expressions on their own.
As we already have the required differential maps available due to our theoretical investigation, we
could realise the optimisation procedure by hand without too much overhead.
Second, to the best of our knowledge, the operations involved in the Hessian are not suited for
general purpose graphics processing units, hence the performance gain of automatic differentiation
frameworks is minimal.
Third, as we have to solve linear equation systems, we cannot avoid to use other libraries as well.
This would add additional overhead in the data transfer and further neglect benefits of these
frameworks.
In conclusion, by implementing the algorithm manually we achieved the full control over all the
components and also could make use of sophisticated parallelisation.
Of course, recent developments and contributions to automatic differentiation frameworks can
render these considerations obsolete.

\subsection{Pseudocode}
\label{ssec:pseudocode}
In our numerical analysis we investigate two RL procedures.
The first one is \emph{Policy Evaluation} as shown in \Cref{alg:peval}.
It makes use of a Gauss Newton Residual Gradient algorithm to approximate $ V_\pi $ with a given
MLP.
The required inputs are initial parameters for the MLP as well as a policy to evaluate.
As output one obtains the optimal parameters.
\begin{algorithm}
  \begin{flushleft}
  \caption{Policy Evaluation with Gauss Newton Residual Gradient Formulation}
  \label{alg:peval}
  \begin{algorithmic}[1]
    \Statex \textbf{Hyper parameters:}
    $ \gamma \in \left(0, 1\right)$,
    $ \alpha > 0 $,
    $ c = 10^{-5} $,
    $ \epsilon \leq 10^{-5} $,
    $ N \sim \Nnet $
    \Statex \textbf{Input:}
    \Statex - MLP $ f \in \MLP{n_0}{1} $ with initialised parameters $ \Wbf \in \Wbfcal $
    \Statex - policy $ \pi $
    \Statex - unique sample states $ s_i \in \Scal $ with $ i = 1, \ldots, N $
    \Statex \textbf{Output:} $ \Wbf $ such that
    $ f(\Wbf, s_i) \approx  V_\pi(s_i) \quad \forall i=1, \ldots, N$
    \newline
    \State Construct transition tuples $ (s_i, r_i, s_i') $ using $ \pi $ for all $ i $
    \newline
    \Do
      \newline
      \State Evaluate
      $ F(\Wbf) \coloneqq \left[ f(\Wbf, s_1) \ldots f(\Wbf, s_N) \right]^\top \in \Rds^N $
      and its differential map $ G(\Wbf) $.
      \State Evaluate $ F' $ and $ G' $ using $ s_i' $.
      \newline
      \State NMSBE: $ \widetilde \Jcal(\Wbf) = \frac{1}{2N} \Delta_\pi(\Wbf)^\top \Delta_\pi(\Wbf) $
      \State Gradient: $ \nabla_\Wbf \widetilde \Jcal(\Wbf) = \frac1N \Big( G(\Wbf) - \gamma
      G'(\Wbf) \Big)^\top \Delta_{\pi}(\Wbf) $
      \State Hessian: $ \Hess_\Wbf \widetilde \Jcal(\Wbf) = \frac1N
      \Big( G(\Wbf) - \gamma G'(\Wbf) \Big)^\top \Big( G(\Wbf) - \gamma G'(\Wbf) \Big) $
      \State Solve for $ \eta $: $ \left( \Hess_\Wbf \widetilde \Jcal(\Wbf) + c I_{\Nnet} \right)
      \eta = \nabla_\Wbf \widetilde \Jcal(\Wbf) $
      (e.g. with Householder QR-Decomposition)
      \newline
      \State Descent step: $ \Wbf \gets \Wbf - \alpha \eta $
      \newline
    \DoWhile{$ \widetilde \Jcal(\Wbf) > \epsilon $}
  \end{algorithmic}
  \end{flushleft}
\end{algorithm}
The second algorithm is normal \emph{Policy Iteration}, which makes use of the first and extends it
with a \emph{Policy Improvement} step.
After every sweep consisting of evaluation and improvement a better performing policy should be
available.
It is depicted in \Cref{alg:pimp}.

A common practice in Policy Iteration is to reuse the most recent evaluation outcome for the next
iteration.
In \citep{sigaud:nn19}, this practice is called \emph{persistent} whereas running the evaluation
from scratch in every sweep is referred to as \emph{transient}.
We employ this naming convention here as well.
\begin{algorithm}
  \begin{flushleft}
  \caption{Policy Iteration}
  \label{alg:pimp}
  \begin{algorithmic}[1]
    \Statex \textbf{Hyper parameters:}
    $ \gamma \in \left(0, 1\right)$,
    $ \alpha > 0 $,
    $ c = 10^{-5} $,
    $ \epsilon \leq 10^{-5} $,
    $ N \sim \Nnet $,
    $ sweeps > 0 $
    \Statex \textbf{Input:}
    MLP $ f \in \MLP{n_0}{1} $ with parameters $ \Wbf \in \Wbfcal $,
    \Statex \textbf{Output:} policy $ \pi $
    \newline
    \State Draw $ \Wbf $ elementwise uniformly from $ [-1, 1] $
    \State $ \pi(s) \gets \text{GIP}(f, \Wbf, s) $ for any $ s \in \Scal $
    \State Draw $ s_i $ for $ i = 1, \ldots, N $ uniformly from $ \Scal $
    \newline
    \For{sweep \textbf{in} sweeps}
      \newline
      \If{\textbf{not} persistent}
        \State Draw $ \Wbf $ elementwise uniformly from $ [-1, 1] $
      \EndIf
      \newline
      \State $ \Wbf \gets \text{PolicyEvaluation}(f, \Wbf, \pi, s_1, \ldots, s_N) $
      \State $ \pi'(s) \gets \text{GIP}(f_\pi, \Wbf, s) $ for all $ s \in \Scal $
      \State Evaluate $ \pi' $ empirically using several roll-outs
      \State $ \pi \gets \pi' $
      \newline
    \EndFor
  \end{algorithmic}
  \end{flushleft}
\end{algorithm}

\subsection{Demonstration of Local Quadratic Convergence}
\label{ssec:quad_conv_demo}
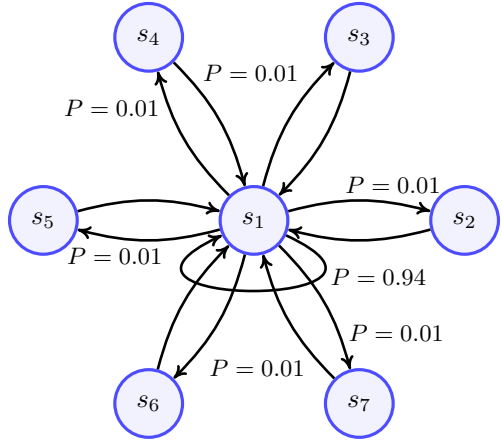
\begin{figure}
  \centering
  \begin{tikzpicture}[scale=0.8,
    state/.style={circle, draw=blue!70, fill=blue!5, very thick, minimum size=9mm}]

    \node[state] at (0.00,  0.00) (state1) {$ s_1 $};

    \def\r{3.5}

    \node[state] at (\r *  1.00, \r *  0.00) (state2) {$ s_2 $};
    \node[state] at (\r *  0.50, \r *  0.87) (state3) {$ s_3 $};
    \node[state] at (\r * -0.50, \r *  0.87) (state4) {$ s_4 $};
    \node[state] at (\r * -1.00, \r *  0.00) (state5) {$ s_5 $};
    \node[state] at (\r * -0.50, \r * -0.87) (state6) {$ s_6 $};
    \node[state] at (\r *  0.50, \r * -0.87) (state7) {$ s_7 $};

    \draw[-{>[scale=2.5,length=2,width=2]},line width=1pt] (state1) to [bend left=15]
    node[near end, above] {\footnotesize $ P = 0.01 $} (state2);
    \draw[-{>[scale=2.5,length=2,width=2]},line width=1pt] (state2) to [bend left=15] (state1);

    \draw[-{>[scale=2.5,length=2,width=2]},line width=1pt] (state1) to [bend left=15]
    node[near end, above left] {\footnotesize $ P = 0.01 $} (state3);
    \draw[-{>[scale=2.5,length=2,width=2]},line width=1pt] (state3) to [bend left=15] (state1);

    \draw[-{>[scale=2.5,length=2,width=2]},line width=1pt] (state1) to [bend left=15]
    node[near end, left] {\footnotesize $ P = 0.01 $} (state4);
    \draw[-{>[scale=2.5,length=2,width=2]},line width=1pt] (state4) to [bend left=15] (state1);

    \draw[-{>[scale=2.5,length=2,width=2]},line width=1pt] (state1) to [bend left=15]
    node[near end, below] {\footnotesize $ P = 0.01 $} (state5);
    \draw[-{>[scale=2.5,length=2,width=2]},line width=1pt] (state5) to [bend left=15] (state1);

    \draw[-{>[scale=2.5,length=2,width=2]},line width=1pt] (state1) to [bend left=15]
    node[near end, below right] {\footnotesize $ P = 0.01 $} (state6);
    \draw[-{>[scale=2.5,length=2,width=2]},line width=1pt] (state6) to [bend left=15] (state1);

    \draw[-{>[scale=2.5,length=2,width=2]},line width=1pt] (state1) to [bend left=15]
    node[near end, right] {\footnotesize $ P = 0.01 $} (state7);
    \draw[-{>[scale=2.5,length=2,width=2]},line width=1pt] (state7) to [bend left=15] (state1);

    \draw[-{>[scale=2.5,length=2,width=2]},line width=1pt] (state1) to
    [out=360-25,in=180+25,looseness=7.0] node[near start, right] {\footnotesize $ P = 0.94 $}
    (state1);

  \end{tikzpicture}
  \caption{Adapted version of Baird's \emph{Seven State Star Problem} \citep{bair:icml95}.
  We have added transitions with low probabilities from the central node back to the six outer
  states to obtain an infinite horizon problem such that the NMSBE is well-defined.}
  \label{fig:seven_star_graph}
\end{figure}

To evaluate empirically our derived theoretical results in the discrete state space setting
together with the assumption of exact learning, we demonstrate local quadratic convergence on
Baird's \emph{Seven State Star Problem} \citep{bair:icml95}.
Only if all components work as intended, local quadratic convergence of a second-order gradient
descent algorithm can be visible.
As we require a problem, where all states occur infinite many times for the NMSBE to be
well-defined, we extend the Star Problem with transitions from
the central node to all others.
In \cref{fig:seven_star_graph}, all transitions and their probabilities are shown.
A policy for evaluation is defined implicitly by setting the transition probabilities to fixed
values.
The reward is present at the central node with value one.
As discount factor we use $ \gamma = 0.99 $.

We deploy an MLP architecture $ \Fcal(2, 7, 1) $ consisting of $ \Nnet = 29 $ parameters with step
size $ \alpha = 1 $ and use Bent-Id as activation function in hidden layers.
Every state receives a unique two-dimensional random feature to embed the discrete states in a
vector space for the input layer.
Features are drawn from a normal distribution with zero mean and unit covariance.

\begin{figure}
  \centering
  \begin{subfigure}{0.5\linewidth}
    \centering
    \includegraphics[width=1.0\linewidth]{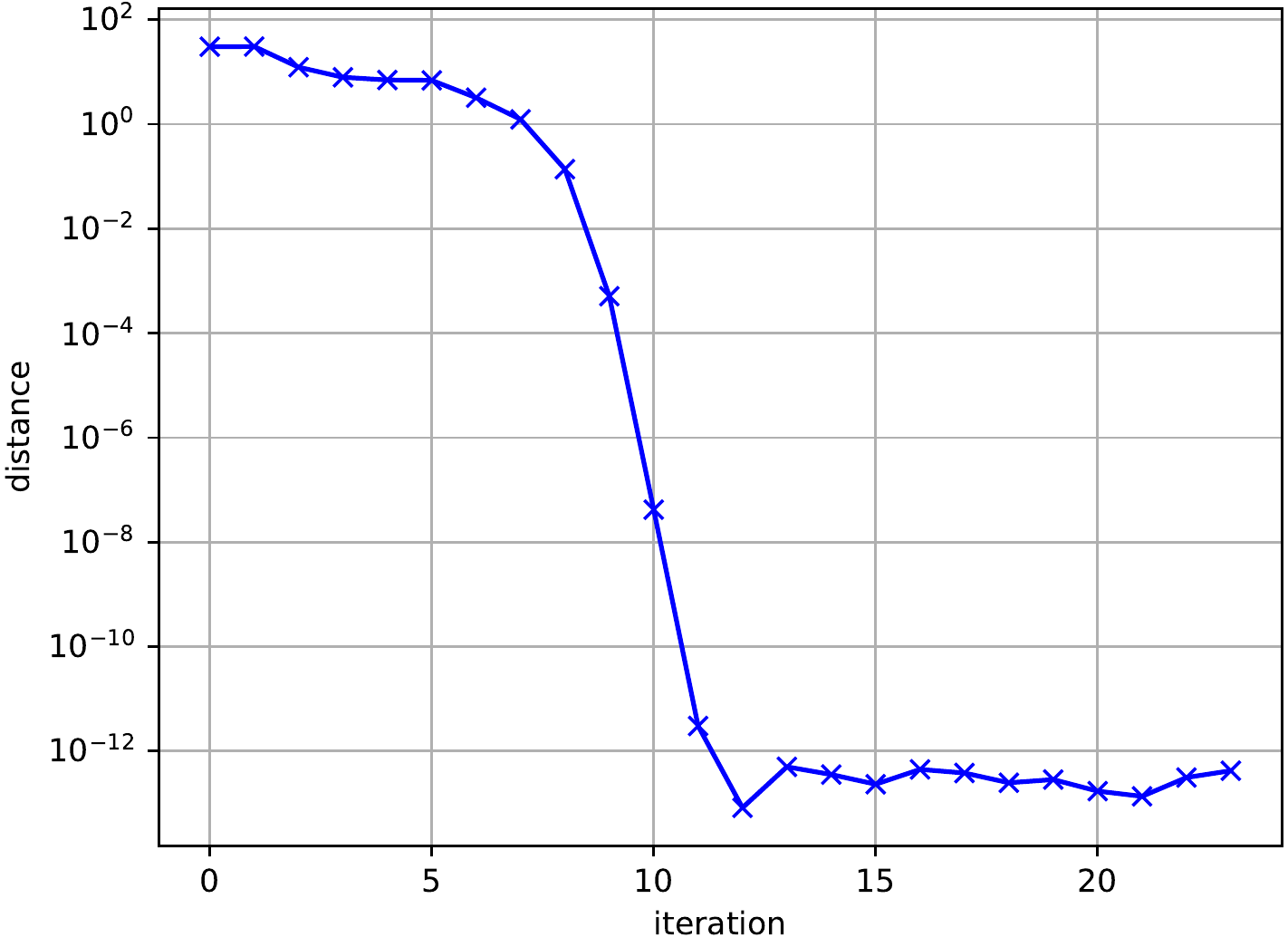}
    \caption{}
    \label{fig:seven_star_distance}
  \end{subfigure}%
  \begin{subfigure}{0.5\linewidth}
    \centering
    \includegraphics[width=1.0\linewidth]{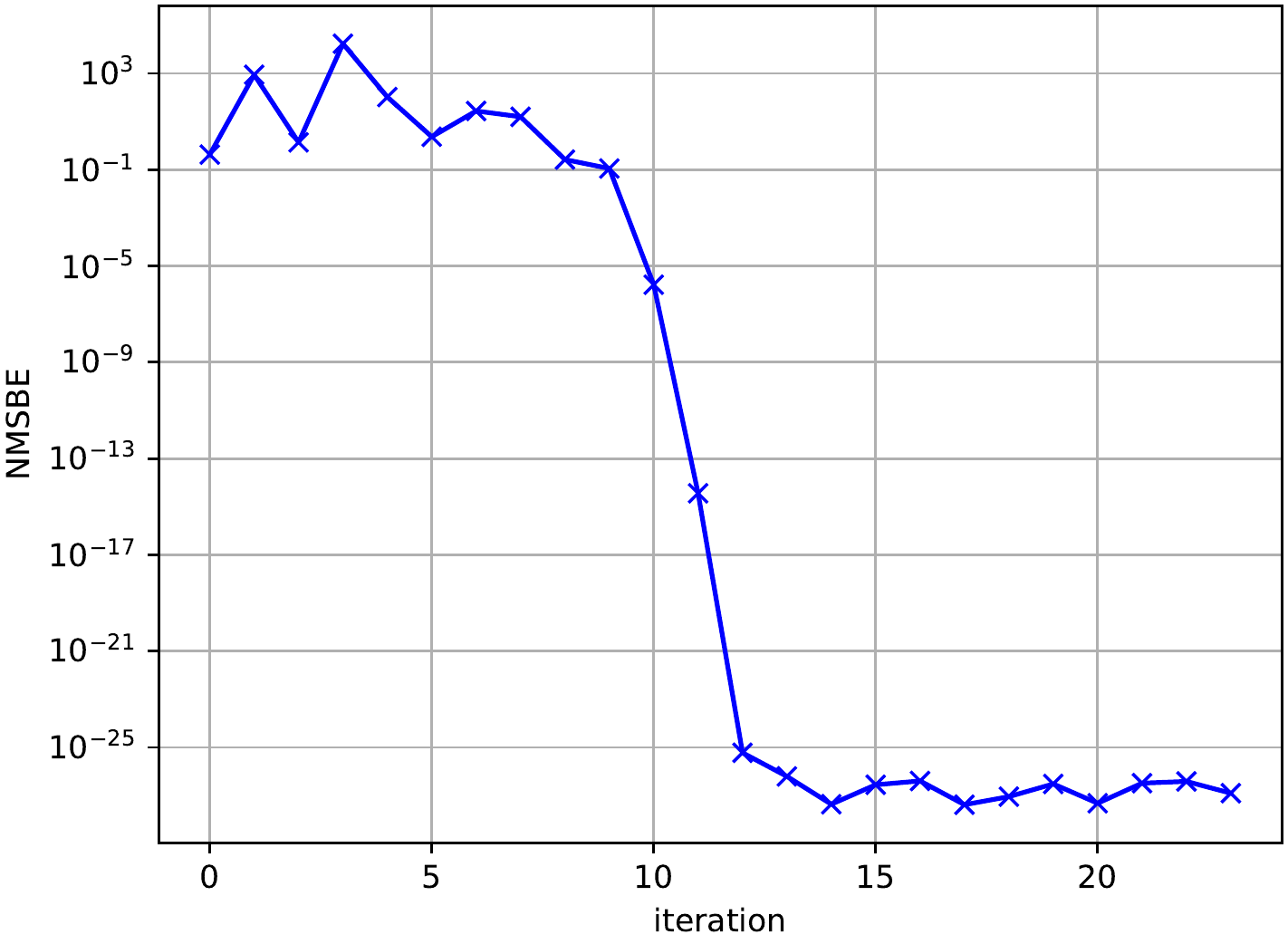}
    \caption{}
    \label{fig:seven_star_msbe}
  \end{subfigure}
  \caption{
  A typical local quadratic convergence behaviour for an adapted version of Baird's
  \emph{Seven State Star Problem}.
  \textbf{a)}: Distance of iterates $ \Wbf^{(k)} $  to the accumulation point $ \Wbf^* $.
  \textbf{b)}: The NMSBE corresponding to each iterate.}
  \label{fig:seven_star}
\end{figure}

In \cref{fig:seven_star}, we visualise convergence behaviour with the distance of the accumulation
point $ \Wbf^* $ to all iterates $ \Wbf^{(k)} $ as well as by the corresponding NMSBE.
We use the last iterate as $ \Wbf^* $ and measure the distance by extending the Frobenius norm of
matrices to collections of matrices as
$ \| \Wbf^{(k)} - \Wbf^{*} \|_F^2 \coloneqq \sum_{l=1}^L \| W_l^{(k)} - W_l^* \|_F^2 $.

It is clear from \cref{fig:seven_star_distance} that the AN algorithm converges locally
quadratically to a minimiser.
Judging from the negligibly small final NMSBE as seen \cref{fig:seven_star_msbe},
we conclude that the MLP is an exact approximator of the ground truth value function.
Both graphs together imply
that the approximated Hessian is exact close to $ \Wbf^* $ and additionally that its rank is full.
Due to over-parametrisation, convergence to suboptimal local minima does not happen.

%
\section{Experiments in Continuous State Spaces}
\label{sec:experiments}
For our empirical investigation of the algorithm we first outline the experimental setup.
Second, we test the convergence behaviour in continuous state spaces.
We compare Residual Gradients with Semi-Gradients and quantify the influence of second-order
optimisation.
Next, we explore the generalisation capabilities of an MLP when trained with a Gauss Newton
Residual Gradient algorithm by evaluating the NMSBE on unseen states outside of the training set.
Finally, we address the application of a second-order Residual Gradient algorithm in full Policy
Iteration and test it in a continuous control problem.

\subsection{Experimental Setup}
\label{ssec:experiments_setup}
We apply the Gauss Newton Residual Gradient algorithm for Policy Evaluation in finite dimensional
and bounded Euclidean state spaces.
We provide empirical results for the performance of the Approximated Newton's algorithm by
minimising the objective of \cref{eq:nmsbe_continuous} in several different scenarios:
\begin{enumerate}
  \item \textbf{Convergence of Policy Evaluation} by evaluating the value function of a fixed
  policy;

  \item \textbf{Generalisation Capabilities} by characterising the influence of different MLP
  architectures on the generalisation performance;



  \item \textbf{Full Policy Iteration} by combining Policy Evaluation and $ Q $-factors to improve
  iteratively an initial policy.
\end{enumerate}
Investigating performance of the algorithm with generalisation in mind is both interesting and
important due to still-unexplainable capabilities of neural networks
\citep{lawr:aaai97, zhan:iclr17, neys:nips17}.
Testing in a full Policy Iteration setting is important for the general applicability of a
Gauss Newton Residual Gradient algorithm.

In our Policy Evaluation experiments,
except for generalisation,
we compare the cases of whether or not considering derivatives of the TD-target.
This means we compare Semi-Gradient and Residual Gradient formulations for both first and second
order optimisation methods.

For the experiments, we use the \emph{Mountain Car} \citep{moor:tr90} and the \emph{Cart Pole}
control problems \citep{barto:tsmc83} as environments.
These two are classic deterministic Reinforcement Learning benchmarks with a typical continuous
state space and a manageable complexity that allows for an extensive investigation and
visualisation.
More specifically, we employ the \mbox{\emph{MountainCar-v0}} environment of the
\mbox{\emph{OpenAI Gym}} package \citep{brockman:corr16}, and include important changes to the
environment to obtain infinite horizon problems.
First, we replace the built-in constant reward function, which assigns independently of a state, an
action and its successor state to any transition a punishment of $ -1 $, with a function that
rewards being in a goal region by not giving punishments in that area.
We define the goal region as $ x > 0.5 $, where $ x $ denotes the position of the car.
Second, once the car enters the goal region we teleport it back to a starting state.
These two adjustments allow us to use the environment in a non-episodic fashion and, thereby,
establish the required formulation of the MDP to work with uniformly sampled state transitions.
Without these adjustments ergodicity cannot be present and the objective would not be well-defined.
Similarly, the \mbox{\emph{CartPole-v1}} environment of \mbox{\emph{OpenAI Gym}} receives a new
non-constant reward and a new transition.
Instead of rewarding every step with $ +1 $, including those states which are considered to be a
failure because the pole fell over or the cart left the allowed range,
we only punish with $ -1 $ reward the transition into those failure states.
Otherwise, the reward is zero.
By adding connections from terminal region to the start state, we convert the episodic formulation
of the balancing task and restore also here an infinite horizon MDP.
We set the discount factor for all environments to $ \gamma = 0.99 $ if not otherwise stated.

\subsection{Empirical Convergence Analysis}
\label{ssec:experiments_conv}
\paragraph{Setting}
We first make use of the Mountain Car problem, which has a continuous state space and discrete
actions, and investigate the performance when evaluating a predefined policy.
The complexity and dimensionality of the problem is manageable, which enables us to estimate the
ground truth value function with Monte Carlo methods based on a fine grained two dimensional grid.
Thus, we can evaluate accurately the performance of tested algorithms against the ground truth.
The policy, whose value function is being evaluated, is fixed to accelerate the car in the
direction of the current velocity.

We investigate the performance of the algorithm under the influence of three variants:
\emph{ignoring TD targets in derivatives}, \emph{using Hessian based optimisation} and
\emph{varying learning rates}.
We use four different learning rates $\alpha \in \{10^{0}, 10^{-1}, 10^{-2}, 10^{-3} \}$ and study
all 16 combinations.

For all tests, we adopt the batch learning scenario.
Specifically, we use as training set $ 100 $ transition tuples $ (s, r, s') $ collected in prior
with the fixed policy.
All states are sampled uniformly from $ \Scal $.
Note that changing the $ \xi $-weighted norm to an equivalent one, for example the uniform
distribution used here, can only change the steepest descent direction.
The types and locations of critical points remain the same.
Executing the action provided by the policy in each state yields its successor $ s' $ and the
one-step-reward $ r $ for that transition.
We fix the transition tuples throughout all convergence experiments to provide a fair comparison
between individual runs.

As function space for approximated value functions,
we employ the MLP $ \Fcal(2, 10, 10, 1) $ with Bent-Id activation functions.
This architecture has $ \Nnet = 151 $ free parameters, i.e., more parameters than the number of
samples, and complies to our theoretical findings.
Furthermore, MLPs in this experiment are all initialised with the same weight matrices to further
improve the comparability.
For initialisation we use a uniform distribution in the interval $ [-1,1] $.

\begin{figure}
  \centering
  \begin{subfigure}{0.5\linewidth}
    \includegraphics[width=1.0\linewidth]{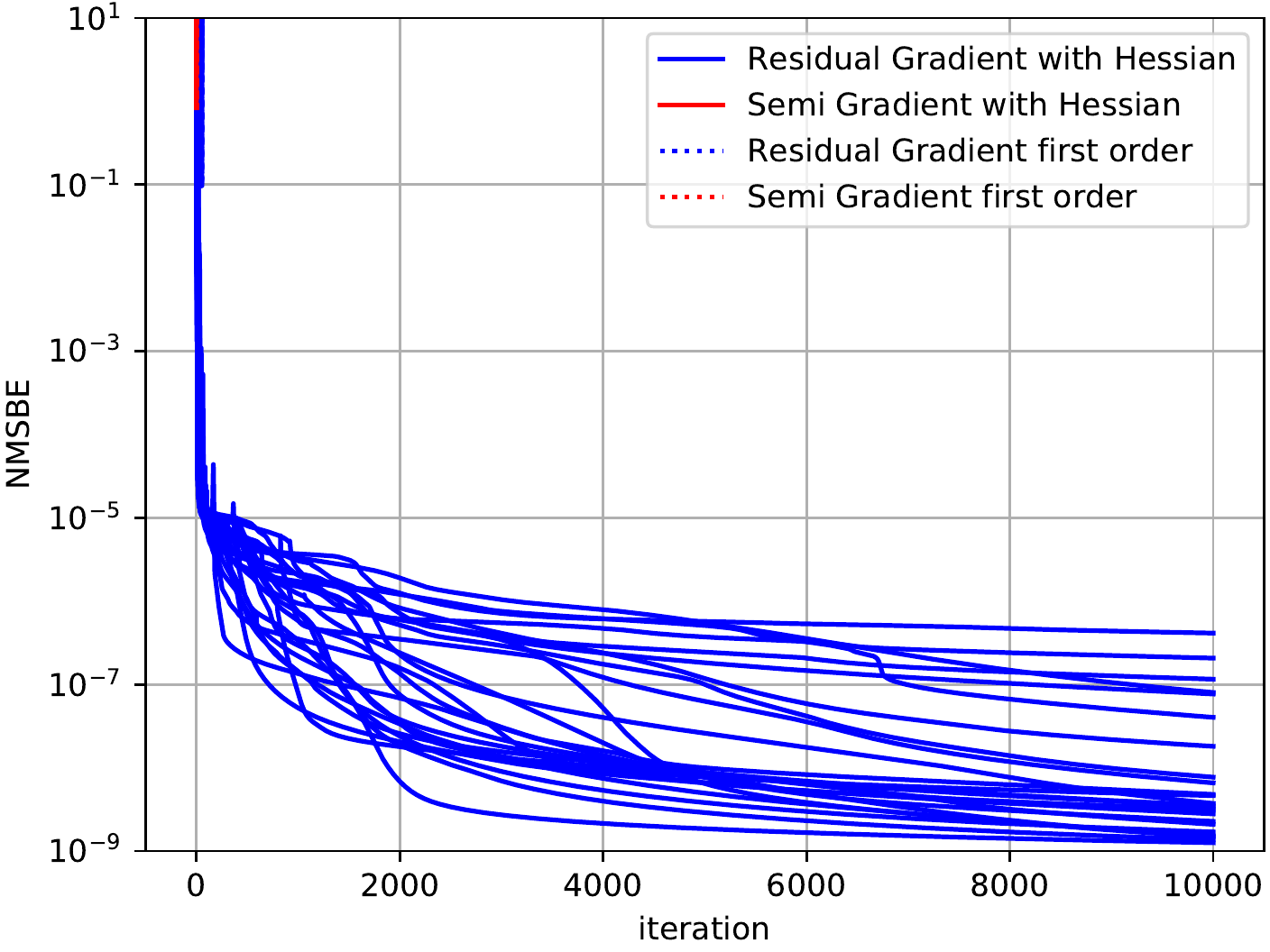}
    \caption{$ \alpha = 10^{0} $}
    \label{fig:influence_alpha_1e-00}
  \end{subfigure}%
  \begin{subfigure}{0.5\linewidth}
    \includegraphics[width=1.0\linewidth]{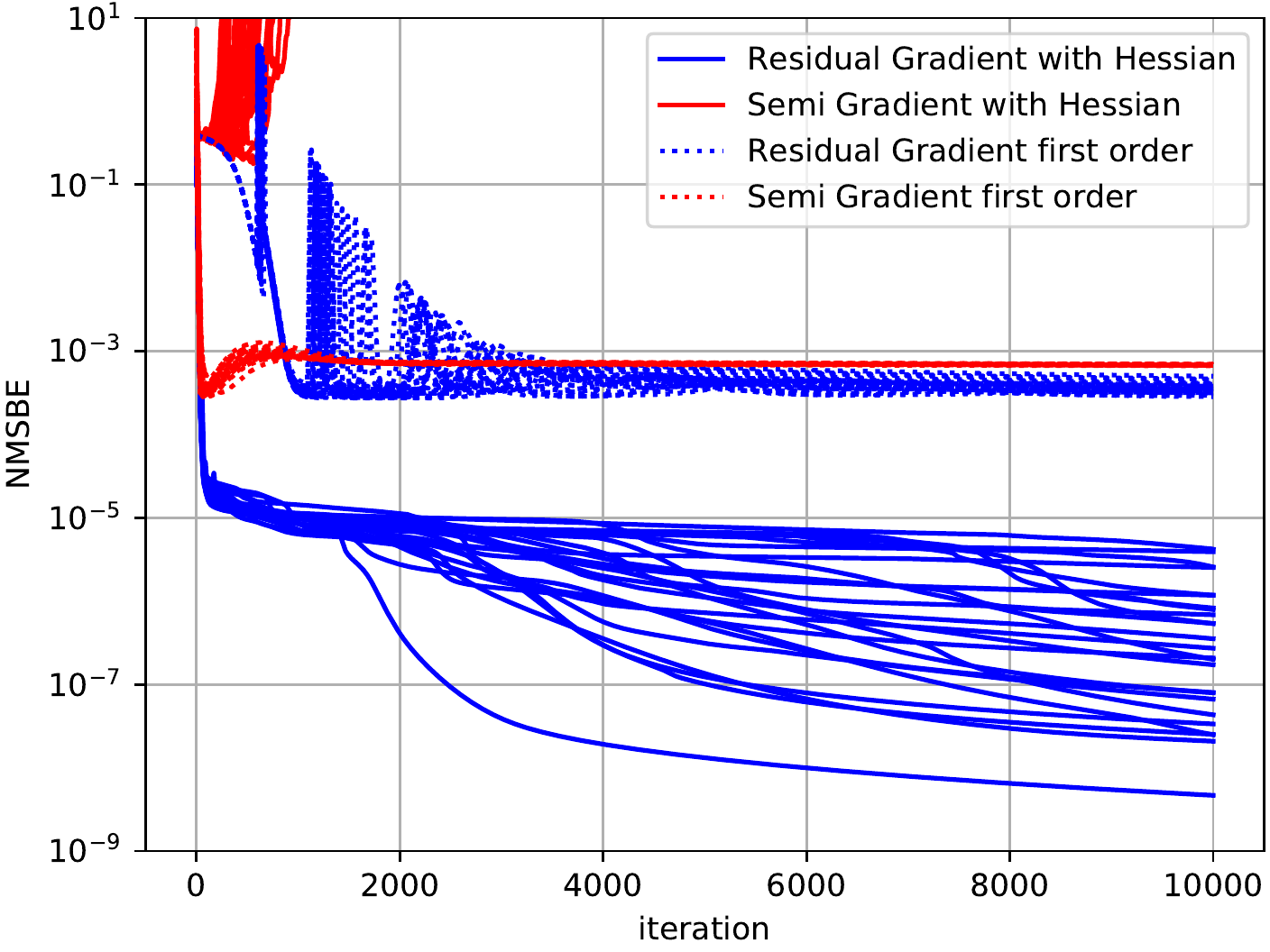}
    \caption{$ \alpha = 10^{-1} $}
    \label{fig:influence_alpha_1e-01}
  \end{subfigure}

  \begin{subfigure}{0.5\linewidth}
    \includegraphics[width=1.0\linewidth]{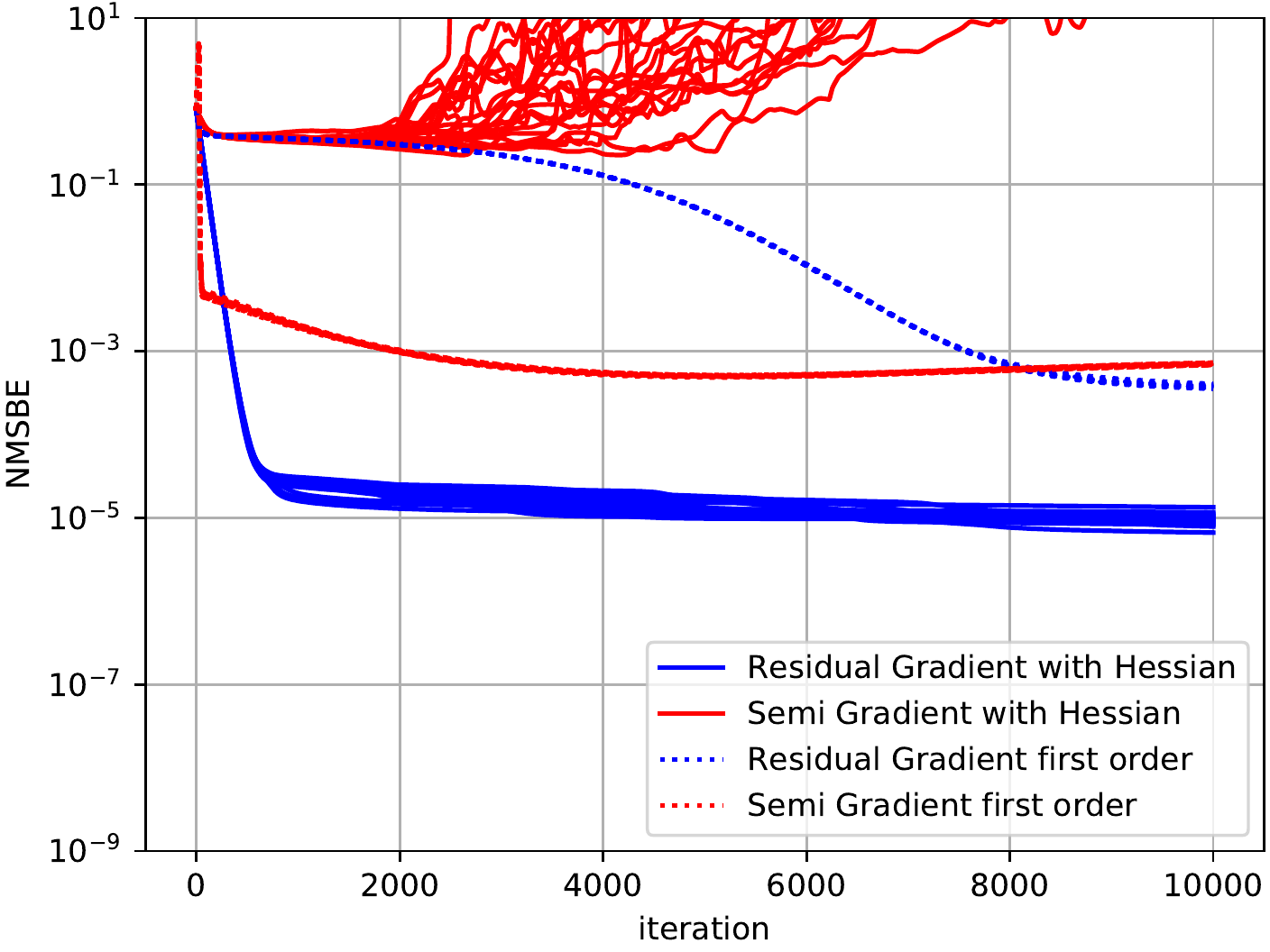}
    \caption{$ \alpha = 10^{-2} $}
    \label{fig:influence_alpha_1e-02}
  \end{subfigure}%
  \begin{subfigure}{0.5\linewidth}
    \includegraphics[width=1.0\linewidth]{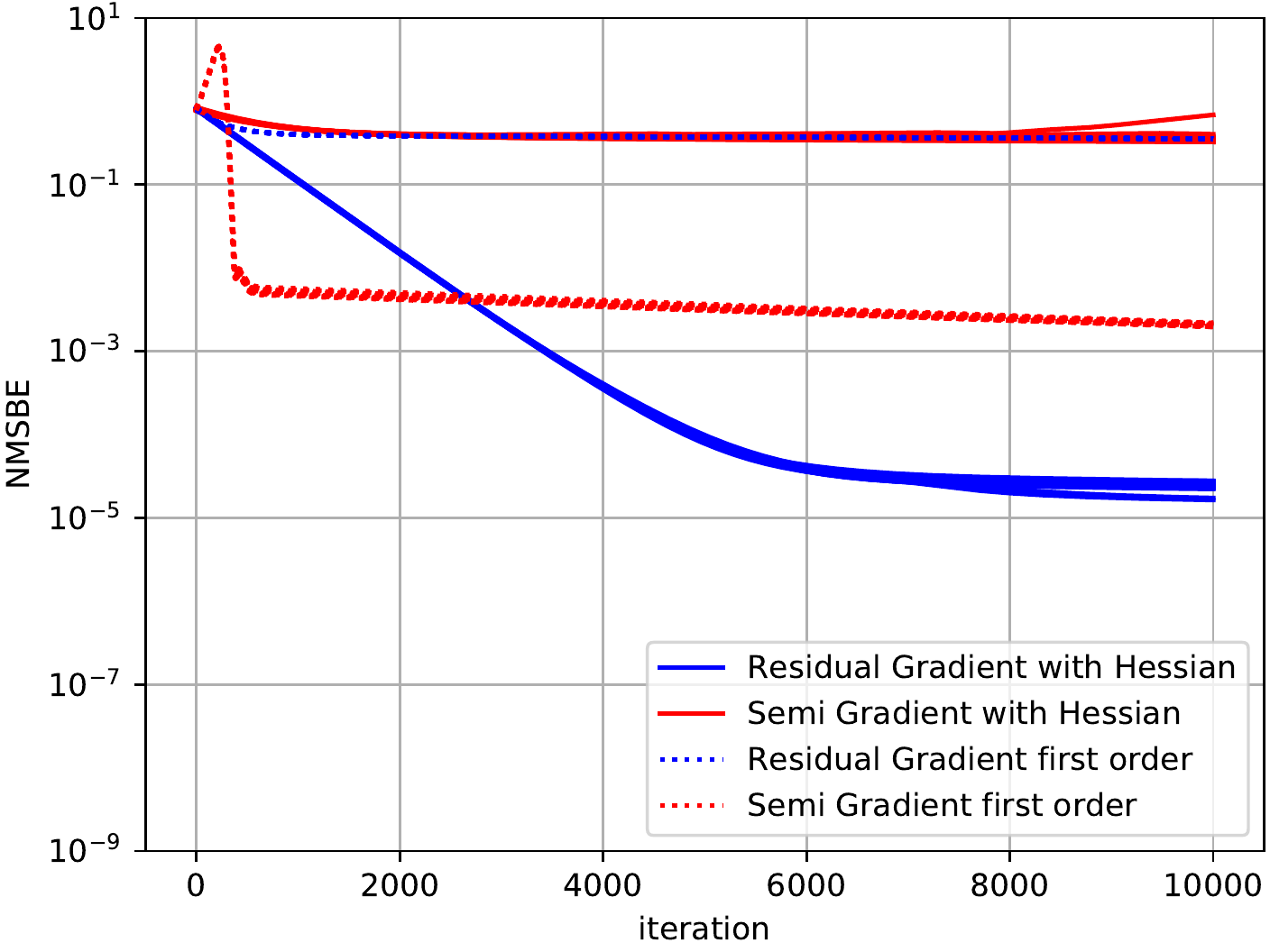}
    \caption{$ \alpha = 10^{-3} $}
    \label{fig:influence_alpha_1e-03}
  \end{subfigure}
  \caption{
  The NMSBE as defined in \cref{eq:nmsbe_continuous} plotted over time for all tested optimisation
  approaches.
  Figure a) to d) represent considered learning rates
  $ \alpha \in \left\{10^0, 10^{-1}, 10^{-2}, 10^{-3} \right\}$.
  A Residual Gradient formulation combined with Hessian based optimisation outperforms
  Semi-Gradient algorithms for all learning rates and stays convergent, even for large values of
  $ \alpha $.
  First-order only Residual Gradient algorithms demonstrate the reported slow convergence.}
  \label{fig:influence_alpha}
\end{figure}

Although there is no noise or randomness in the environment part involved and we do not make use of
exploration mechanisms, the outcomes can still vary.
We use an asynchronous multithreaded implementation such that numerical errors can either
accumulate or cancel out over time based on the order of execution.
Hence, all experiments are repeated $ 25 $ times.
Our results of the optimisation processes are shown in \cref{fig:influence_alpha}.

\paragraph{Results}
We observe that a Semi-Gradient algorithm always diverges for extremely large step sizes as shown
in \cref{fig:influence_alpha_1e-00} with $ \alpha = 1 $.
For smaller step sizes $ \alpha \in \left\{10^{-1}, 10^{-2}, 10^{-3} \right\} $ as shown in
\namecrefs{fig:influence_alpha} \ref{fig:influence_alpha_1e-01}, \ref{fig:influence_alpha_1e-02} and
\ref{fig:influence_alpha_1e-03}, a Semi-Gradient algorithm can converge if using first-order
optimisation.
%
%
Extending it to second-order gradient descent causes it to diverge sooner or later for all step
sizes.
For small enough learning rates, e.g. $ \alpha = 10^{-3} $, second-order Semi-Gradient additionally
achieves only the same final NMSBE as first-order Residual Gradient methods, indicating that the
computation for the approximated Semi-Hessian is obsolete.
However, we want to point out that the resulting solution of Policy Evaluation is not good compared
to other possible outcomes.

\cref{fig:influence_alpha_1e-01,fig:influence_alpha_1e-02} show that first-order gradient descent
algorithms with and without ignoring the dependency of the gradient on the TD target perform
consistently and achieve almost identical final errors.
Looking at the descent behaviour, we can confirm the slow convergence issue of a Residual Gradient
formulation.
From \cref{fig:influence_alpha_1e-03} it becomes clear that first-order Semi-Gradient descent
combined with carefully selected learning rates $ \alpha \in \left\{10^{-2}, 10^{-3} \right\} $ can
achieve an equal or even lower final error, further explaining its popularity over Residual
Gradients.

In contrast stands the Gauss Newton Residual Gradient algorithm, i.e., a Gauss Newton algorithm
combined with complete derivatives of the NMSBE.
This algorithm works well with all learning rates, in particular this algorithm works with large
learning rates as it can be seen in \cref{fig:influence_alpha_1e-00,fig:influence_alpha_1e-01}.
Even for extremely large learning rates $ \alpha = 1 $ convergence is not a problem.
Despite strong initial numerical problems, all repetitions arrive at a sufficiently small NMSBE over
time.
For all learning rates, the final value for the NMSBE is orders of magnitude smaller than that of
other approaches.
In other words, building the derivatives of the TD target with respect to the parameters and using
(approximate) second-order information of the NMSBE function are important ingredients for
designing and implementing efficient NN-VFA algorithms.
Modifying the descent direction based on the curvature is crucial to achieve good performance.
This insight matches also the empirical evidence that when combining Semi-Gradient algorithms with
an annealing scheme on the learning rate, even better performance can be obtained
\citep{gronauer:ijcai21}.

\paragraph{Computational concerns}
A severe burden of Newton-type algorithms is the computational complexity involved in evaluating
the Hessian, especially since the Gauss Newton approximation involves a full sized and dense matrix.
In this experiment, first-order only methods require roughly $ 8.4 $ and $ 7.6 $ seconds on average
for $ 10^4 $ iterations of Semi- and Residual Gradients, respectively.
Our experiments run on an \textit{AMD 3990X 64-Core} computer, but these execution times are not
supposed to be accurate and reproducible measurements.
They should only reveal the overall trend.
Calculating the Hessian and Newton's direction%
\footnote{using \emph{Eigen}'s Householder QR-decomposition when linked against \emph{OpenBlas} and
\emph{OpenLapack}}
increases the run time to $ 22.4 $ and $ 48.5 $ seconds, respectively.
The numbers imply that within $ 10^4 $ steps of a first-order Residual Gradient method only around
$ 1500 $ iterations of a Gauss Newton Residual Gradient algorithm can be performed on our computers.
However, the performance gain in terms of convergence speed and overall lower error, as seen in
\cref{fig:influence_alpha_1e-01} or \cref{fig:influence_alpha_1e-02}, still justifies the
additional computational effort.
A second-order algorithm is in total faster than first-order only methods, because far less
iterations are required to reach an already significantly smaller error.
The performance could be further enhanced if we would make use of concepts such as the
\emph{Levenberg-Marquardt} heuristic or put additional effort in hyper parameter tuning.

\subsection{Generalisation Capabilities of MLPs}
\label{ssec:experiments_gene}
As mentioned earlier in the critical point analysis, working with finite exact approximators based
on $ N $ samples from a continuous state space causes generalisation to be an important topic.
Since the value function approximator can only be trained to fit a finite set of samples exactly,
the generalisation capabilities of an MLP for states in between the collected training samples are
essential to RL and worth further investigation.
Thus, we evaluate an approximated value function, which we obtain by optimising the NMSBE with the
Gauss Newton Residual Gradient algorithm, with states that were not part of the training data.

\subsubsection{Single Architecture}
\label{sssec:experiment_gene_single}
\begin{figure}
  \centering
  \begin{subfigure}{1.0\linewidth}
    \centering
    \includegraphics[width=1.00\linewidth]{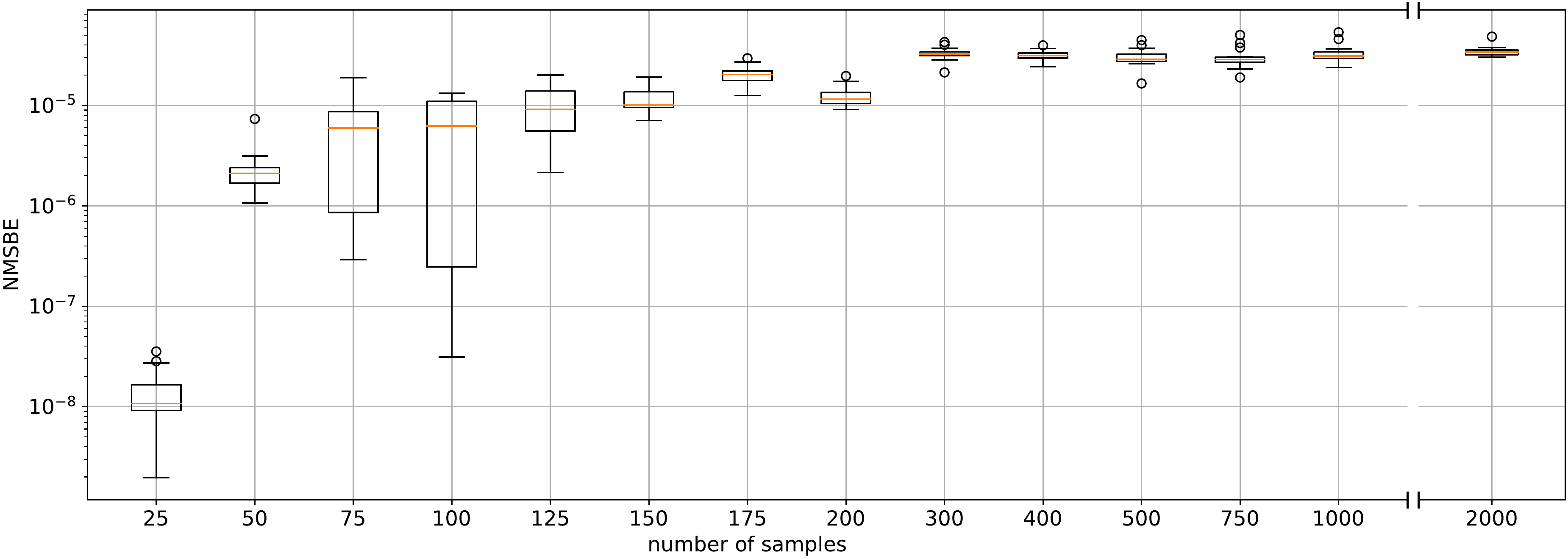}
    \caption{Training error}
    \label{fig:generalisation_box_training}
  \end{subfigure}
  \begin{subfigure}{1.0\linewidth}
    \centering
    \includegraphics[width=1.00\linewidth]{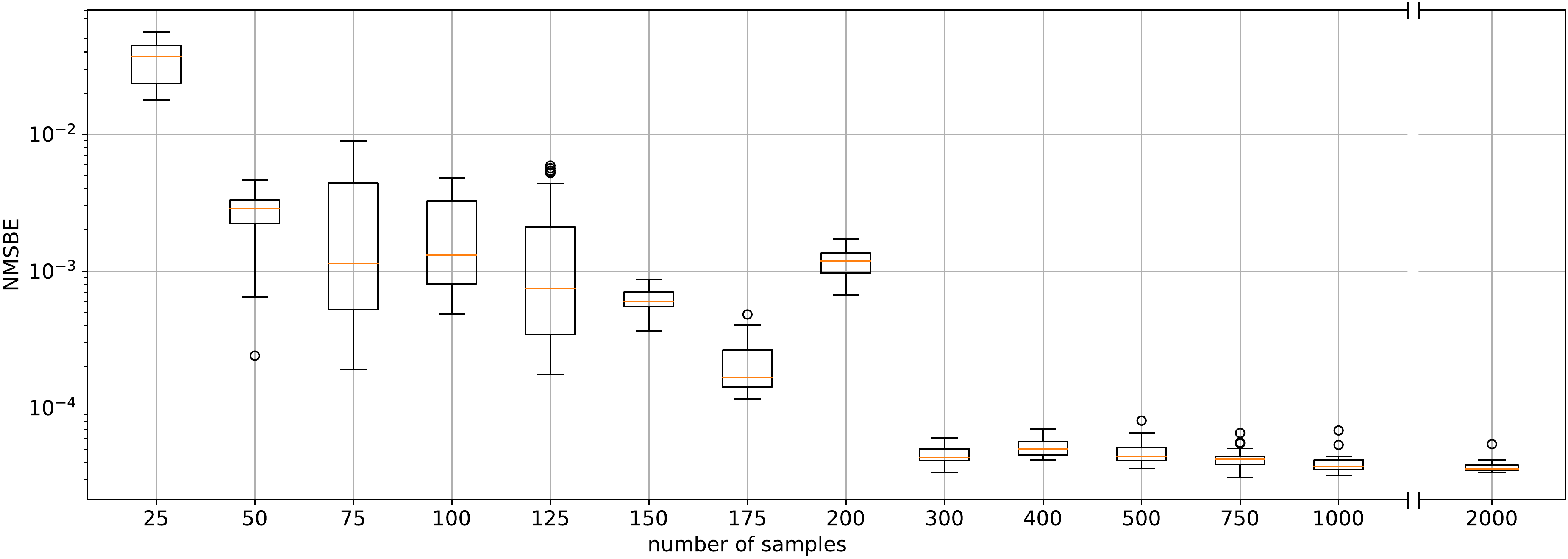}
    \caption{Test error}
    \label{fig:generalisation_box_testing}
  \end{subfigure}
  \caption{
  The final NMSBE of the MLP  $ \Fcal(2, 10, 10, 1) $ for different amount $N$ of training samples.
  \textbf{Top:} NMBSE using the training samples.
  \textbf{Bottom:} NMSBE for the test set.}
  \label{fig:batch_diff_T}
\end{figure}

\paragraph{Setting}
We start with a single architecture and vary the amount of training samples.
More specifically, we consider again the MLP $ \Fcal(2, 10, 10, 1) $ with learning rate
$ \alpha = 10^{-2} $ and \textit{Bend-Id} activation function.
For the initialisation we use a uniform distribution in the interval $ [-1,1] $.
In this experiment, the number of training samples $ N $ is varied non-uniformly between $ 25 $ and
$ 2000 $ and the NMSBE is computed for a separated test set comprised of $ 25 \cdot 10^4 $ states
arranged on a grid.
Again, we repeat for each $ N $ the training of MLPs $ 25 $ times and visualise the NMSBE as
box-plots in \cref{fig:generalisation_box_training} for the training data and in
\cref{fig:generalisation_box_testing}
for the held-out test set.

\paragraph{Results}
It is clear that at $ N = 2000 $ the training process is always successful with almost ignorable
variance.
While decreasing the number of randomly placed training samples down to $ N = 300 $, the mean error
becomes slightly larger with more pronounced variances.
Furthermore, only poor results are observed at $ N = 25 $.
Here, the MLP is able to fit the training data exactly, but the test error indicates that this is
no longer a valid solution.

We observe for the test error that the variance, when using $ N = 150 $ samples, is smaller than
that of its direct neighbours.
Simultaneously, there is a sharp reduction in the variance for training errors starting at
$ N = 125 $.
Since these numbers of samples are almost identical or at least close to the number of parameters in
the network ($ \Nnet = 151 $) this could be a numerical confirmation of the theoretical condition
as derived earlier.
However, we admit that an empirical verification is rather involved.
In particular, we require $ N $ unique samples in the training set, but there is no practical way to
determine, whether those $ N $ collected samples are \enquote{sufficiently unique}.

\begin{figure}
  \centering
  \begin{subfigure}{0.33\linewidth}
    \centering
    \includegraphics[width=1.0\linewidth]{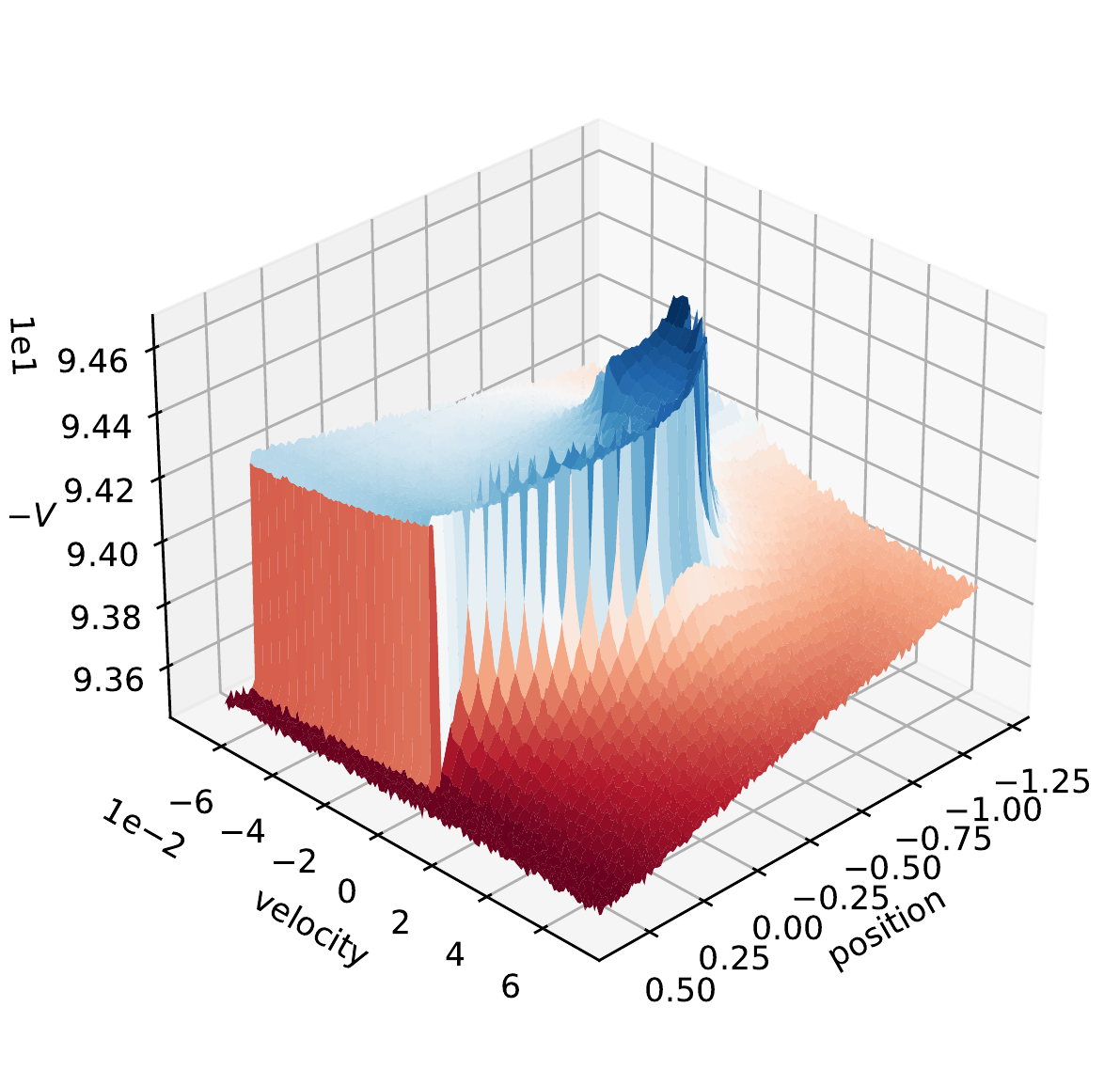}
    \caption{Ground Truth $ V_\pi $ obtained from rollouts.}
    \label{fig:V_net_MC}
  \end{subfigure}%
  \begin{subfigure}{0.33\linewidth}
    \centering
    \includegraphics[width=1.0\linewidth]{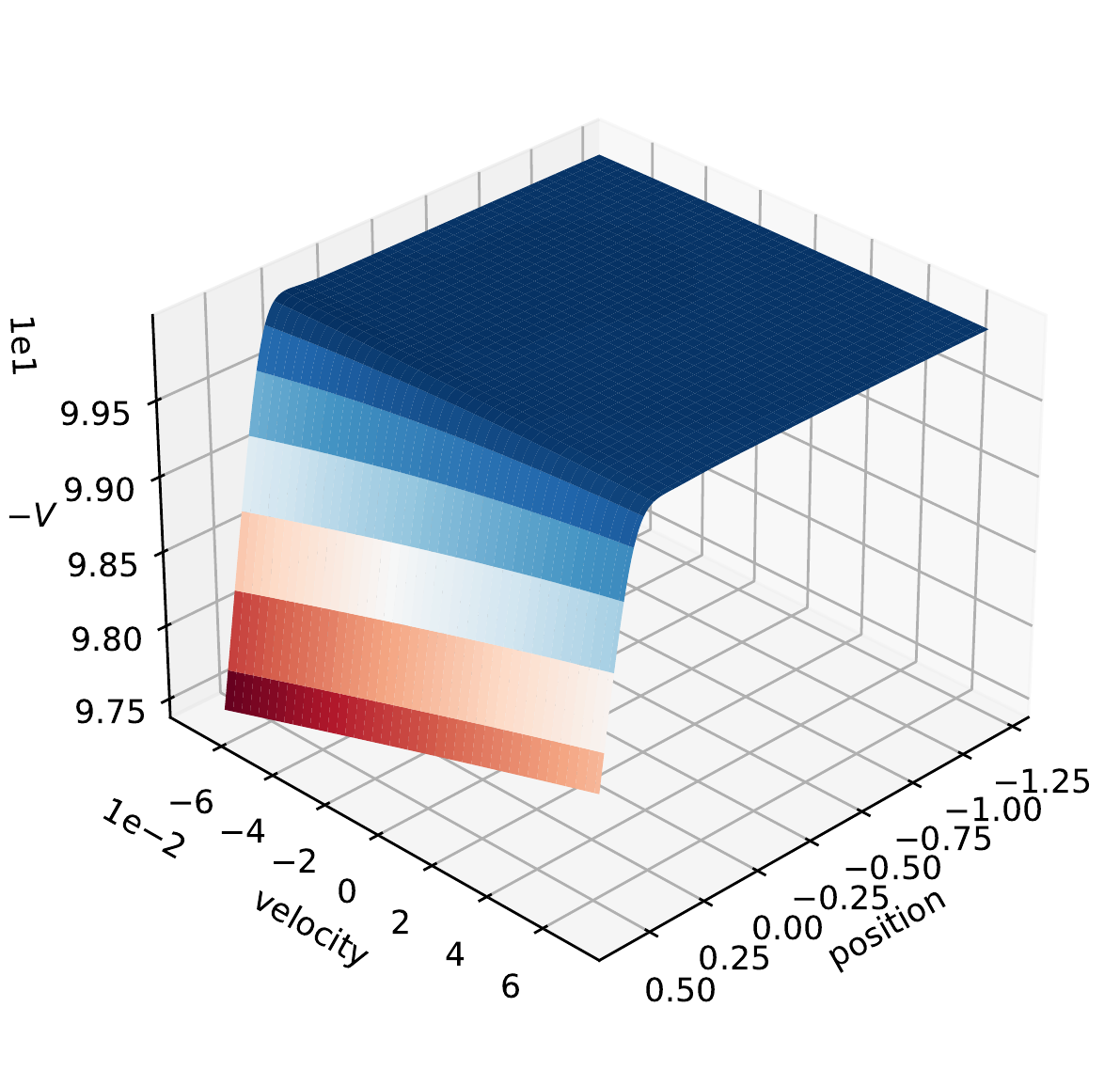}
    \caption{$ N = 25 $}
    \label{fig:V_net_N_25}
  \end{subfigure}%
  \begin{subfigure}{0.33\linewidth}
    \centering
    \includegraphics[width=1.0\linewidth]{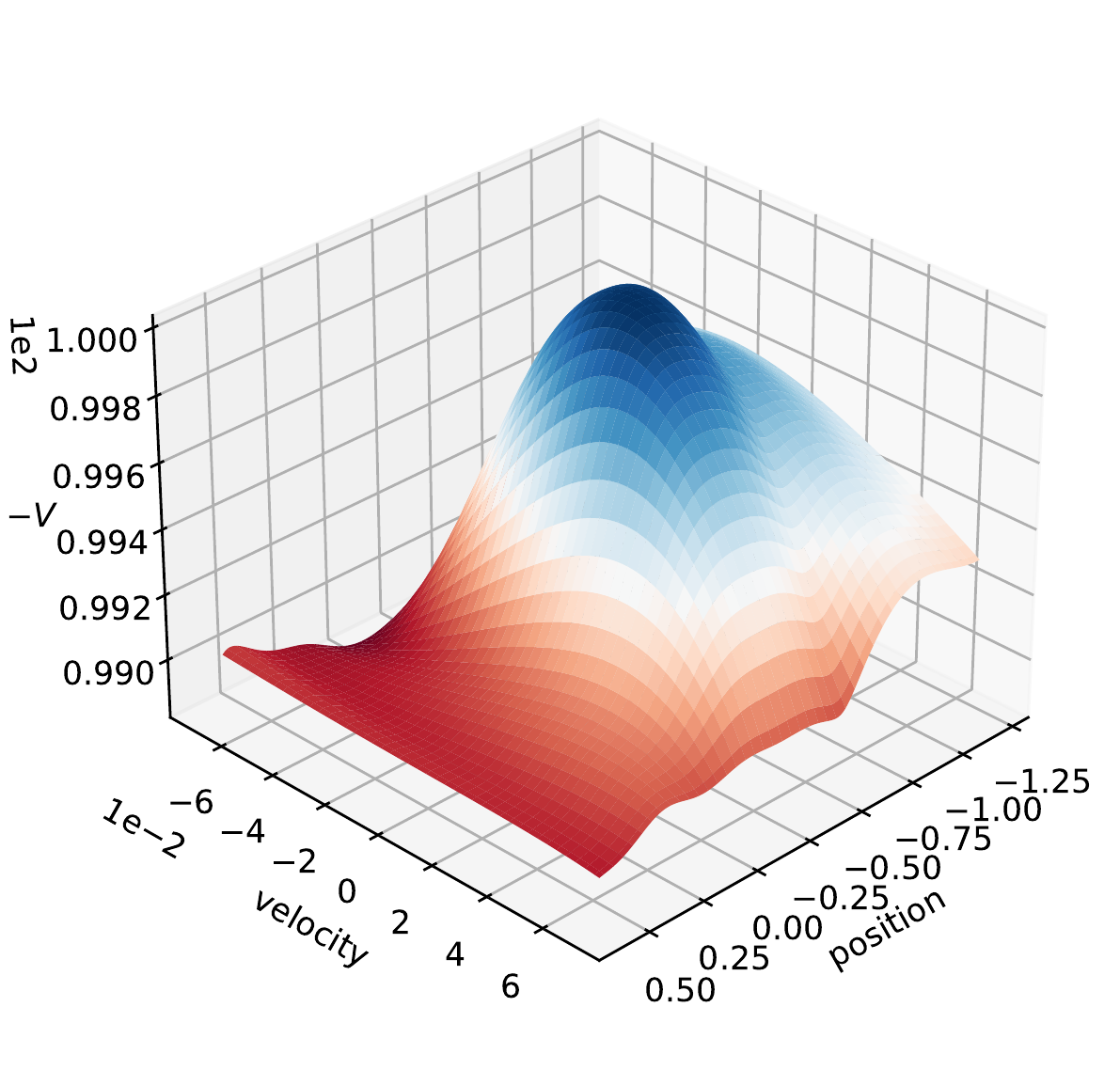}
    \caption{$ N = 50 $}
    \label{fig:V_net_N_50}
  \end{subfigure}

  \begin{subfigure}{0.33\linewidth}
    \centering
    \includegraphics[width=1.0\linewidth]{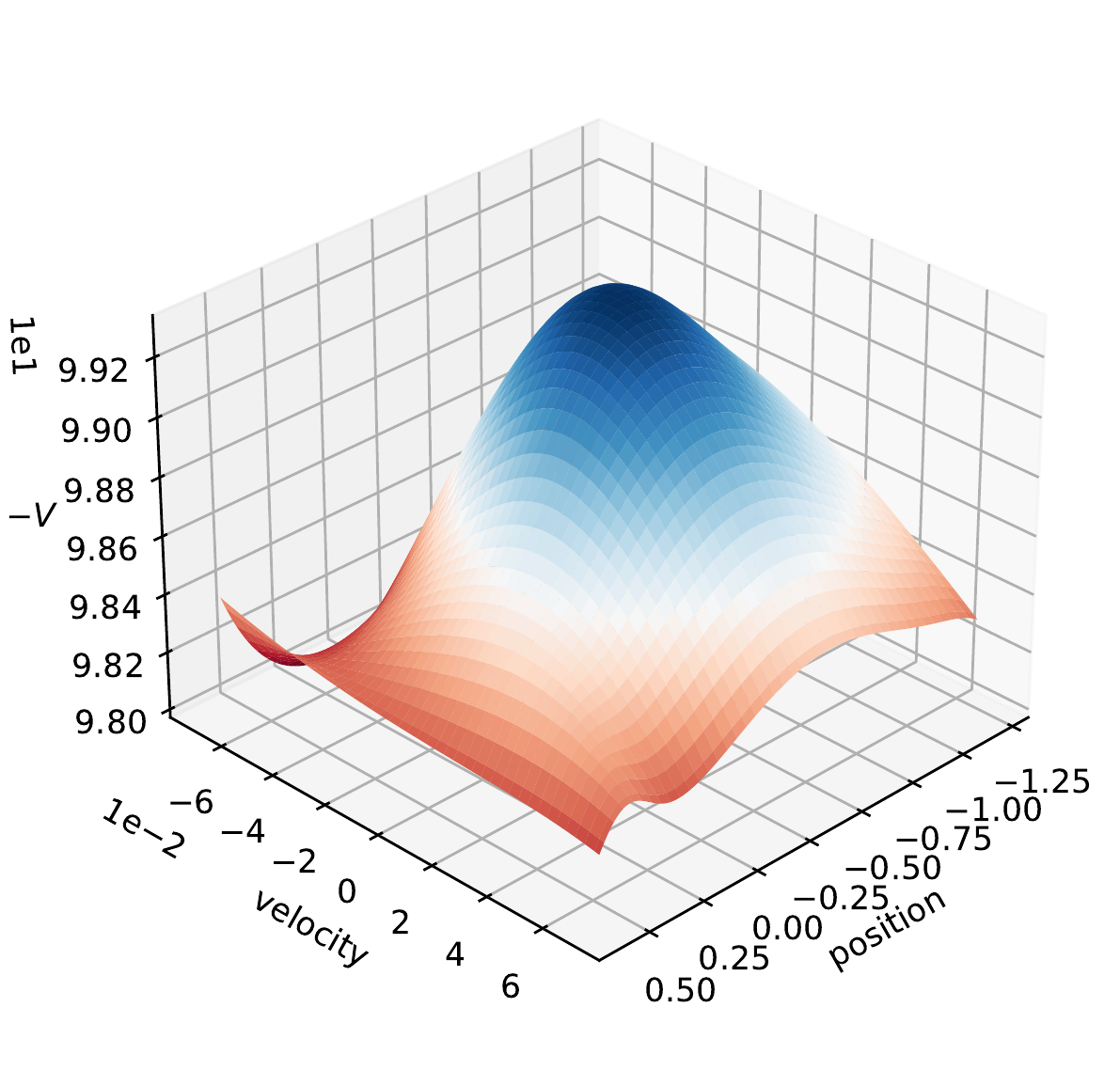}
    \caption{$ N = 100 $}
    \label{fig:V_net_N_150}
  \end{subfigure}%
  \begin{subfigure}{0.33\linewidth}
    \centering
    \includegraphics[width=1.0\linewidth]{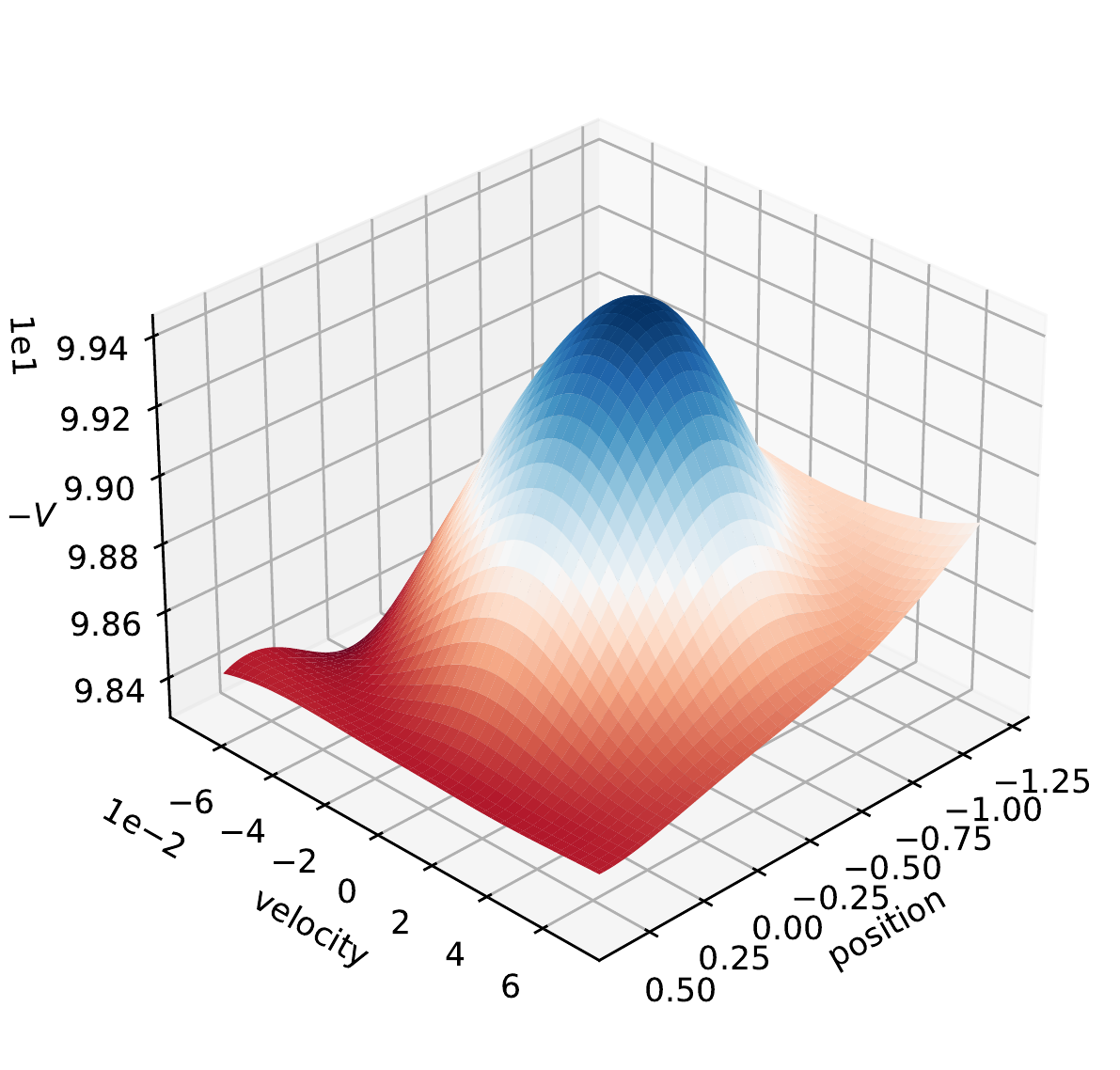}
    \caption{$ N = 500 $}
    \label{fig:V_net_N_500}
  \end{subfigure}%
  \begin{subfigure}{0.33\linewidth}
    \centering
    \includegraphics[width=1.0\linewidth]{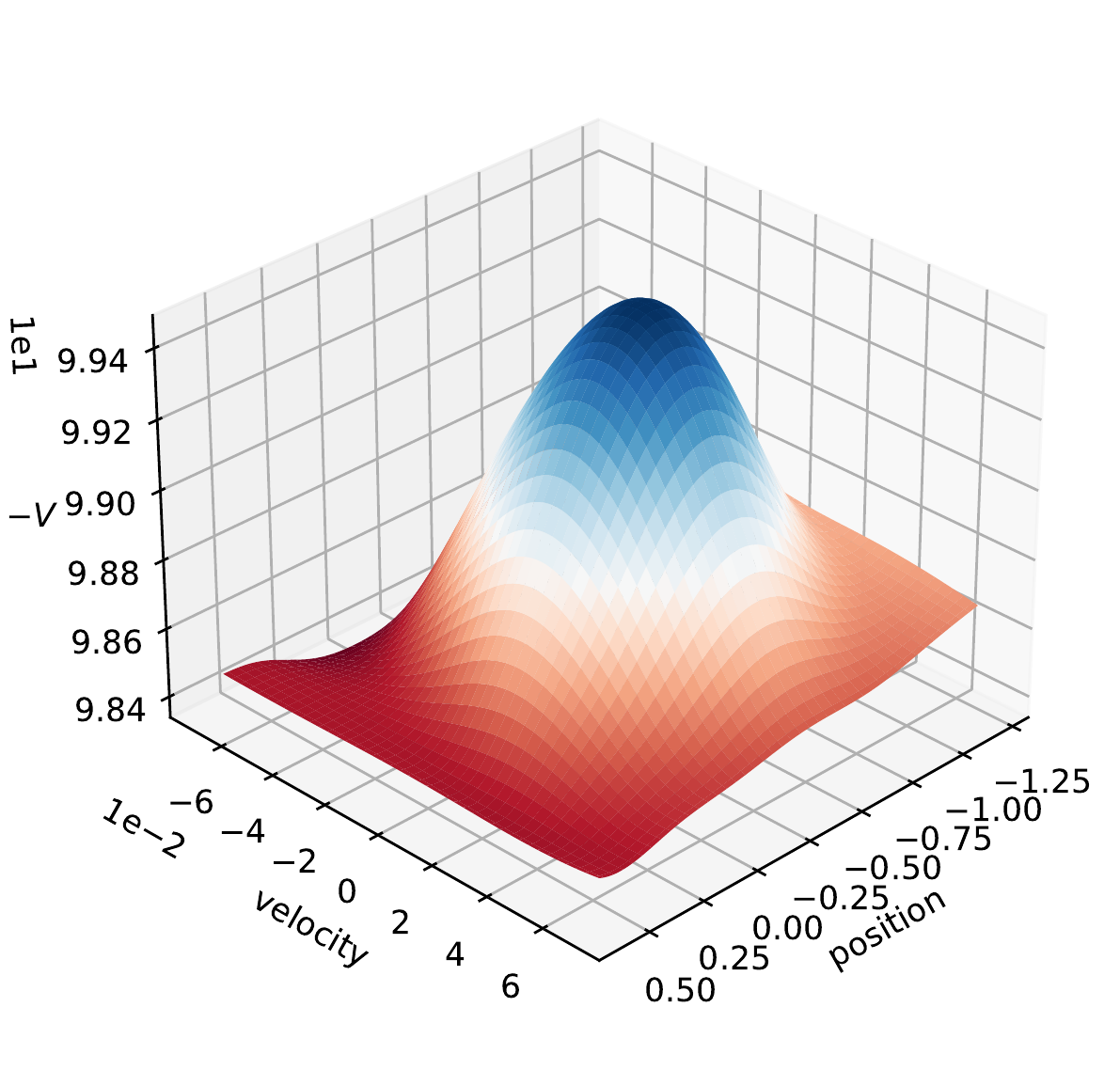}
    \caption{$ N = 2000 $}
    \label{fig:V_net_N_2000}
  \end{subfigure}
  \caption{
  Approximated value functions for different batch sizes $ N $ with minimal test NMSBE, evaluated
  at the same states as for the MC version.}
  \label{fig:V_net_N}
\end{figure}

When evaluating the approximated value function with smallest test error for $ N = 25 $ samples
visually, i.e., see \cref{fig:V_net_N_25}, one can spot a plateau located at around $ -100 $
expected discounted reward.
This is exactly the solution to Bellman's equation or, more precisely, to the loss as defined in
\cref{eq:nmsbe_continuous} if every transition would yield $ -1 $ reward.
We conclude that those scarce samples do not allow the reward information to flow and hence we
have solved implicitly a different MDP.
By comparing the ground truth value function as shown in \cref{fig:V_net_MC}, which we obtain by
running rollouts from every grid state in the test set, to other value functions learned with
different sample sizes as shown in \crefrange{fig:V_net_N_50}{fig:V_net_N_2000}, we see that
training with even only $ 50 $ samples starts to fit the shape of the ground truth value function
in the correct range.
Hence, this experiment suggests that for problems with a continuous state space, NN-VFA methods
can still perform well with a relatively small number of sampled interactions.

\subsubsection{Various Architectures}
\label{sssec:experiment_gene_various}
\begin{figure}
  \centering
  \begin{subfigure}{0.5\linewidth}
    \centering
    \includegraphics[width=1.0\linewidth]{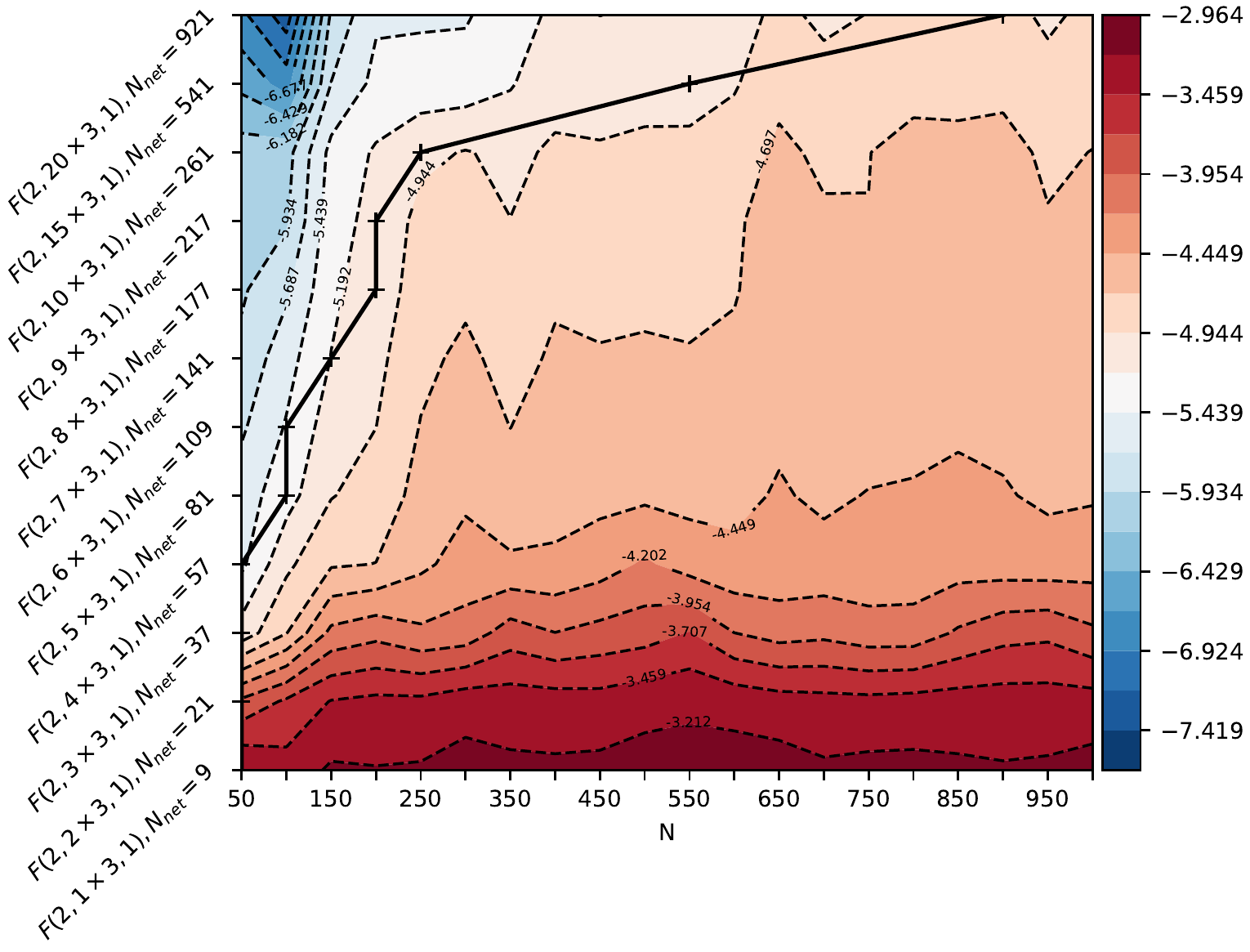}
    \caption{}
    \label{fig:generalisation_train_d3}
  \end{subfigure}%
  \begin{subfigure}{0.5\linewidth}
    \centering
    \includegraphics[width=1.0\linewidth]{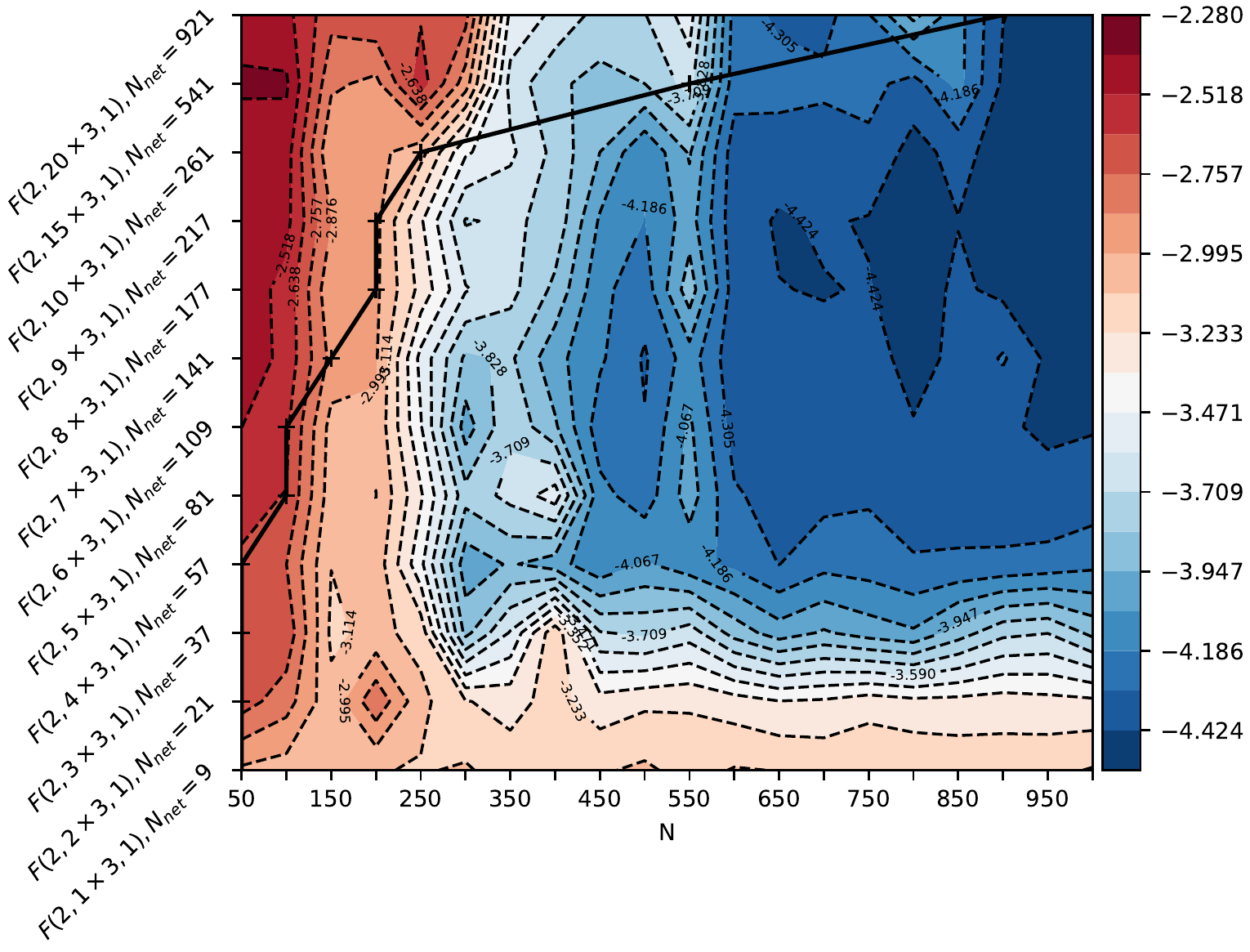}
    \caption{}
    \label{fig:generalisation_test_d3}
  \end{subfigure}

  \begin{subfigure}{0.5\linewidth}
    \centering
    \includegraphics[width=1.0\linewidth]{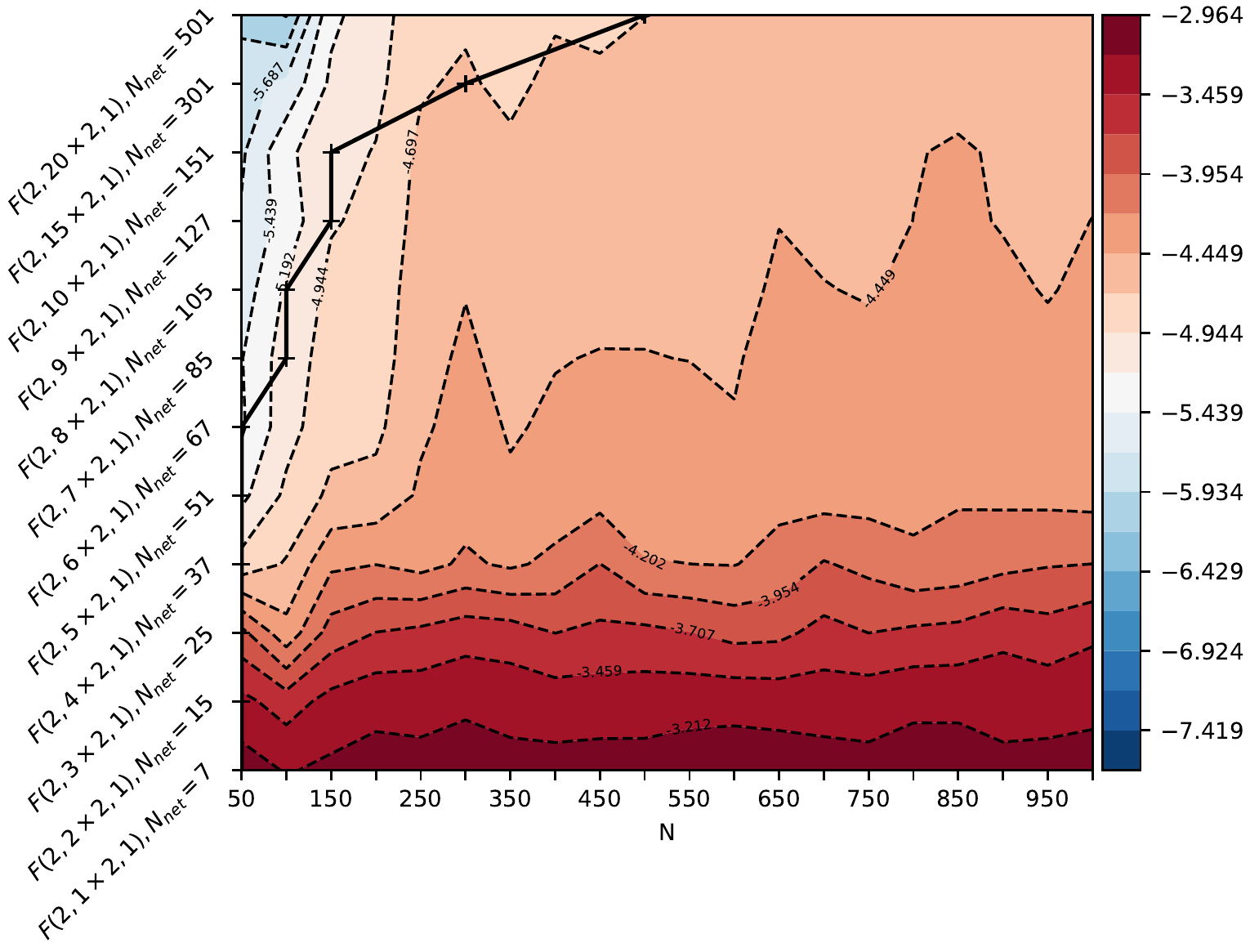}
    \caption{}
    \label{fig:generalisation_train_d2}
  \end{subfigure}%
  \begin{subfigure}{0.5\linewidth}
    \centering
    \includegraphics[width=1.0\linewidth]{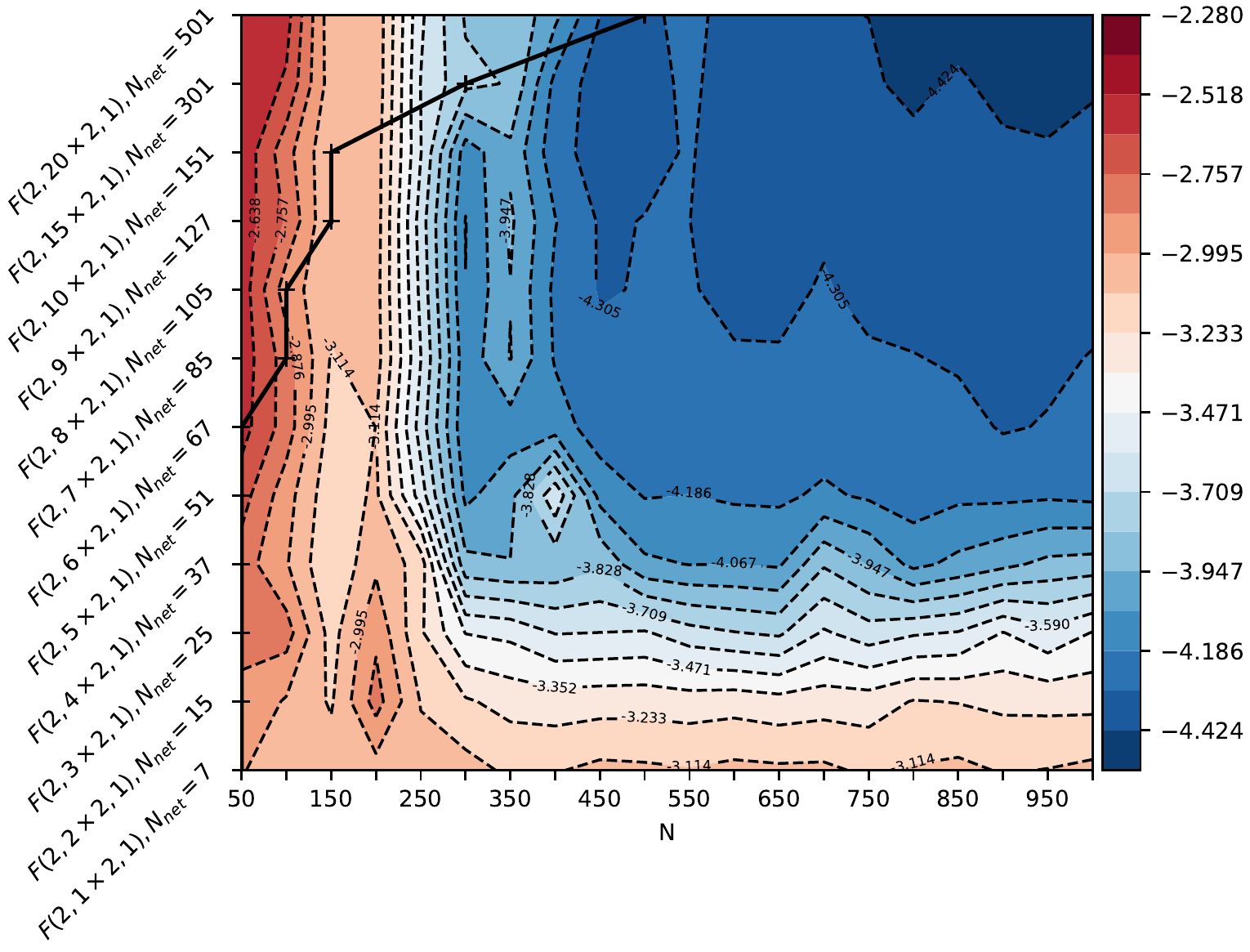}
    \caption{}
    \label{fig:generalisation_test_d2}
  \end{subfigure}

  \begin{subfigure}{0.5\linewidth}
    \centering
    \includegraphics[width=1.0\linewidth]{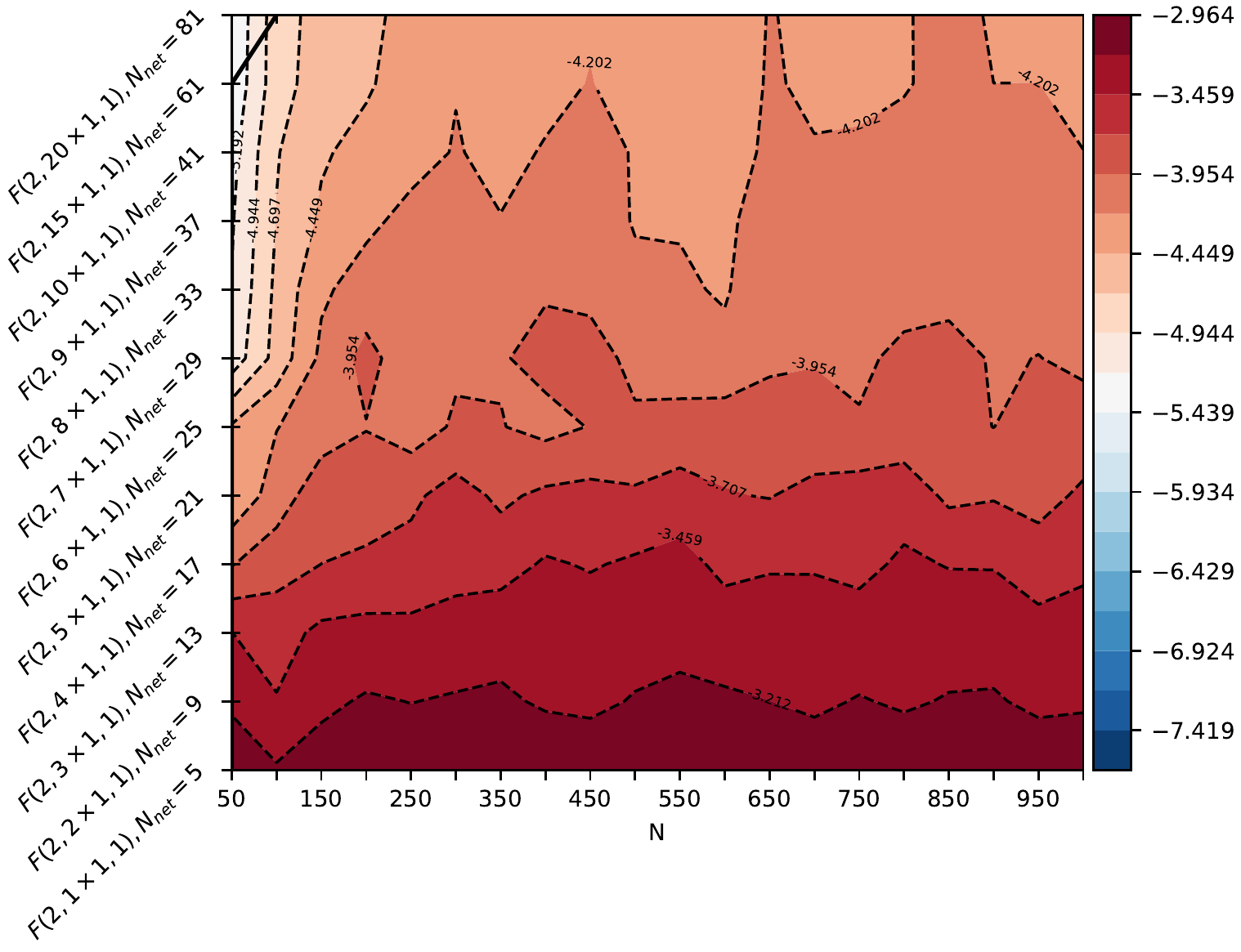}
    \caption{}
    \label{fig:generalisation_train_d1}
  \end{subfigure}%
  \begin{subfigure}{0.5\linewidth}
    \centering
    \includegraphics[width=1.0\linewidth]{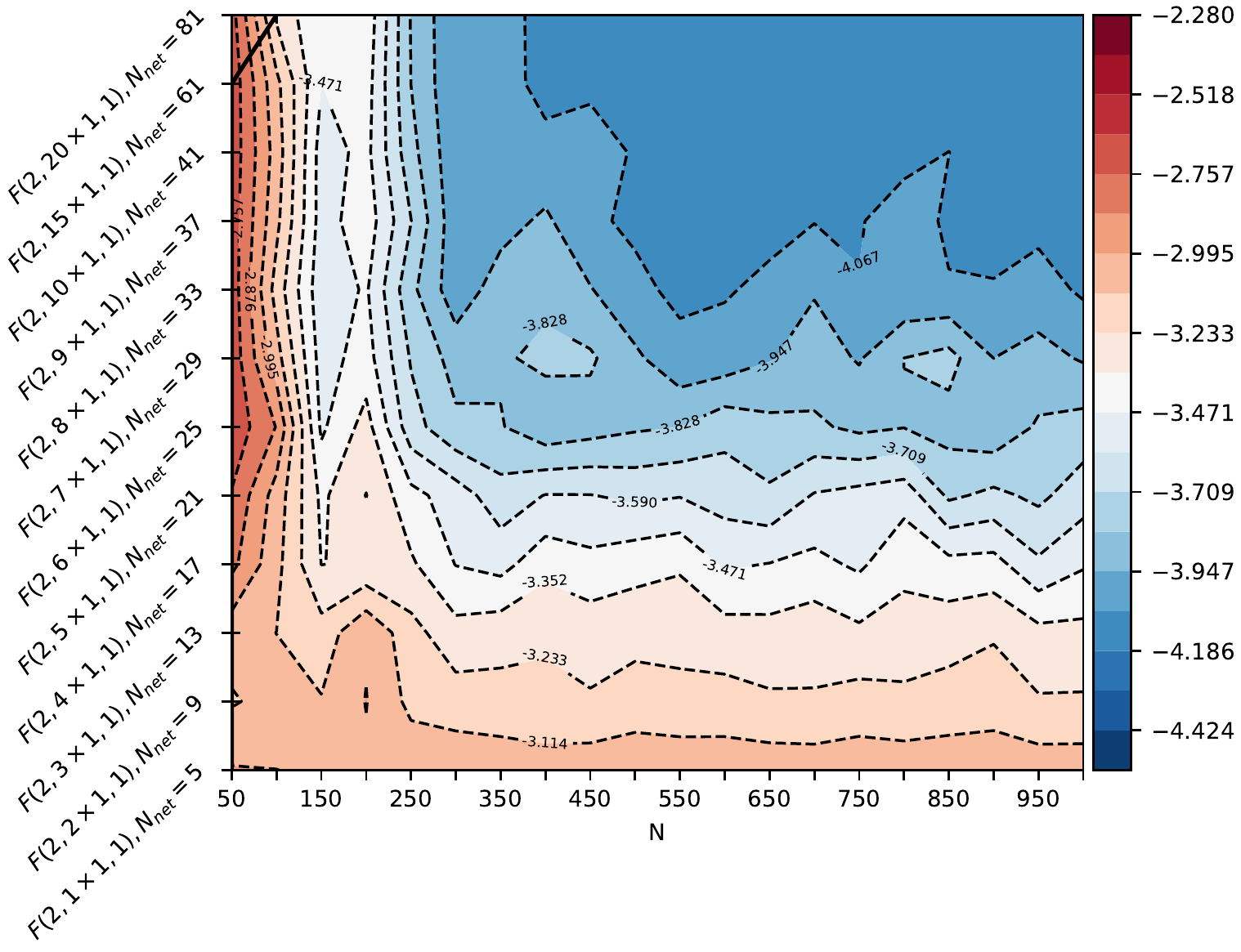}
    \caption{}
    \label{fig:generalisation_test_d1}
  \end{subfigure}
  \caption{
  The training and test error of different MLP architectures (ordinate) for various
  sample sizes $ N $ (abscissa).
  We use a logarithmic scale $ Z = \log_{10}(E) $, where $ E $ is the original error and $ Z $ its
  plotted value.
  Red indicates higher errors.
  In all plots the solid line represents the condition $ \Nnet = N $.
  \textbf{Left column:} Training error. \textbf{Right column:} Test error.\\
  For training errors one can see that shape and position of isolines follow roughly the condition
  $ \Nnet = N $.
  Lower training error as well as the case of exact fitting happens to the upper left of the
  black solid line, confirming our considerations from \cref{sec:theory}.
  The testing errors reveal that, at least for MLPs with a large number of parameters, the region
  with smallest test error reaches close to the condition $ \Nnet = N $.
  However, for all networks the well-known rule \enquote{the more data, the better performance}
  applies, presumable because not all sample states are far apart.}
  \label{fig:generalisation}
\end{figure}

\paragraph{Setting}
It is widely believed that the architecture of an MLP has an important influence on its
generalisation performance, but the exact impact is still unclear.
Therefore, we train several MLPs with different architectures and a varying number of samples in the
same scenario as before.
We refer to the axis labels in \cref{fig:generalisation} for concrete choices of architectures
and the values of $ N $.
Due to the huge amount of computation involved in deeper networks with large sample sizes, we reduce
the number of repetitions from $ 25 $ to ten.
We provide the mean test and training error as contour plots in \cref{fig:generalisation} with
logarithmic scale ($ Z = \log_{10}( E ) $ where $ E $ is the actual error and $ Z $ its plotted
value).
This additional processing step is required to reveal the detailed structure of the surface.

\paragraph{Results}
In the following, we look separately at training and testing errors and start with the training
errors, i.e., the contour plots in the left column of \cref{fig:generalisation}.
%
%
We see that the shapes of isolines for all MLPs but the smallest ones match the shape of the line
$ \Nnet \approx N  $, which we call \emph{condition line} in the following.
We obtain the condition line by rounding the amount of parameters $ \Nnet $ to the closest value
of $ N $ and joining those points.
The shape implies that the area with same training error follows the condition in
\cref{eq:overparam_continuous} from our theoretical investigation.
When we increase the amount of data, we also have to increase the depth or width of an MLP, and vice
versa, to maximise the chance of avoiding suboptimal minima.
%
%
Between \cref{fig:generalisation_train_d2,fig:generalisation_train_d3} the shapes of isolines stay
consistent and furthermore, the condition line is present at roughly the same training error
of $ -4.994 $ to $ -4.697 $.
%
%
Exact learning of all samples indicated by training errors close to zero happens only above the
solid black line, i.e., whenever the network has more parameters than the number of training
samples.

Next, we address the testing error as shown in the right column of \cref{fig:generalisation}.
%
%
For larger MLPs, where the condition $ \Nnet \approx N $ is available for large values of $ N $, we
observe that the region with smallest test error extends always to the condition line.
This happens e.g. for the MLP $ \MLPwd{2}{15}{3}{1} $ at $ N = 500 $ or for $ \MLPwd{2}{20}{3}{1} $
at $ N = 900 $.
We see that larger MLPs work more predictable where smaller MLPs can achieve the smallest test
error for several training set sizes $ N $ with higher errors in between.
In other words, large MLPs do not necessarily increase the threshold of required samples for good
performance in our experiment.
Small MLPs need samples in a similar scale as the largest MLPs to achieve the smallest test error.
Even those MLPs, which one would consider as tiny such as $ \MLPwd{2}{6}{3}{1} $, also achieve the
smallest test error with the same amount of samples.
Thus, if for small MLPs the condition $ \Nnet > N $ is far from being realistic, the well-known
statement \enquote{the more data, the better performance} applies.
We need a factor of ten more training data than adjustable parameters.
In summary, large MLPs concentrate the region with extreme values for the test errors and amplify
the effect of the sample size.
For $ N \to 50 $, the largest architectures produces out of all runs the highest test error.
Hence, by shrinking the amount of parameters in an MLP such that $ \Nnet \approx N $ applies again
one can reduce the test error without changing the amount of data.
If $ N \to 1000 $, one has to employ MLPs with a comparable amount of parameters such that the area
with smallest test error is present.
These MLPs then possess highest generalisation performance out of all architectures.
%

\subsection{Policy Iteration}
\label{sec:policy_iteration}
\paragraph{Setting}
For the final experiment, we change the environment to Cart Pole and test our approach in a Policy
Iteration setting.
For that purpose, we use $ Q $-factors instead of $ V_{\pi}(s) $ and use a greedily induced policy
as defined in \cref{eq:def_gip}.
To represent $ Q $-factors, the network input consists of the continuous state and the discrete
action index, hence MLPs must have $ K + 1 $ units in the input layer.
More specifically, we use the MLP $ \Fcal(5, 10, 10, 1) $ with $ \Nnet = 181 $ parameters to
approximate the $ Q $-function.

We test different sample sizes $ N \in \left\{ 100, 181, 300, 500 \right\} $, i.e., the number of
states sampled before every Policy Evaluation, and investigate different amounts of policy
evaluation steps $ i \in \left\{ 500, 1500, 2500, 3500, 4500 \right\} $, i.e., the number of descent
steps.
Finally, we also compare the effect of reusing the last parameters of the previous Policy
Evaluation step as initialisation for the current one and refer to this by \emph{transient} and
\emph{persistent} as in \citep{sigaud:nn19}.

\begin{figure}
  \centering
  \begin{subfigure}{0.25\linewidth}
    \includegraphics[width=1.0\linewidth]{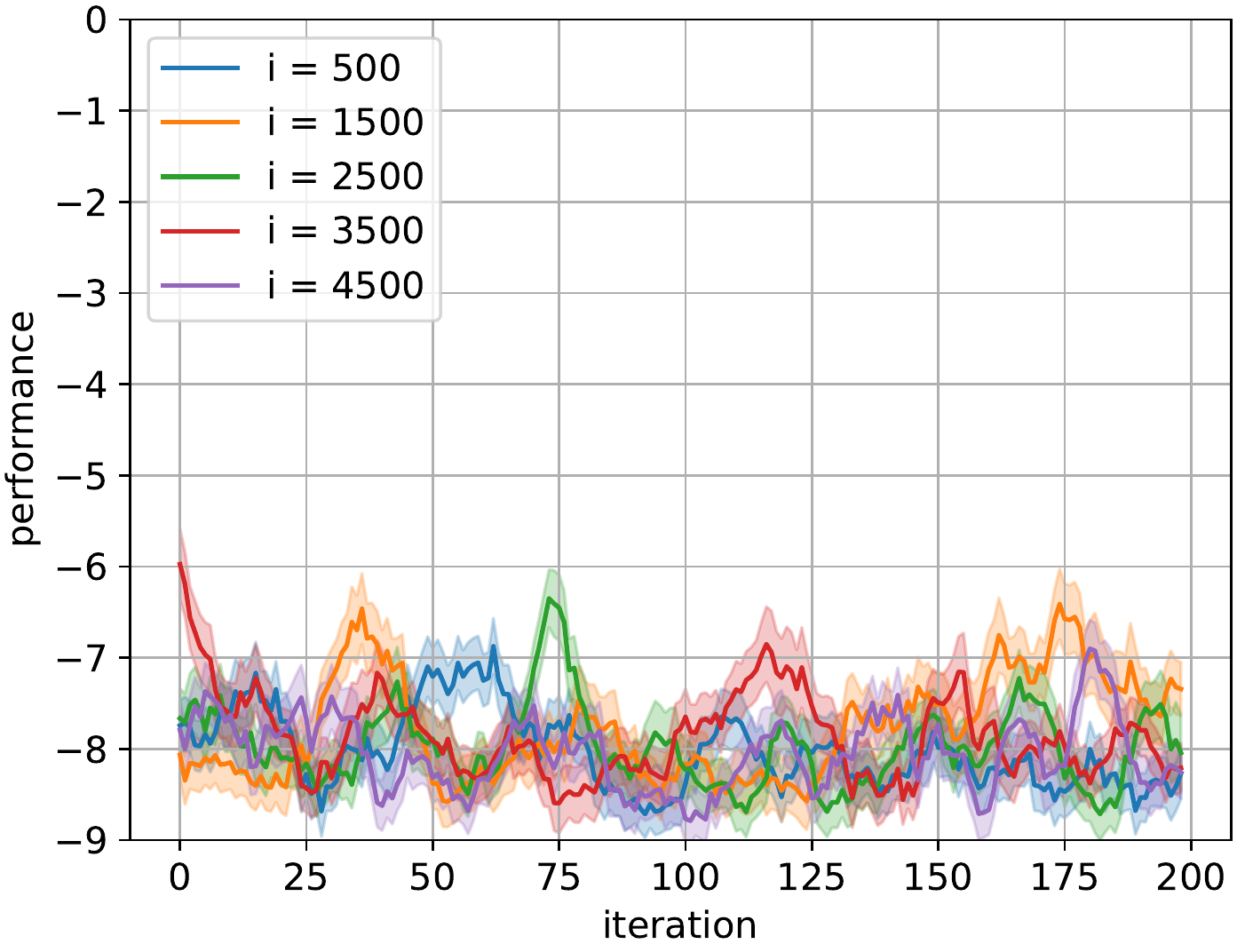}
    \caption{$ N = 100 $}
    \label{fig:pi_cold_N_100}
  \end{subfigure}%
  \begin{subfigure}{0.25\linewidth}
    \includegraphics[width=1.0\linewidth]{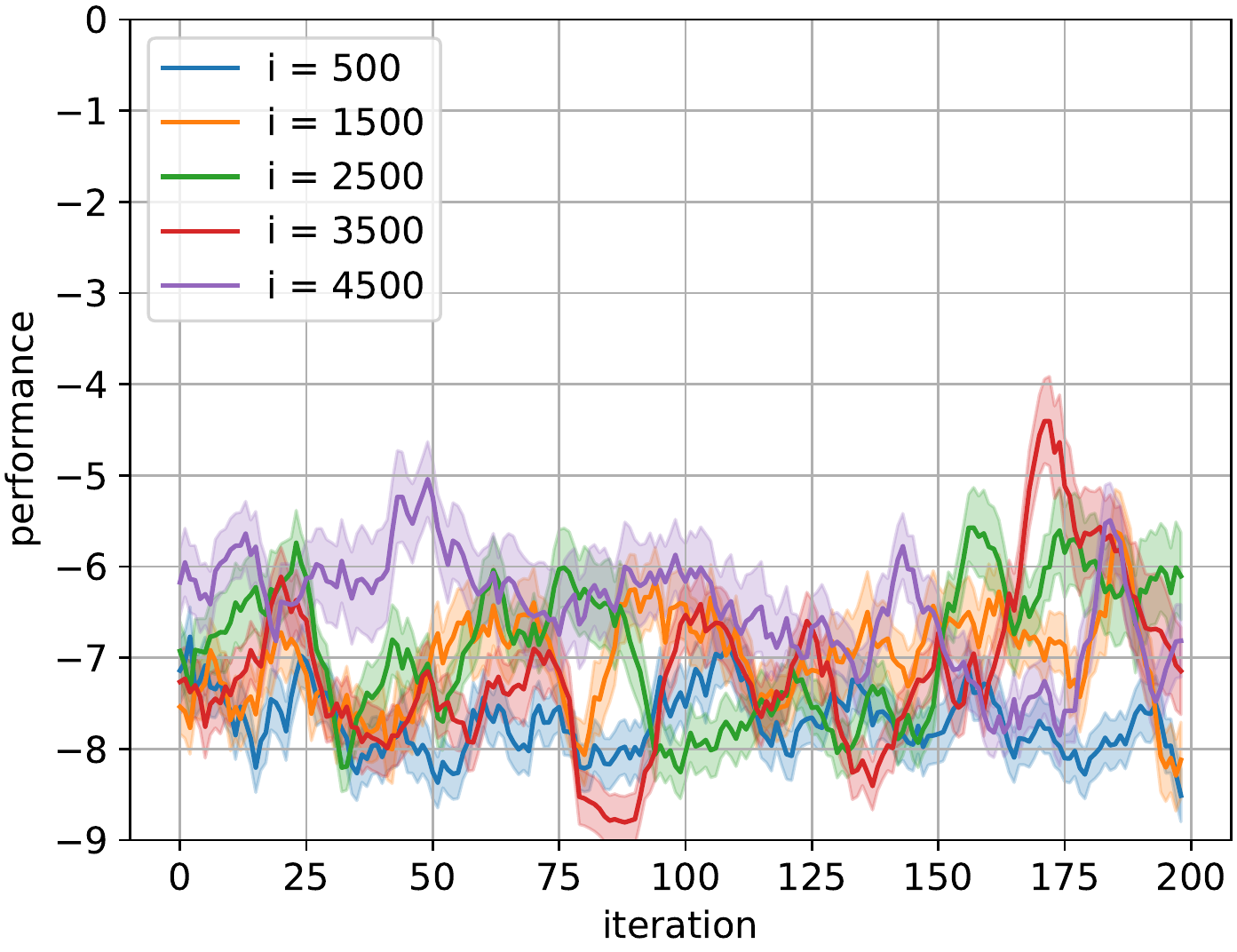}
    \caption{$ N = 181 $}
    \label{fig:pi_cold_N_181}
  \end{subfigure}%
  \begin{subfigure}{0.25\linewidth}
    \includegraphics[width=1.0\linewidth]{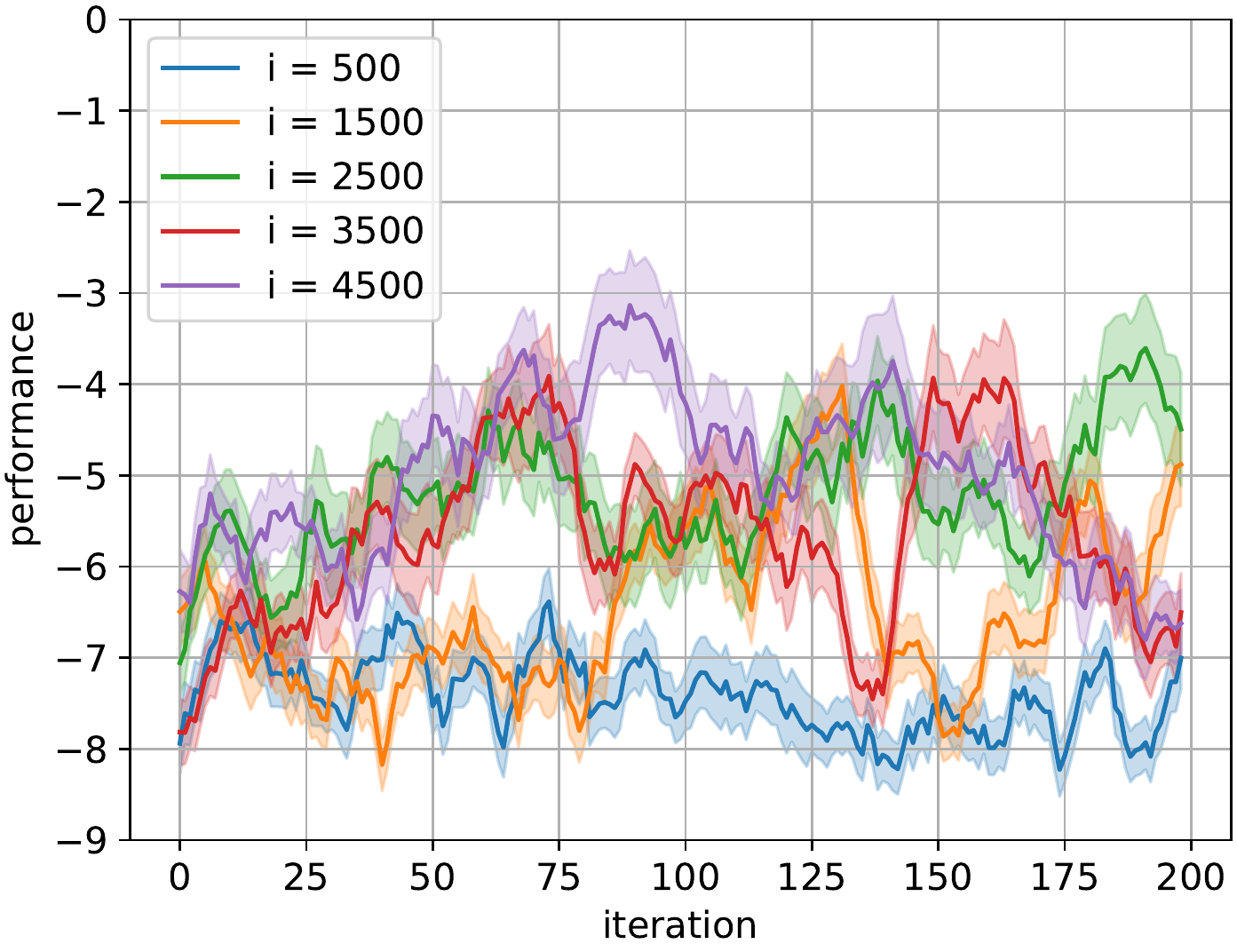}
    \caption{$ N = 300 $}
    \label{fig:pi_cold_N_300}
  \end{subfigure}%
  \begin{subfigure}{0.25\linewidth}
    \includegraphics[width=1.0\linewidth]{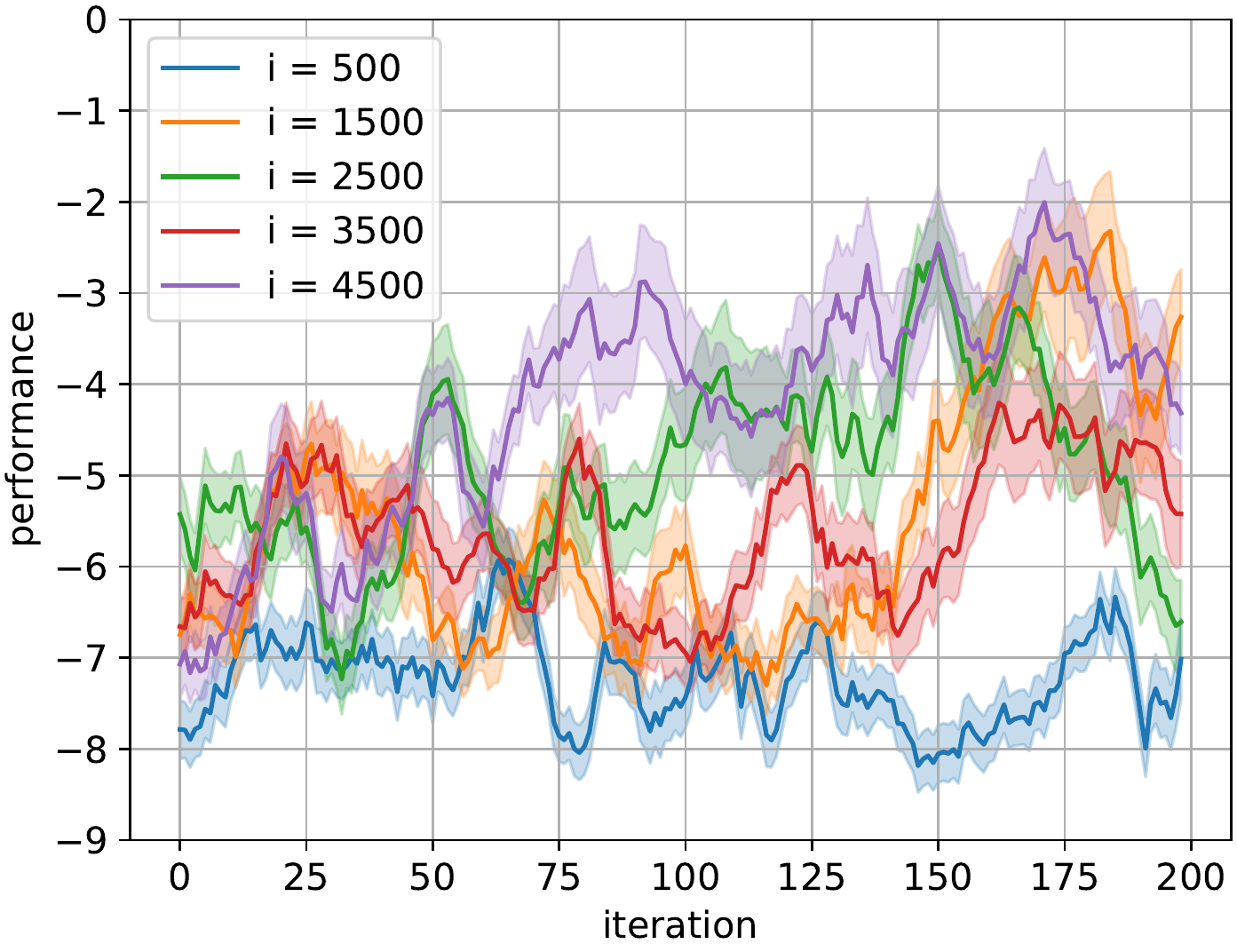}
    \caption{$ N = 500 $}
    \label{fig:pi_cold_N_500}
  \end{subfigure}

  \begin{subfigure}{0.25\linewidth}
    \includegraphics[width=1.0\linewidth]{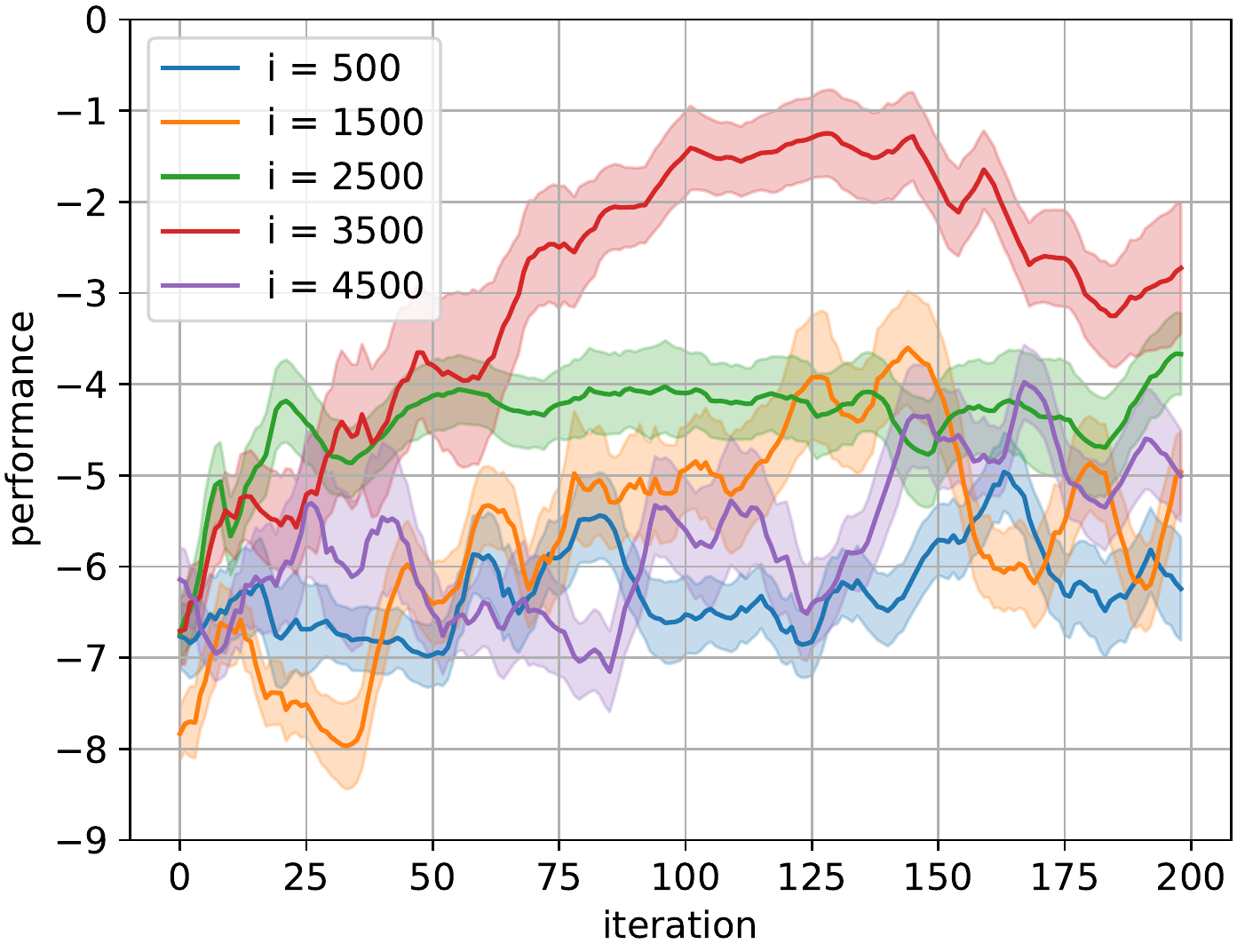}
    \caption{$ N = 100 $}
    \label{fig:pi_warm_N_100}
  \end{subfigure}%
  \begin{subfigure}{0.25\linewidth}
    \includegraphics[width=1.0\linewidth]{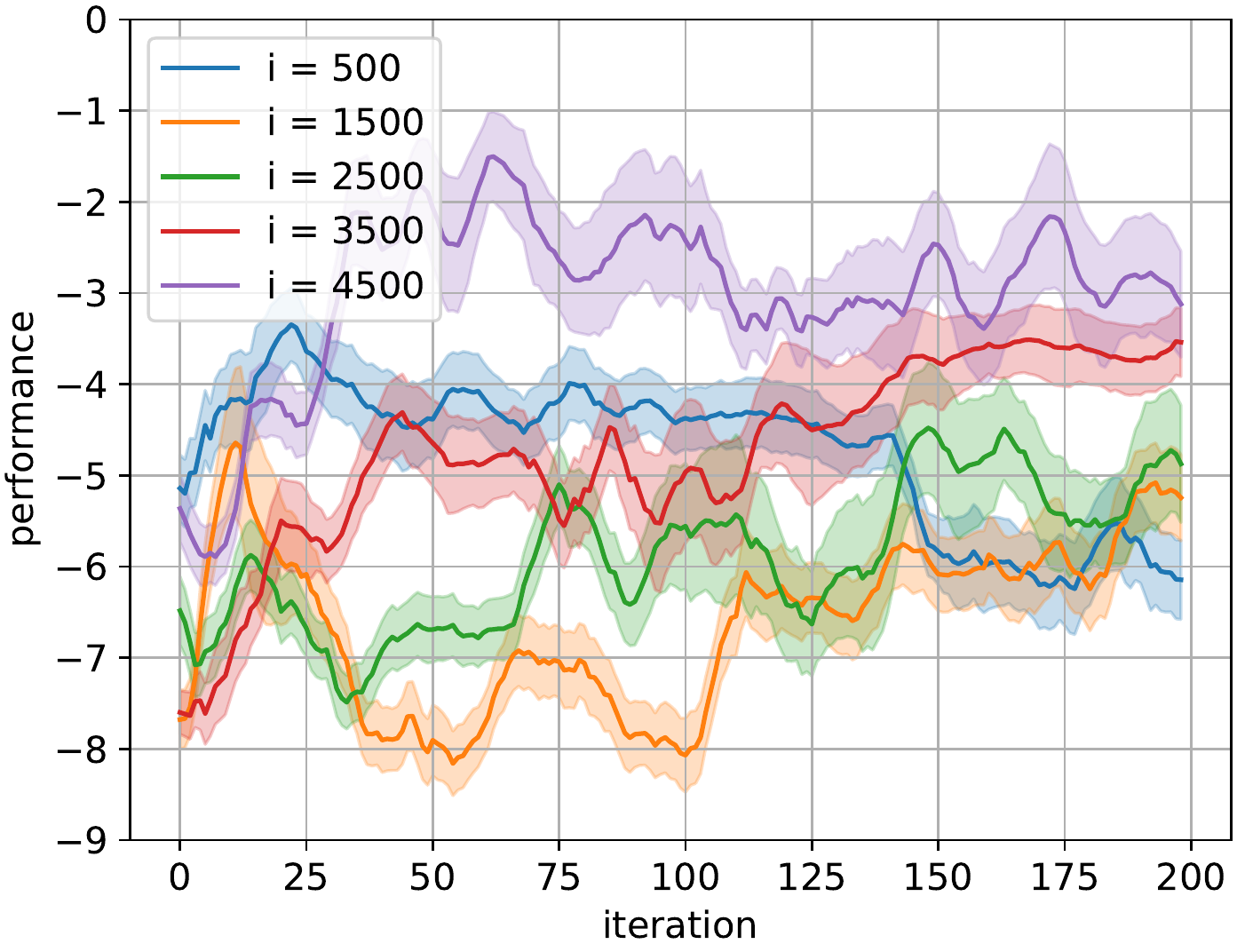}
    \caption{$ N = 181 $}
    \label{fig:pi_warm_N_181}
  \end{subfigure}%
  \begin{subfigure}{0.25\linewidth}
    \includegraphics[width=1.0\linewidth]{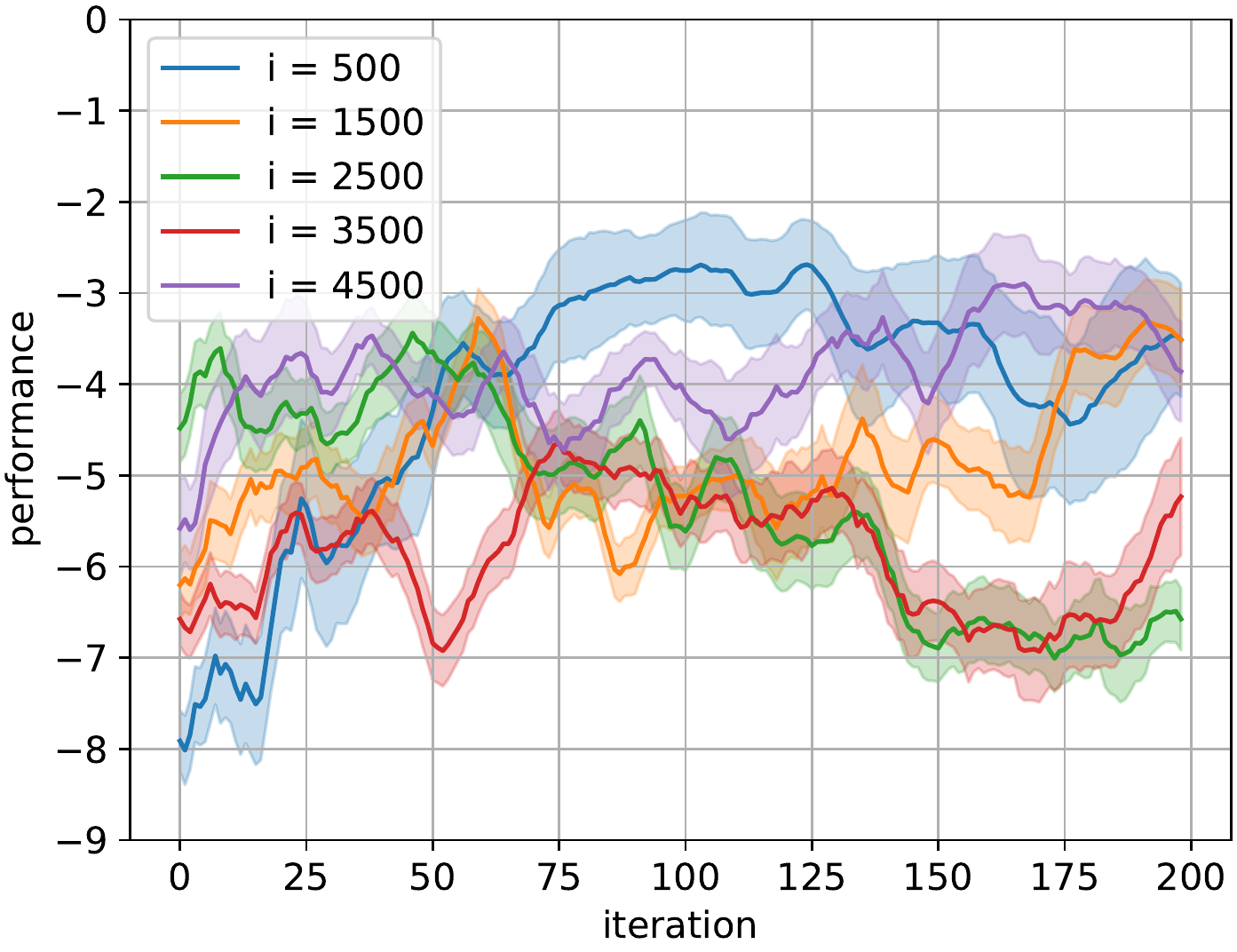}
    \caption{$ N = 300 $}
    \label{fig:pi_warm_N_300}
  \end{subfigure}%
  \begin{subfigure}{0.25\linewidth}
    \includegraphics[width=1.0\linewidth]{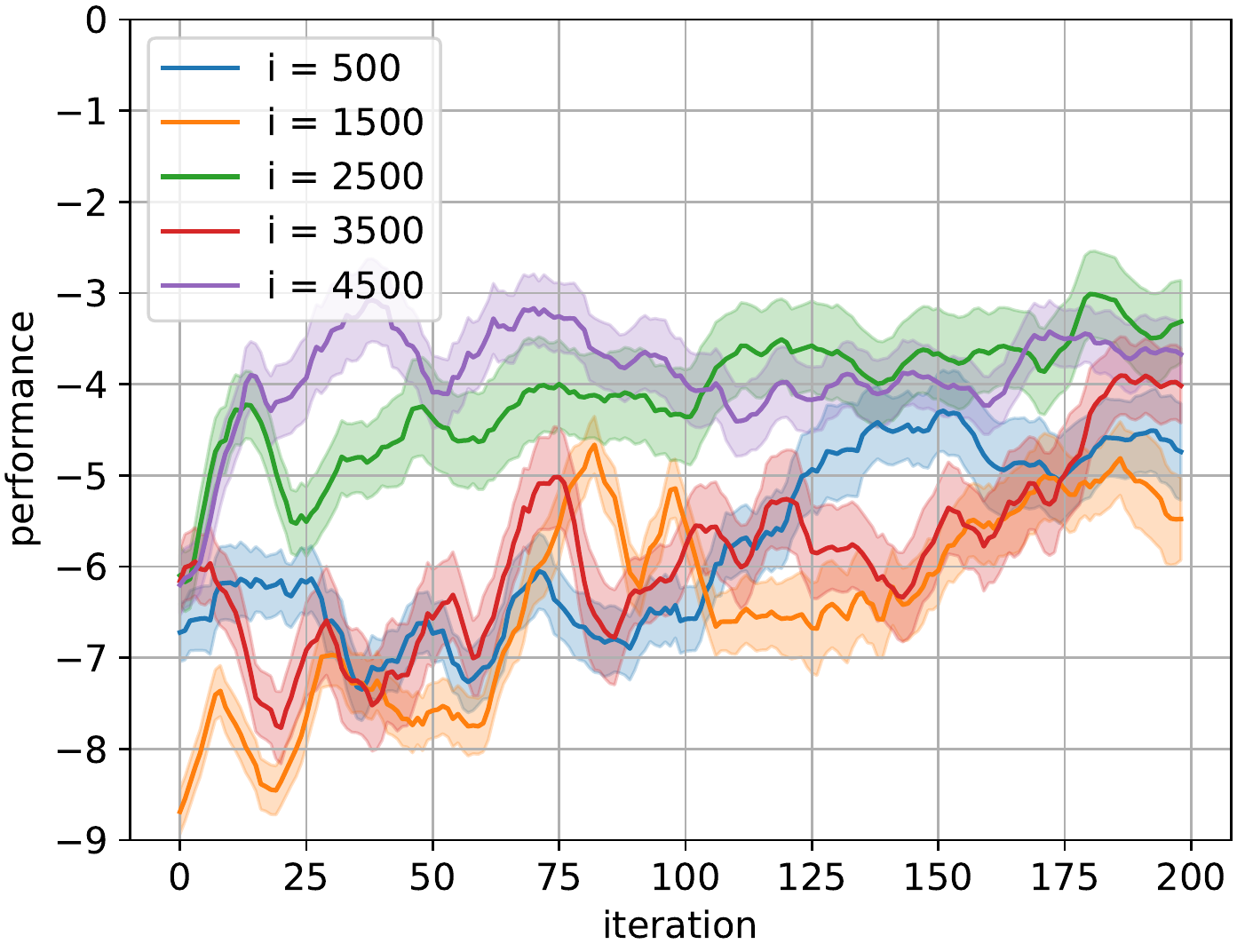}
    \caption{$ N = 500 $}
    \label{fig:pi_warm_N_500}
  \end{subfigure}

  \caption{
  Combining the Residual Gradient Gauss Newton Policy Evaluation with $ Q $-factors and GIP
  policies to
  full Policy Iteration.
  As environment we use \emph{Cart Pole} and plot the expected reward obtained by performing
  roll-outs from $ 5 $ repetitions.
  From left to right we vary the number of sampled states.
  In each Figure the number of Policy Evaluation iterations is changed.
  In the top row Policy Evaluation starts with a new MLP in every sweep.
  For the bottom row we reuse the last evaluation of the previous sweep as initialisation for the
  current one.\\
  Sequentially improving policies are visible, especially when reusing the MLP, however, there is
  no clear trend for the required number of samples or the optimal amount of Policy Evaluation
  iterations.
  Without second-order optimisation or when using Semi-Gradients Policy Iteration did not work at
  all.}
  \label{fig:policy_iteration}
\end{figure}

We measure the performance of a policy by performing ten rollouts with each $ 500 $ steps after
every improvement step and combine the rollouts as mean, minimal and maximal value.
Furthermore, we repeat every experiment five times such that we can provide averages of expected
returns at each sweep as a summarising impression of the Policy Iteration processes.
Shaded areas represent the mean of all individual spreads of discounted returns.
Our results are available in \cref{fig:policy_iteration}.

\paragraph{Results}
%
%
Since experiments with Semi-Gradients diverged for both first and second order descent algorithms
and a first-order Residual Gradient formulation did not show improving policies over time, we only
provide results for the Gauss Newton Residual Gradient formulation.
Thus, this experiment already implies that second-order information is essential to enable and
stabilise Policy Iteration in continuous problems.
It also shows that in order to get the benefits of second-order approaches, the TD target cannot be
ignored during the computation of differential maps.

%
%
Sequentially improving policies are visible in most plots.
Starting with a fresh MLP for each evaluation increases the number of sample states to see a more
pronounced improvement over time.
When using persistent evaluations, i.e., reusing the MLP parameters corresponding to the last
policy, clear improvements are visible also for small sample sizes.
Furthermore, the maximum achieved performance is higher than for transient sweeps.

%
%
In all plots of \cref{fig:policy_iteration}, it is hard to find a clear trend for the required
number of samples or the optimal amount of Policy Evaluation iterations.
As an example, in \cref{fig:pi_cold_N_500} both $ i = 1500 $ and $ i = 4500 $ have a similar high
performance around sweep $ 175 $.
In \cref{fig:pi_warm_N_100}, using only $ N = 100 $ samples result for $ i = 3500 $ reliably in
significant better performing policies than when increasing the amount of Policy Evaluation steps
to~$ i = 4500 $.

%
%
The fact that for $ N = 181 = \Nnet $ samples and when using persistent evaluations of policies all
repetitions of the experiment come up with good performing policies for the largest amount of
Policy Evaluation steps is worth further investigation.
But based on our currently available experimental results we cannot make reliable statements at
this point.
%
%
Although we observe a slightly chaotic behaviour, we argue that this is to be expected,
since function approximation is utilised in a Policy Iteration framework.
Thus, there are two unavoidable sources of error, namely inaccurate Policy Evaluation and erroneous
Policy Improvement.
First, different than typical settings where samples come from an exploration mechanism, we select
samples uniformly distributed in the whole state space.
Thus, our optimisation problem does not have a high resolution focused on visited parts of the
state space but tries to find a global solution, which is obviously a more challenging problem.
Second, we make use of discrete actions.
This means, we ask a smooth function approximation architecture to model jumps in the value
function, which must occur for example around the balancing point of the pole.

%
\section{Conclusion}
\label{sec:conclusion}
%
%
RL with NN-VFA has become one of the most powerful RL paradigms in both research and
application in the recent years.
Despite the superior performance, its training and convergence analysis remains challenging due to
an incomplete theoretical understanding how MLPs affect the common RL setting.
In particular, most previous theoretical analysis of this problem approaches the challenge from the
perspective of minimising Mean Squared Projected Bellman Error which is incompatible with recent
successful applications.

%
%
This work bridges the gap with a concise critical point analysis of the NMSBE when using MLPs.
We address both the discrete and continuous state space setting for a Residual Gradient
formulation, i.e., using the complete gradient of the NMSBE.
We derive conditions on MLPs to ensure a proper behaviour of the optimisation procedure.
Over-parametrisation of the MLP is required next to some design principles for MLPs to eliminate
suboptimal local minima.
Furthermore, full rankness of the differential maps of the MLP enable pleasing convergence
properties of gradient descent algorithms.
Our analysis unveils the possibility to utilise approximated second-order information
of the cost function, resulting in an efficient Approximate Newton's method, namely a Gauss Newton
Residual Gradient algorithm.
As part of our work, we also see, why multistep lookahead can help Semi-Gradient algorithms.
As the MLP in the TD target gets multiplied with the discount factor with larger powers the effect
of ignoring the dependency vanishes naturally.
Furthermore, we point out a source of error when using sampling based approximations to the NMSBE.
Critical points also contain solutions with zero NMSBE, which not necessarily correspond to good
approximations of the value function.
The optimisation problem can have undesired degrees of freedom.

%
%
In several experiments, we investigate empirically the minimisation of the NMSBE using a Gauss
Newton Residual Gradient algorithm.
First, we ensure the correctness of the approximated Hessian close to critical points by
demonstrating quadratic convergence on an adapted version of Baird's Seven State Star Problem.

Next, in continuous state space problems, we provide consistent and stable convergence properties
of Residual Gradient algorithms combined with second-order information for a wide range of learning
rates.
Semi-Gradient algorithms are observed to diverge except for some learning rates.
Carefully selecting the learning rate is thus important.
First-order only Residual Gradients are shown to converge slowly, confirming the known behaviour
which explains their unpopularity.
Using an approximated Hessian can solve these issues and is essential to develop stable and
efficient RL algorithms, which also achieve better final errors.
Our experiments also serve as proof of concept that with modern computer systems second-order
optimisation is a possible approach in Reinforcement Learning applications.

Furthermore, we investigate the generalisation capabilities of MLPs when training with approximated
second-order Residual Gradient algorithms.
Training and test errors follow our theoretical insights, meaning that the number of parameters
should be synchronised with the number of training samples.
We find that deeper architectures do not necessarily increase the number of samples required for
good performance, they rather amplify the extreme cases for errors.

Finally, we demonstrate that the application of Residual Gradient methods can work in a full
Policy Iteration setting.
A well performing policy can be learned by alternating between Policy Evaluation and Policy
Improvement, starting from a random one.
However, our empirical findings did not reveal trends on the number of samples or Policy Evaluation
steps needed.
Our hypothesis is that the discrete action space stands in conflict with the capabilities of a
smooth function approximators such as MLPs.
Solving the RL task involves representing jumps in the value function, which might not be possible.
Hence, including a continuous action space in our analysis is an important next step for our future
work.

%
%
In conclusion, considering derivatives of the TD-target as done in Residual Gradient algorithms
allows us to perform a sophisticated analysis and also provides the foundation for reliable NL-VFA
algorithms.
Smooth optimisation is a promising methodology to answer open issues in Deep RL.
It unveils important details for the construction of efficient algorithms and outlines difficulties
in the formulation of the optimisation task.
%
%
Hence, next steps for future work are to extend this approach to policy gradient methods and the
actor-critic frameworks to allow for continuous action spaces in the analysis.

\section*{Acknowledgements}
Supported by Deutsche Forschungsgemeinschaft (DFG) through TUM International Graduate School of
Science and Engineering (IGSSE), GSC 81.

\newpage

\begin{small}
  \bibliographystyle{plainnat}
  \bibliography{../analysis_optimisation_bellman_error}
\end{small}

\newpage

\appendix

%
\section{Step by Step Calculations}
\label{sec:step-by-step}
%
%
\paragraph{Differential map of error function}
For the error function $ E \colon \Rds^K \to \Rds $ and some $ F \in \Rds^K $ we have
\begin{align*}
E(F) =& \frac12
        \Big( F - P_\pi \big( R_\pi + \gamma F \big) \Big)^\top
        \Xi
        \Big( F - P_\pi \big( R_\pi + \gamma F \big) \Big)\\
\D E(F)[h]
  =& \frac12 \left. \frac{d}{dt} \right|_{t=0}
     \left[\Big( (F + th) - P_\pi \big( R_\pi + \gamma (F + th)\big) \Big)^\top
     \Xi
     \Big( (F + th) - P_\pi \big( R_\pi + \gamma (F + th)\big) \Big) \right]\\
  =& \frac12 \Big[
     \Big( h - P_\pi \gamma h \Big)^\top \Xi
     \Big( (F + th) - P_\pi \big( R_\pi + \gamma (F + th)\big) \Big)\\
  &+ \Big( (F + th) - P_\pi \big( R_\pi + \gamma (F + th)\big) \Big)^\top \Xi
     \Big( h - P_\pi \gamma h \Big)
     \Big]_{t=0}\\
  =& \frac12 \left[
     \Big( h - P_\pi \gamma h \Big)^\top \Xi
     \Big( F - P_\pi \big( R_\pi + \gamma F \big) \Big) +
     \Big( F - P_\pi \big( R_\pi + \gamma F \big) \Big)^\top \Xi
     \Big( h - P_\pi \gamma h \Big)
     \right]\\
  =& \Big( F - P_\pi \big( R + \gamma F \big) \Big)^\top \Xi
     \Big( I_K - \gamma P_\pi \Big) h.
\end{align*}
Thus, according to Riesz, the gradient is
\begin{equation*}
  \nabla_F E(F) =
  \left(\left(F - P_\pi(R_\pi + \gamma F)\right)^\top \Xi (I_K - \gamma P_\pi) \right)^\top,
\end{equation*}

\paragraph{Differential map of MLP}
Consider an MLP $ f \in \MLP{n_0, n_1}{n_{L-1}, n_L} $ and an input $ s \in \Scal $.
To calculate the differential map of $ f $ for $ s $ at the point $ \Wbf \in \Wcal $ and a
direction $ \Hbf \in \Wcal$ first start with a single layer $ l $ of the MLP.
We have
\begin{equation*}
  \D_{W_l} f(\Wbf, s) [H_l] =
  \D_2 \Lambda_L(W_L,\phi_{L-1}) \circ \ldots \circ \D_2 \Lambda_{l+1}(W_{l+1},\phi_l) \circ
  \D_1 \Lambda_l(W_l,\phi_{l-1}) [H_l],
\end{equation*}
where $ \D_1 \Lambda_l(W_l,\phi_{l-1})[H_l] $ and $ \D_2 \Lambda_l(W_l,\phi_{l-1})[h_{l-1}] $
refer to the derivative of layer mapping $ \Lambda_l $ with respect to the first and the second
argument, respectively.
For the layer definition in \cref{eq:layer_map} we obtain
\begin{align*}
  \D_1 \Lambda_l(W_l,\phi_{l-1})[H_l]
  &= \left. \frac{d}{dt} \right|_{t=0}
  \threebyone{\vdots}{\sigma \left( (W_{l,k} + t \cdot H_{l,k} )^\top \cdot \homogen{\phi_{l-1}}
  \right)}{\vdots} = \threebyone{\vdots}{\dot \sigma (\cdots) H_{l,k}^\top
  \homogen{\phi_{l-1}}}{\vdots}_{t=0} \\
  &= \diag \left( \dot \phi_l \right) H_l^\top \homogen{\phi_{l-1}}
  \eqqcolon \Sigma_l \cdot H_l^\top \cdot \tilde \phi_{l-1} \\
  \D_2 \Lambda_l(W_l,\phi_{l-1})[h_{l-1}]
  &= \left. \frac{d}{dt} \right|_{t=0}
  \threebyone{\vdots}{\sigma \left(W_{l,k}^\top \cdot  ( \homogen{\phi_{l-1}} + t \cdot
  \homogendir{h_{l-1}} ) \right)}{\vdots}
  = \threebyone{\vdots}{\dot \sigma (\cdots) W_{l,k}^\top \homogendir{h_{l-1}}}{\vdots}_{t=0} \\
  &= \diag \left( \dot \phi_l \right) W_l^\top \homogendir{h_{l-1}}
  \eqqcolon \Sigma_l \cdot \bar W_l^\top \cdot h_{l-1}
\end{align*}
where $ \Sigma_l \in \Rds^{n_l \times n_l} $ is a diagonal matrix with its entries being the
derivatives of the activation function with respect to the input, i.e., $ \dot \phi_l $ containing
$ \dot \sigma(\ldots) $ for all units in layer $ l $.
The input to $ \dot \phi_l $ is the unmodified output $ \phi_{l-1} $ of the truncated MLP.
By writing $ \bar W $ we indicate that the last row is cut off due to the multiplication by
zero and $ \tilde \phi $ shows that the layer output is extended with an additional $ 1 $.
This resembles homogenous coordinates as they are used with the special Euclidean group $ SE(3) $
for computer vision applications.
Inserting these parts yields for the differential map
\begin{equation*}
  \D_{W_l} f(\Wbf, s) [H_l] =
  \Sigma_{L}   \bar W_{L}^\top   \cdot
  \Sigma_{L-1} \bar W_{L-1}^\top \cdots
  \Sigma_{l+1} \bar W_{l+1}^\top \cdot
  \Sigma_{l} H_{l}^\top \tilde \phi_{l-1}.
\end{equation*}
To shorten this expression let us construct a sequence of matrices for all $ l = L-1, \ldots, 1 $ as
\begin{equation*}
  \Psi_l \coloneqq \Sigma_l \bar W_{l+1} \Psi_{l+1} \in \Rds^{n_l \times n_L},
\end{equation*}
with $ \Psi_L \equiv 1 $ due to the activation function in the last layer being the identity
function.
Now we can write compactly
\begin{equation*}
  \D_{W_l} f(\Wbf, s) [H_l] = \Psi_{l}^\top H_{l}^\top \phi_{l-1}.
\end{equation*}
To arrive at the expression shown in the paper consider a matrix $ A \in \Rds^{n \times m} $ and a
compatible column vector $ b \in \Rds^{n \times 1} $.
When denoting by $ A_1, \cdots, A_m $ the $ m $ columns of $ A $, one can show by straightforward
computation the identity
\begin{equation*}
  A^\top \cdot b =
  \threebyone{A_1^\top b}{\vdots}{A_m^\top b} =
  \threebyone{b^\top A_1}{\vdots}{b^\top A_m} =
  \threediag{b^\top}{\ddots}{b^\top} \threebyone{A_1}{\vdots}{A_m} =
  \left(I_{m \times m} \kron b^\top \right) \cdot \flatten(A).
\end{equation*}
By setting $ A = H_l $ and $ b = \phi_{l-1} $, we get
\begin{equation*}
  \D_{W_l} f(\Wbf, s) [H_l] =
  \Psi_{l}^\top \left( I_{n_l} \kron \phi_{l-1}^\top \right) \flatten(H_{l}).
\end{equation*}
Finally, we can combine the expressions for all layers and produce the full differential map with
respect to all parameters
\begin{equation*}
  \D_\Wbf f(\Wbf, s)[\Hbf] =
  \underbrace{
  \onebythree{\Psi_{1}^\top \left( I_{n_1} \kron \phi_{0}
  \right)}{\ldots}{\Psi_{L}^\top \left(
  I_{n_L} \kron \phi_{L-1}\right)}
  }_{\in \Rds^{n_L \times \Nnet}}
  \cdot
  \underbrace{
  \threebyone{\flatten(H_1)}{\vdots}{\flatten(H_L)}
  }_{\in \Rds^{\Nnet \times 1}},
\end{equation*}
where the MLP input $ \phi_{0} $ is just the input $ s $.
For the application in our work we always have $ n_L = 1 $ because the value function maps to a
scalar value.
Using all $ N $ inputs at once we arrive at the expression
$ G(\Wbf) \in \Rds^{N \cdot n_L \times \Nnet} $ as shown in \cref{eq:diff_map_f}
\begin{equation*}
  \D_\Wbf F(\Wbf)[\Hbf] =
  \underbrace{
  \threebythree
  {\Psi_{1}^\top \left( I_{n_1} \kron {\phi^{(1)}_{0}}^\top \right)}
  {\ldots}
  {\Psi_{L}^\top \left( I_{n_L} \kron {\phi^{(1)}_{L-1}}^\top \right)}
  {\vdots}{\ddots}{\vdots}
  {\Psi_{1}^\top \left( I_{n_1} \kron {\phi^{(N)}_{0}}^\top \right)}
  {\ldots}
  {\Psi_{L}^\top \left( I_{n_L} \kron {\phi^{(N)}_{L-1}}^\top \right)}
  }_{\eqqcolon G(\Wbf) \in \Rds^{N \cdot n_L \times \Nnet}} \cdot
  \threebyone{\flatten(H_1)}{\vdots}{\flatten(H_L)}.
\end{equation*}
The superscript $ (\cdot)^{(i)} $ indicates that the layer outputs $ \phi_l $ arise from the $
i $-th state in the input layer.

\paragraph{Definition of $ \widetilde G(\Wbf) $ }
\cref{eq:def_tilde_G_W} originates directly from the difference of $ G(\Wbf) $ and $ G'(\Wbf) $.
We have
\begin{align*}
  \widetilde G(\Wbf)
  &= G(\Wbf) - \gamma G'(\Wbf) \\
  &= \twobytwo
     {\!\Psi_{1}^\top \left( I_{n_1} \kron {\phi^{(1)}_{0}}^\top \right)}
     {\Psi_{L}^\top \left( I_{n_L} \kron {\phi^{(1)}_{L-1}}^\top \right)\!}
     {\!\Psi_{1}^\top \left( I_{n_1} \kron {\phi^{(N)}_{0}}^\top \right)}
     {\Psi_{L}^\top \left( I_{n_L} \kron {\phi^{(N)}_{L-1}}^\top \right)\!}
     - \gamma
     \twobytwo
     {\!{\Psi_{1}'}^\top \left( I_{n_1} \kron {\phi'^{(1)}_{0}}^\top \right)}
     {{\Psi_{L}'}^\top \left( I_{n_L} \kron {\phi'^{(1)}_{L-1}}^\top \right)\!}
     {\!{\Psi_{1}'}^\top \left( I_{n_1} \kron {\phi'^{(N)}_{0}}^\top \right)}
     {{\Psi_{L}'}^\top \left( I_{n_L} \kron {\phi'^{(N)}_{L-1}}^\top \right)\!}\\
  &= \threebythree
     {\!\!\!
      \Psi_{1} ^\top \!\left( I_{n_1} \!\kron {\phi _{0}}^{(1)} \!\right)^\top \!\! - \gamma
      \Psi_{1}'^\top \!\left( I_{n_1} \!\kron {\phi'_{0}}^{(1)} \!\right)^\top}
     {\!\!\!\!\!\ldots\!\!\!\!\!}
     {\Psi_{L} ^\top \!\left( I_{n_L} \!\kron {\phi _{L-1}}^{(1)} \!\right)^\top \!\! - \gamma
      \Psi_{L}'^\top \!\left( I_{n_L} \!\kron {\phi'_{L-1}}^{(1)} \!\right)^\top
      \!\!\!}
     {\vdots}{\!\!\!\!\!\ddots\!\!\!\!\!}{\vdots}
     {\!\!\!
      \Psi_{1} ^\top \!\left( I_{n_1} \!\kron {\phi _{0}}^{(N)} \!\right)^\top \!\! - \gamma
      \Psi_{1}'^\top \!\left( I_{n_1} \!\kron {\phi'_{0}}^{(N)} \!\right)^\top}
     {\!\!\!\!\!\ldots\!\!\!\!\!}
     {\Psi_{L} ^\top \!\left( I_{n_L} \!\kron {\phi _{L-1}}^{(N)} \!\right)^\top \!\! - \gamma
      \Psi_{L}'^\top \!\left( I_{n_L} \!\kron {\phi'_{L-1}}^{(N)} \!\right)^\top
      \!\!\!}
\end{align*}
by pairing each block in the matrices.

\section{Raw Data}
\label{sec:raw-data}
To emphasize the difficulty in visualising the results of our Policy Iteration experiment, we
provide here the raw output of all five individual repetitions for $ N = 100 $ samples with
$ i = 3500 $ Policy Evaluations steps when using the persistent setting.
\begin{figure}[H]
  \centering
  \begin{subfigure}{0.20\linewidth}
    \includegraphics[width=1.0\linewidth]{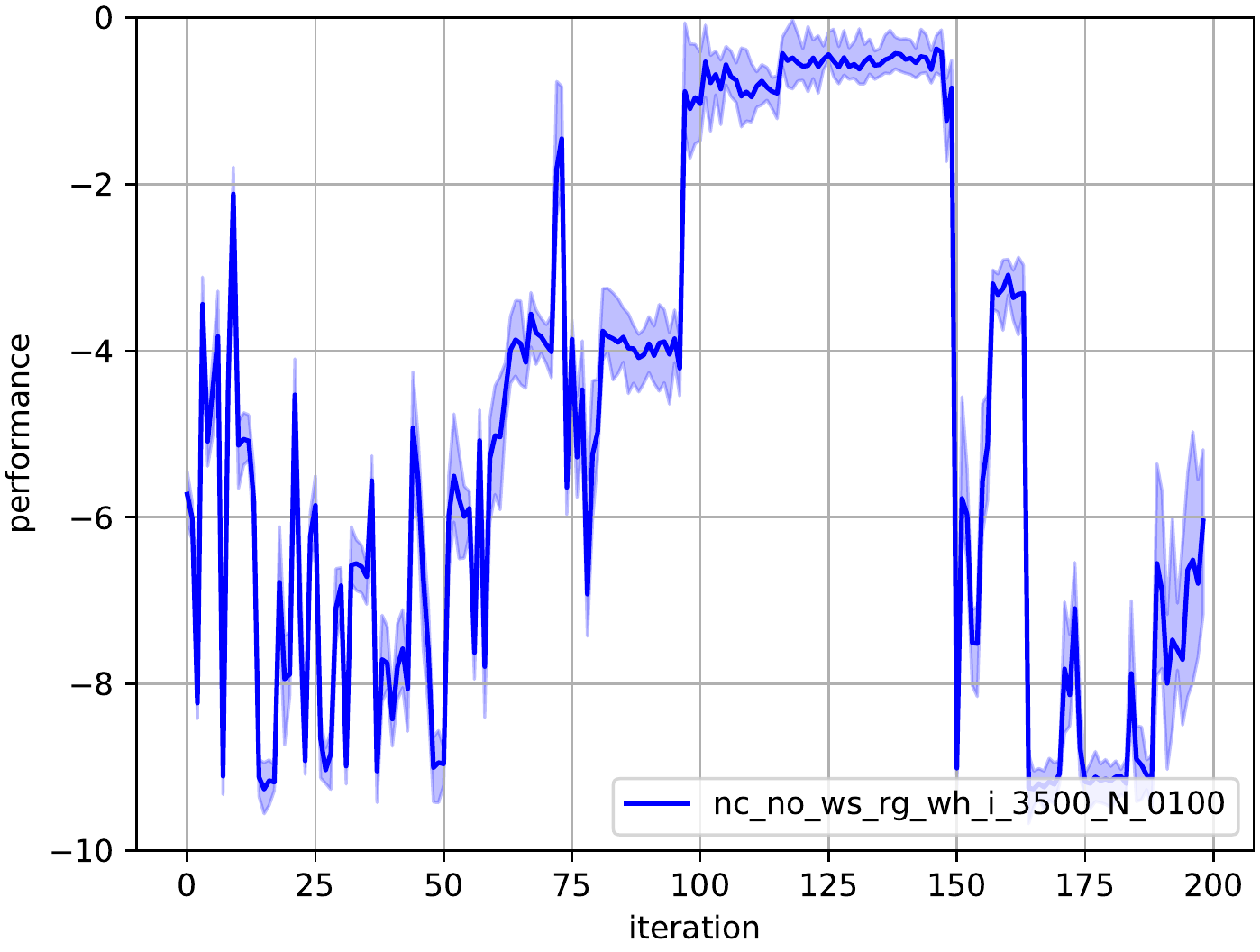}
  \end{subfigure}%
  \begin{subfigure}{0.20\linewidth}
    \includegraphics[width=1.0\linewidth]{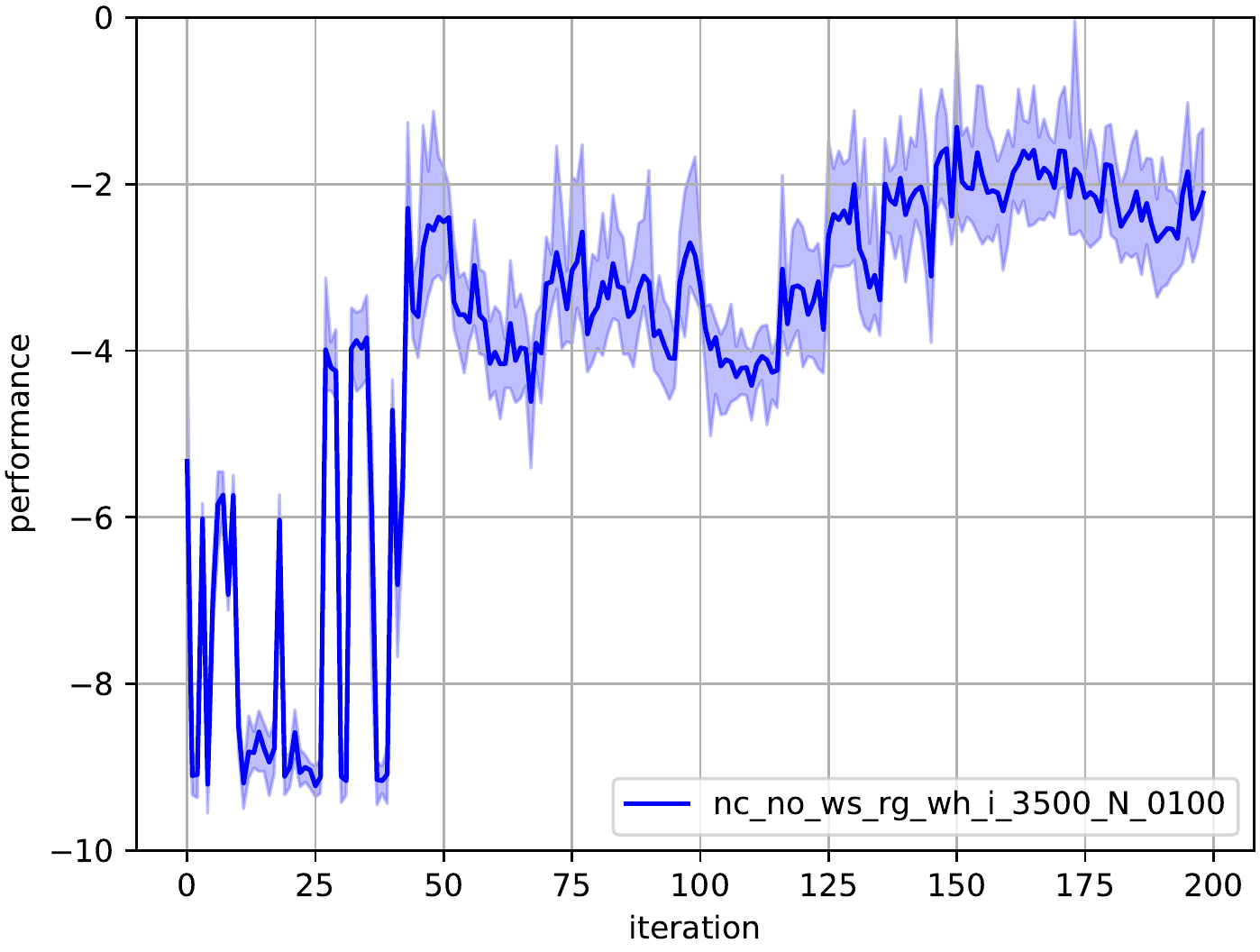}
  \end{subfigure}%
  \begin{subfigure}{0.20\linewidth}
    \includegraphics[width=1.0\linewidth]{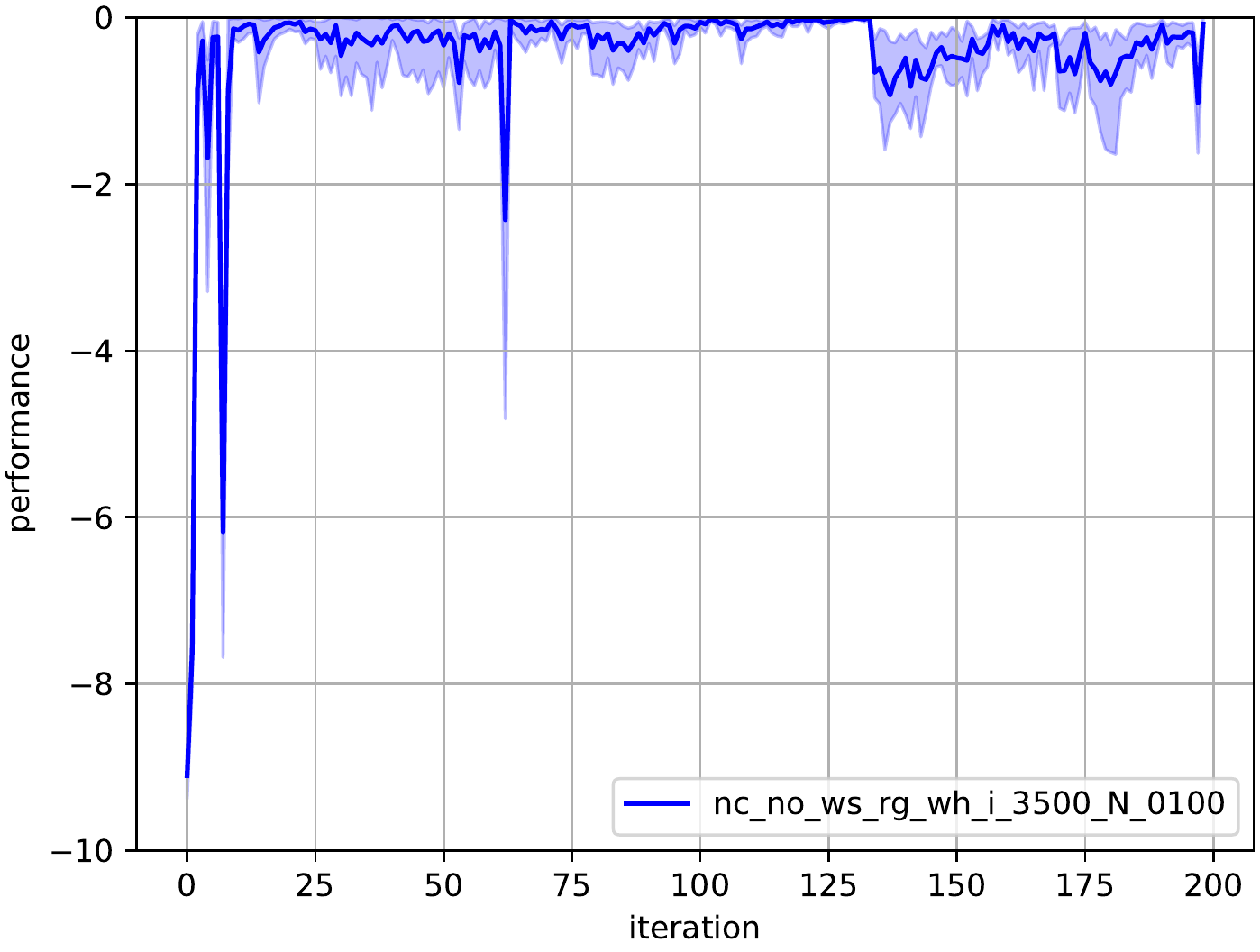}
  \end{subfigure}%
  \begin{subfigure}{0.20\linewidth}
    \includegraphics[width=1.0\linewidth]{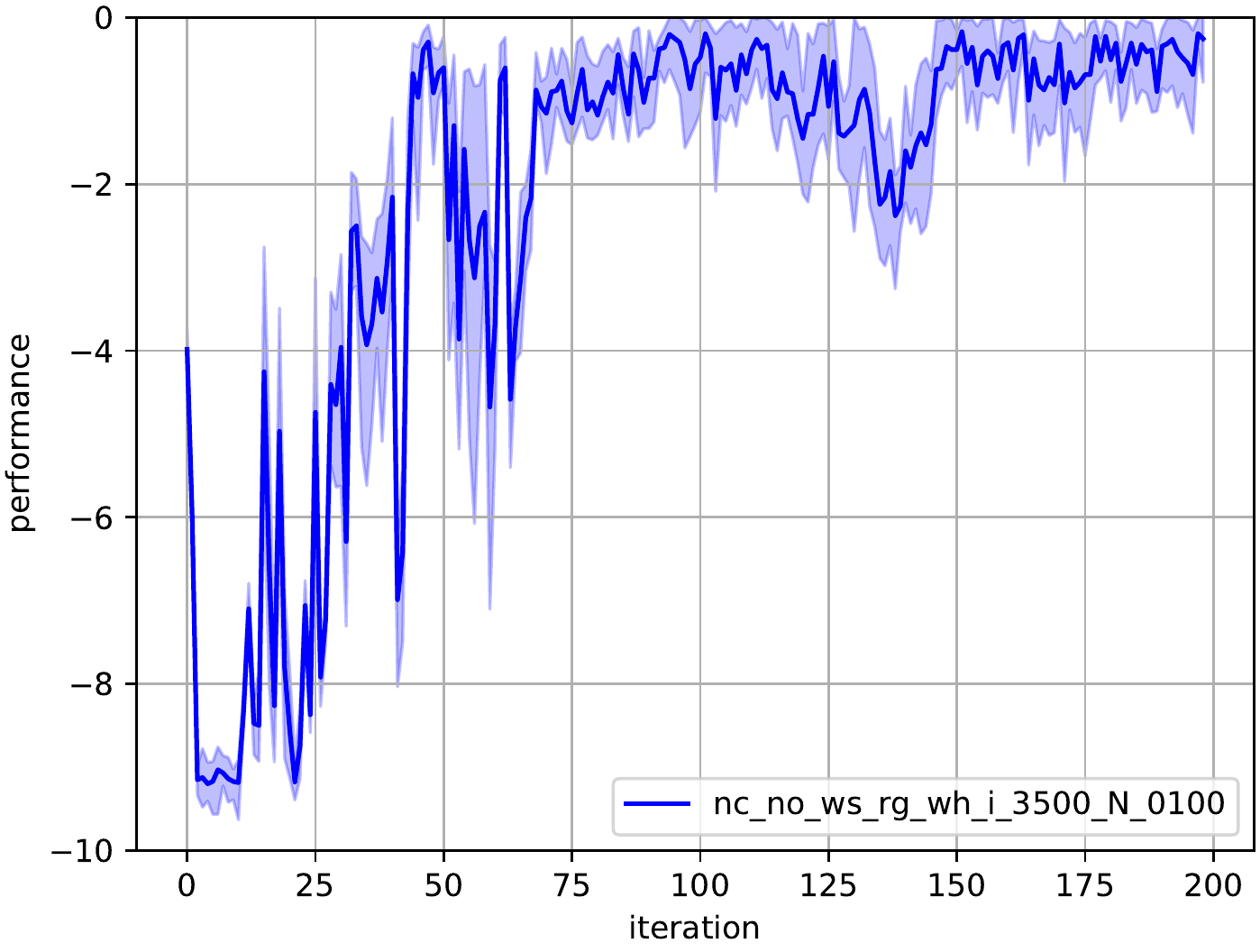}
  \end{subfigure}%
  \begin{subfigure}{0.20\linewidth}
    \includegraphics[width=1.0\linewidth]{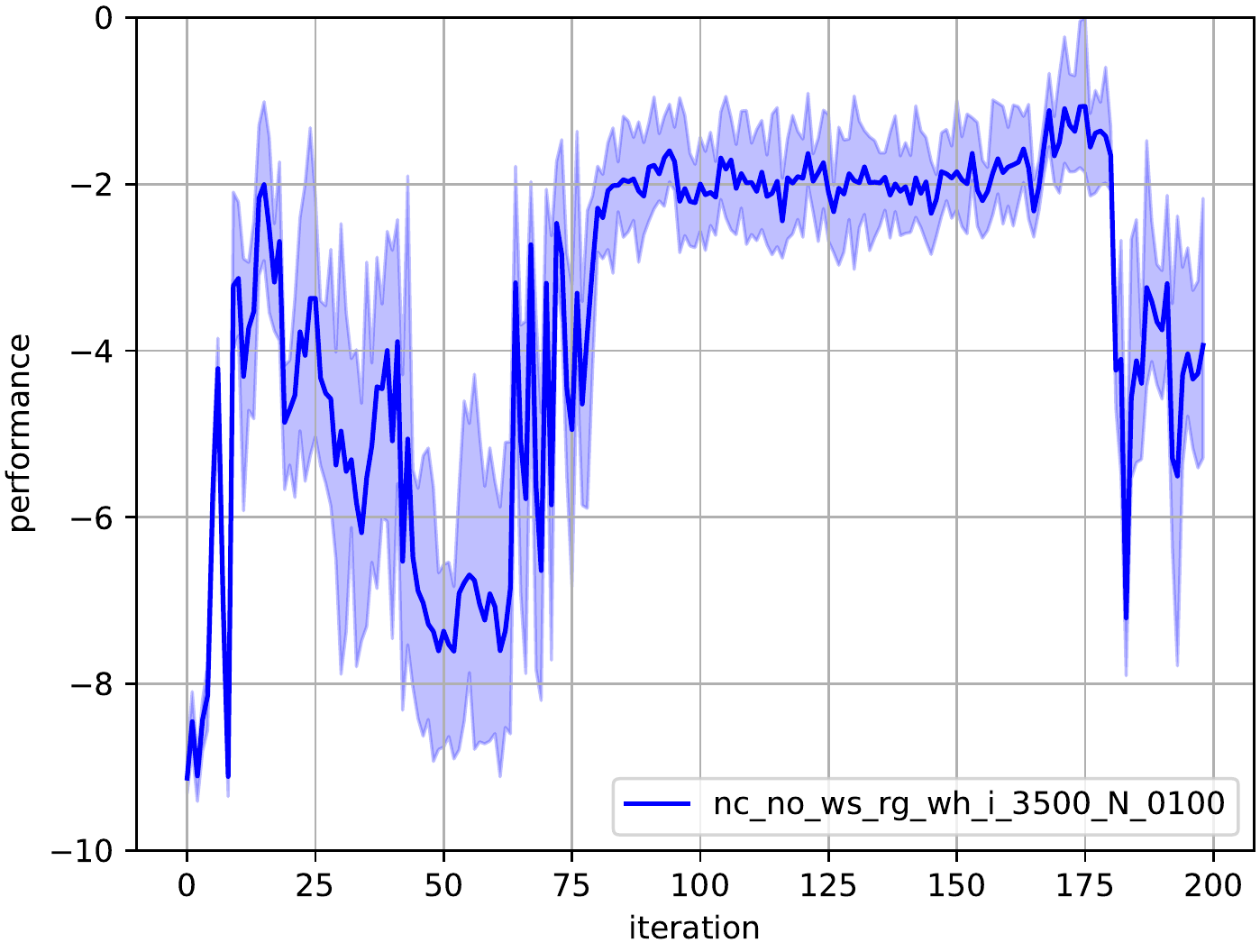}
  \end{subfigure}
\end{figure}
All repetitions of the roll-outs per run produce reliably similar discounted returns, but running
again the whole experiment produces varying performance curves.
Unfortunately, these curves can be rather dissimilar.
Thus, by averaging these curves, we can highlight the trend for the performance over time in the
overall experiment and produce the plots shown in \cref{fig:policy_iteration}.

\end{document}